\documentclass[twoside,11pt]{article}

\usepackage[T1]{fontenc}
\usepackage{amsmath,amsthm,mathtools,bm}
\usepackage[preprint,nohyperref]{jmlr2e}
\usepackage{booktabs,microtype,float,array,multirow}
\usepackage[section]{placeins}
\usepackage[ruled,vlined,linesnumbered]{algorithm2e}
\usepackage{longtable,etoc}
\usepackage[most]{tcolorbox}
\newtcolorbox{roadmap}[1][]{enhanced,breakable,colback=black!3,colframe=black!25,
  boxrule=0.5pt,arc=1.5pt,left=6pt,right=6pt,top=4pt,bottom=4pt,
  fonttitle=\bfseries\small,coltitle=black,colbacktitle=black!8,
  title={Roadmap#1},fontupper=\small}
\usepackage[colorlinks=false,allbordercolors={1 1 1}]{hyperref}
\hypersetup{
  pdftitle={Decision-Observable Reinforcement Learning: Priced Audits, Verified Outcomes, and Certified Memory},
  pdfauthor={Rui Wang}
}
\newcommand{\E}{\mathbb E}
\newcommand{\Prb}{\mathbb P}
\newcommand{\R}{\mathbb R}
\newcommand{\KL}{\operatorname{KL}}
\newcommand{\cM}{\mathcal M}
\newcommand{\cS}{\mathcal S}
\newcommand{\cF}{\mathcal F}
\newcommand{\cC}{\mathcal C}
\newcommand{\cG}{\mathcal G}
\newcommand{\one}{\mathbf 1}
\newcommand{\sgn}{\operatorname{sgn}}
\newcommand{\Reg}{\operatorname{Reg}}
\newcommand{\pihat}{\widehat\pi}
\newcommand{\Rhat}{\widehat R}
\DeclareMathOperator*{\argmax}{arg\,max}

\ShortHeadings{Decision-Observable Reinforcement Learning}{Wang}
\firstpageno{1}

\begin{document}
\title{Decision-Observable Reinforcement Learning:\\
Priced Audits, Verified Outcomes, and Certified Memory}
\author{\name Rui Wang \email wangrui@microsoft.com\\
  \addr Vancouver, BC, Canada}
\maketitle
\thispagestyle{plain}

\begin{abstract}
Many sequential decisions depend on hidden facts cued long before the decision, and the outcome that matters
can be checked only at a cost. We formalize \emph{decision-observable reinforcement learning} (DORL). The learner sees a
free but possibly biased proxy, can buy audits that reveal hidden events during an episode, and can buy a verifier
that reveals the outcome afterwards. It pays for every episode, audit, and label, and returns a policy with a
certificate on its regret against the best full-history policy. For finitely many candidate worlds we derive the
exact leading cost of certification. For a class in which the outcome is a product of $k$ noisy parities of early
reports, learning from verified outcomes costs order $\gamma^{-2k}$, whereas audits that supervise the memory cost
order $\gamma^{-2}$. This motivates DORL-PPO, a recurrent policy-gradient method with a doubly-robust verified return,
audit replies as auxiliary targets for the memory, and anytime certificates. At matched evidence charge, DORL-PPO
learns multi-event memory tasks on which every outcome-supervised baseline stays at the memory-free value, and cuts
regret from about $0.3$ to below $0.07$. The gain persists without reinforcement learning and with a language-model
policy, and sparse verification prevents reward hacking in language-model fine-tuning.
\end{abstract}

\etocdepthtag.toc{mtmain}
\etocsettagdepth{mtmain}{subsection}
\etocsettagdepth{mtappendix}{section}
\section{Introduction}\label{sec:intro}

Reinforcement learning works best when the reward is cheap to observe, correct, and computed from a state the agent can
see. Board games, simulators, and rule-checked mathematics have all three properties. Many sequential tasks have none
of them. A software agent sees a cheap self-assessment of its patch, while task success is confirmed only by an
expensive end-to-end test. An inspection robot sees a door that looks the same whether or not the room behind it is
safe, and it can pay for a sensor reading before it opens the door. A clinician acts on a history whose relevant part
was reported many steps earlier, and can order a laboratory test before choosing a treatment.

\begin{figure}[t]
\centering
\includegraphics[width=\linewidth]{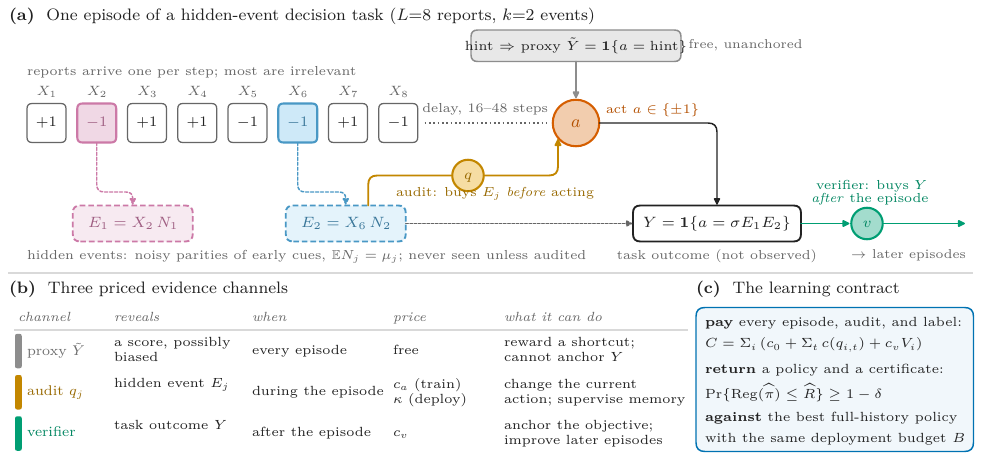}
\caption{\textbf{The decision-observable setting.} \textbf{(a)} Early reports $X_t$ arrive one per step. Each hidden event
$E_j$ is a noisy function of a few highlighted cues and stays hidden unless audited. After a delay the agent acts, and
the outcome $Y$ depends on the product of the events. The free proxy rewards following a hint that is unrelated to $Y$.
\textbf{(b)} The evidence channels differ in what they reveal, when, and at what price. \textbf{(c)} The learner pays
for every episode, audit, and verification ($V_i=1$) and returns a policy with a regret certificate against the best
full-history policy with the same deployment budget.}\label{fig:setting}
\end{figure}

Figure~\ref{fig:setting} shows the structure these examples share. The right decision depends on \emph{hidden events},
each predictable from an early cue that the agent must remember. The free signal is a \emph{proxy} that may reward a
shortcut. The agent can buy two kinds of evidence. An \emph{audit} during the episode reveals one hidden event and can
change the current action; a \emph{verifier} after the episode reveals the task outcome and can only improve later
episodes. Memory-based methods for partially observed problems assume a correct reward \citep{kaelbling1998,ni2022},
reward-modelling methods optimize a fitted proxy \citep{gao2023}, and active-observation methods price observations
under a known objective \citep{nam2021,krueger2020,parisi2024}. We treat the three difficulties together and organize
the learning problem around the decision the agent has to make.

\begin{figure}[t]
\centering
\includegraphics[width=\linewidth]{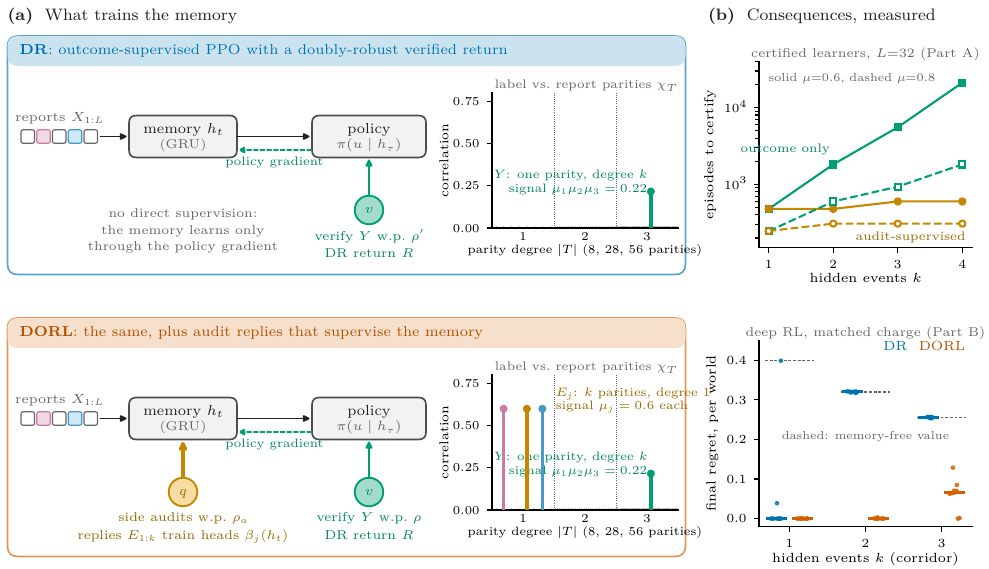}
\caption{\textbf{What trains the memory in DR and in DORL.} \textbf{(a)} Both learners share the recurrent network, the
joint action-and-audit space, PPO, and the doubly-robust verified return. DR trains the memory only through the policy
gradient of verified outcomes. DORL also trains event-belief heads $\beta_j(h_t)$ on side-audit replies at every step, at
the same nominal evidence charge. The insets show the population correlation of
each training label with every report parity $\chi_T$ ($k=3$, $L=8$, $\mu=0.6$). The verified outcome has one nonzero
coefficient, of degree $k$ and size $\prod_j\mu_j$. The audit replies have $k$ coefficients of degree one and size
$\mu_j$. \textbf{(b)} Measured consequences. Top: median episodes to certify the optimal policy for the two certified
routes of Theorem~\ref{thm:upper} (Part~A, $L=32$, $100$ worlds per point). Bottom: final regret of DR and DORL-PPO in each
of the ten worlds of the corridor tasks with $k=1,2,3$ (Part~B); dashed lines mark the memory-free value
$V^\star_0-\frac12$.}\label{fig:drdorl}
\end{figure}

\paragraph{Verified outcomes and memory.}
A strong outcome-supervised learner verifies a fraction of episodes and corrects a learned outcome model with them.
The resulting \emph{doubly-robust} (DR) return is unbiased for the task outcome \citep{dudik2011,jiang2016}, so DR
cannot be hacked through the proxy. It still has to learn the memory from the policy gradient. When the outcome is a
product of $k$ hidden events with reliabilities $\mu_j$, the verified label correlates with a single function of the
reports, the parity of all $k$ cues, at strength $\prod_j\mu_j$, and with every lower-order feature at strength zero
(Figure~\ref{fig:drdorl}a). An audit reply correlates with its own cue at strength $\mu_j$. Audits bought during
training are therefore a much cheaper teacher for the memory, even if the deployed policy never buys one. We make this
precise as a statistical separation, build an algorithm on it, and test it.

\paragraph{Contributions.}
\begin{enumerate}
\item \emph{A learning contract} (\S\ref{sec:model}). The learner pays for every episode, audit, and verification and
returns a deployable joint action-and-audit policy $\pihat$ with a certificate $\Rhat$: with probability $1-\delta$, the
regret of $\pihat$ against the best full-history policy with the same deployment audit budget is at most $\Rhat$. We
measure learners by the charge they need to certify a decision. Three boundary results fix the role of each channel:
verification is necessary when worlds with reversed outcomes look alike, deployment audits are necessary to exceed the
reply-blind optimum, and unaudited events are worthless to a policy that ignores the reports.
\item \emph{Finite-world foundations} (\S\ref{sec:foundations}). With known candidate report laws, the cost of certifying
an $\varepsilon$-good decision is the reciprocal of a charge-normalized exclusion rate over a \emph{decision complex}, the
sets of worlds that one deployable policy can serve. The same quantity gives a Goodhart gap for proxy optimization, an
exact price for report-adaptive audits, and least-favorable deployment certificates.
\item \emph{A priced information law for learned structure} (\S\ref{sec:class}--\ref{sec:theory}). For outcomes that are
products of $k$ noisy parities of unknown reports, an anytime-valid learner recovers the cues from audit replies and
certifies a one-bit decision register, and an information-per-price lower bound matches its signal dependence.
Certifying from verified outcomes alone costs order $\gamma^{-2k}$, against $\gamma^{-2}$ with audits, and we give the
crossover audit price.
\item \emph{DORL-PPO} (\S\ref{sec:deep}). Recurrent PPO over joint actions and audits with a doubly-robust verified
return, audit replies as auxiliary targets for the recurrent state, and anytime empirical-Bernstein certificates of the
deployed policy, which can also certify that the policy acts on its audit replies.
\item \emph{An empirical study in five parts} (\S\ref{sec:experiments}--\ref{sec:results}), from certified learners on
random worlds to deep RL and two language-model studies.
\end{enumerate}

\paragraph{Results in brief.}
The certified learners produced $3{,}800$ certificates on $2{,}200$ random worlds, all valid, and their charges grew as the
theory predicts (fitted log-log slopes $1.03$ and $1.06$; predicted $1$). At $k=4$ the outcome-only route cost $87$ times
more than the audit route. In four multi-event corridor tasks, DORL-PPO reached regret $0.00$--$0.07$ while every
outcome-supervised PPO baseline at matched evidence charge stayed at the memory-free value ($0.25$--$0.32$), including
fully verified PPO at five to six times the charge. DORL won in all ten paired worlds of each task. The gain survived
four times longer outcome-only training and an equal training-charge cap, and supervised learners without RL showed the
same split: audit labels solved $10/10$ worlds and outcome labels $0/10$. In a grid world that adds navigation to the
cues, DORL was ahead without statistical significance. With a Qwen2.5 policy, DORL reduced paired regret against DR by
$0.21$ and $0.20$ at $k=2$ and $3$. In arithmetic fine-tuning with an exploitable reward model, proxy-only training fell
from $0.82$ to $0.32$ accuracy, while verifying $5\%$ of responses with a DR return reached $0.98$, the accuracy of full
verification.

\paragraph{Organization.}
Section~\ref{sec:related} reviews related work and Section~\ref{sec:model} defines the contract.
Section~\ref{sec:foundations} develops the finite-world theory, Sections~\ref{sec:class}--\ref{sec:theory} the
hidden-event class, and Section~\ref{sec:deep} DORL-PPO. Sections~\ref{sec:experiments} and~\ref{sec:results} report the
experiments, and Section~\ref{sec:discussion} discusses applications and future directions. Proofs and experimental details are in the
appendices.
\section{Related work}\label{sec:related}

\paragraph{Partial observability and memory.}
Belief-state and predictive-state methods learn a state that is sufficient for the model
\citep{kaelbling1998,liu2022,chen2022}. The theory of approximate information states and state abstraction orders
history summaries from model irrelevance down to optimal-action irrelevance \citep{li2006,subramanian2022}, and decodable
POMDPs exploit short memory \citep{efroni2022}. Recurrent model-free RL is a strong baseline when tuned
\citep{ni2022}. Self-predictive objectives shape the memory by predicting future observations
\citep{ni2024}, and auxiliary tasks more generally can accelerate representation learning \citep{jaderberg2017}. Our
decision register is an optimal-action abstraction learned from \emph{priced} evidence: the auxiliary targets of
DORL-PPO are purchased replies about decision-relevant hidden events. In our experiments a self-predictive auxiliary
loss does not find the decision-relevant bits.

\paragraph{Privileged information.}
Training-time information that is unavailable at deployment is used in learning with privileged information
\citep{vapnik2009} and in asymmetric actor-critic methods \citep{pinto2018}. Side audits play this role in DORL-PPO,
but their price enters the training charge, and the theory shows when buying them is cheaper than buying outcomes. The
same channel can also be bought at deployment, where its replies change the action.

\paragraph{Costly observation and monitored rewards.}
Active-measurement RL prices state observations \citep{nam2021,shi2026}. Active reinforcement learning prices reward
observation \citep{krueger2020}, and monitored MDPs study rewards that are observed only through a monitor
\citep{parisi2024}. None of these separates in-episode audits from terminal verification, or proves a priced lower bound
against an outcome-only comparator.

\paragraph{Reward misspecification and proxy optimization.}
Corrupted and perturbed reward channels \citep{everitt2017,wang2020perturb}, reward hacking
\citep{skalse2022,laidlaw2025}, and reward-model overoptimization \citep{gao2023} motivate our requirement that the
objective be anchored by a verifier. Our doubly-robust return follows off-policy evaluation
\citep{robins1994,dudik2011,jiang2016}, and our use of a learned model only as a control variate parallels
prediction-powered and active inference \citep{zrnic2024}.

\paragraph{Process versus outcome supervision.}
Empirical work favours process supervision for language-model reasoning \citep{uesato2022,lightman2024}.
\citet{jia2025} prove that outcome supervision is statistically no harder than process supervision when the per-step
rewards are functions of observed states. In our setting audits reveal hidden events that are \emph{not} functions of
the report history, and their price enters the bound explicitly. Intermediate supervision makes otherwise hard
compositions learnable \citep{wies2023}, and gradient descent on parities is governed by leap complexity
\citep{abbe2023}. Our gradient calculation is a local analogue of the latter.

\paragraph{Identification, testing, and certification.}
Our finite-world cost theorem is a fixed-confidence identification result in the tradition of best-arm and
pure-exploration lower bounds \citep{chernoff1959,garivier2016,kaufmann2016,degenne2019}. In our setting the answer is a
\emph{decision complex} of world sets served by one randomized policy rather than a single correct arm, and the
experiments are priced report-adaptive episodes. Minimax regret over a set of compatible models
is classical in partial identification \citep{manski2004,stoye2009}. We compute it with sequence-form linear programs
\citep{koller1996,vonstengel1996}. Our anytime certificates use empirical-Bernstein confidence sequences
\citep{howard2021,waudbysmith2024}. Our reporting follows recommendations for reliable deep RL evaluation
\citep{agarwal2021}.
\section{Decision-observable reinforcement learning}\label{sec:model}

\subsection{Episodes, evidence channels, and the deployment comparator}

\begin{definition}[Hidden-event decision process]\label{def:hedp}
An episode of horizon $H$ has a latent trajectory $z_{1:H}$, ordinary reports $o_t$, and at each step a
\emph{joint} choice $u_t=(q_t,a_t)$ of an audit $q_t\in\mathcal Q\cup\{\varnothing\}$ and a physical action
$a_t\in\mathcal A$. An audit $q\ne\varnothing$ returns a reply $r_t$ before the next choice. A free proxy
$\tilde Y$ is one of the reports. The task authority fixes a bounded \emph{task outcome} $Y=\phi(z_{1:H},a_{1:H})\in[0,1]$
before learning. The deployment utility is $U=Y-\sum_t\kappa(q_t)$ with known audit prices $\kappa\ge0$.
\end{definition}
The report history $h_t$ contains reports, own choices, and replies, but not $z$ or $Y$. For a deployment
budget $B$, the \emph{deployment class} $\Pi_B$ consists of all (possibly randomized) maps from report
histories to joint choices that buy at most $B$ audits on every path. For a world $M$,
$J_M(\pi)=\E^\pi_M U$ and $V^\star_{M,B}=\sup_{\pi\in\Pi_B}J_M(\pi)$; regret is $\Reg_M(\pi)=V^\star_{M,B}-J_M(\pi)$.
The comparator thus remembers the \emph{entire} report history and has the same audit menu and pathwise
budget as the deployed policy.

\paragraph{Training access and the charge ledger.}
A learner interacts with reset episodes of the unknown world $M\in\cM$. In episode $i$ it may follow any
history-dependent policy, buy any audits (the deployment budget does not constrain training), and after
the episode buy a \emph{verification} that returns a label $Y^{\rm ver}$, equal to $Y$ or flipped with known
probability $\eta<1/2$. It never observes $z$, unbought replies, or unverified outcomes. It stops at a time $\tau$ of
its choosing. The \emph{training charge} is
\begin{equation}\label{eq:charge}
C=\sum_{i\le\tau}\Big(c_0+\sum_t c(q_{i,t})+c_v\,\one\{\text{episode $i$ verified}\}\Big),
\end{equation}
with per-episode, per-audit, and per-verification prices $c_0,c(\cdot),c_v$. Proxy-fitting labels and
certification episodes are charged in the same ledger. In the experiments we split \eqref{eq:charge} into the
\emph{training charge} of the episodes used to fit the policy and the \emph{certification charge} of the verified
deployment episodes used for the certificate (with their audits); the \emph{total charge} is their sum.

\begin{definition}[Certified decision learning]\label{def:dorl}
A learner is \emph{$\delta$-sound} on $\cM$ if it outputs $(\pihat,\Rhat)$ with $\pihat\in\Pi_B$ and
$\Prb_M\{\Reg_M(\pihat)\le\Rhat\}\ge1-\delta$ for every $M\in\cM$. Its \emph{cost to certify accuracy
$\varepsilon$} in world $M$ is $\E_M[C]$ for a sound learner that stops only when $\Rhat\le\varepsilon$.
\end{definition}
The certificate must hold for the deployed policy itself. When the available evidence cannot separate worlds
that require different decisions, a sound learner must report a large $\Rhat$.

\subsection{Boundary: when verification is necessary}\label{sec:boundaries}

\begin{proposition}[No anchor without verification]\label{prop:anchor}
Let $M,\bar M\in\cM$ be two worlds in which, under every policy, the joint law of reports (including the
proxy) and audit replies is the same, the outcomes satisfy $Y_{\bar M}\le1-Y_M$ pathwise (for example, complementary
on decided episodes and both zero on timed-out ones), and $V^\star_{M,B}=V^\star_{\bar M,B}=V^\star$. Then every learner that buys no verification satisfies
\[
\max_{W\in\{M,\bar M\}}\Prb_W\big\{\Reg_W(\pihat)\ge V^\star-\tfrac12\big\}\ge\tfrac12 .
\]
In particular no such learner is $\delta$-sound with $\Rhat<V^\star-\tfrac12$ for any $\delta<\tfrac12$,
whatever its proxy and audit data.
\end{proposition}
The proof (Appendix~\ref{app:boundaries}) uses $J_M(\pi)+J_{\bar M}(\pi)\le1$ for every $\pi$. Verification is
therefore necessary for a uniform guarantee in any class that contains an outcome-reversed pair, and unnecessary in a
class where reports or audits identify the task outcome, for instance because the polarity is known. In the class
below and in both deep RL environments, flipping the polarity $\sigma$ produces such a pair, so audit replies alone
cannot anchor the task outcome. The \texttt{proxy+audit-aux} baseline in \S\ref{sec:experiments} is consistent with
this bound.
\providecommand{\cE}{\mathcal E}
\providecommand{\cK}{\mathcal K}
\providecommand{\cT}{\mathcal T}
\section{Finite-world foundations: the cost of a certified decision}\label{sec:foundations}

Definition~\ref{def:dorl} asks for a deployed policy with a valid regret certificate, and the
ledger~\eqref{eq:charge} charges every episode, audit, and verification used to obtain it. Before studying a class
whose structure must be learned, we compute this cost when the structure is known: the world is one of finitely many
candidates with known report laws, and only its identity is uncertain. The cost of certifying an $\varepsilon$-good
decision then has an exact leading constant, the reciprocal of the charge-normalized rate at which evidence excludes
the worlds that a single deployable policy cannot serve. The evidence channels of Definition~\ref{def:hedp} enter this
rate in different places (\S\ref{sec:fw-channels}), so we can compare them at their actual prices within one quantity
and identify when no amount of training evidence can replace a deployment monitor.

\subsection{Known candidate laws}\label{sec:fw-contract}

Let $\Theta$ be a finite set of at least two candidate worlds; $\theta\in\Theta$ indexes a process $M_\theta\in\cM$ of
Definition~\ref{def:hedp}, and the true $\theta$ is fixed and unknown. A \emph{training experiment} $e$ is a complete
protocol for one reset episode: a report-adaptive policy for the joint choices $u_t=(q_t,a_t)$ and a report-based rule
for buying terminal verification. Its \emph{transcript} $Y$, everything the learner observes in the episode (reports,
proxy, own choices, audit replies, and the verifier label if bought), lies in a finite set $\mathcal Y_e$. The learner
draws experiments from an explicit finite menu $\cE$. Under $\theta$ the transcript has a \emph{known} law
$P^e_\theta$; for each $e$ the candidate laws share one support, on which they are at least $p_{\min}>0$. Purchases are
visible, so the charge~\eqref{eq:charge} of an episode is a function $k_e(Y)\in[c_0,\bar c]$ of its transcript, with
$c_0>0$. Write $\bar k_\theta(e)=\E_\theta k_e(Y)$ and $D_e(\theta,\lambda)=\KL(P^e_\theta\Vert P^e_\lambda)$. Episodes
are conditionally independent given the world and the chosen experiments. Worlds are \emph{training-equivalent},
$\theta\sim\lambda$, if $P^e_\theta=P^e_\lambda$ for all $e\in\cE$; $[\theta]$ is the class of $\theta$.

For deployment, let $\Pi\subseteq\Pi_B$ be a finite set of deterministic policies feasible under the declared
deployment monitor, pathwise budget $B$, and any memory restriction, and let $X=\Delta(\Pi)$; a point $x\in X$ is
deployed by drawing one policy from $x$ at the start of an episode, so $J_\theta(x)=\sum_\pi x(\pi)J_{M_\theta}(\pi)$.
On an explicit perfect-recall history tree with all deterministic policies in $\Pi$, the vectors
$(J_\theta(x))_\theta$ are exactly the value vectors of policies in $\Pi_B$ \citep{kuhn1953}. Comparators satisfy
$\bar V_\theta\ge\max_XJ_\theta$, by default $\bar V_\theta=V^\star_{M_\theta,B}$, and
$\Reg_\theta(x)=\bar V_\theta-J_\theta(x)$; two deployment classes are compared with the same comparator. A learner
chooses experiments from past transcripts and internal randomness, stops at a stopping time $\tau\le\infty$, outputs
$\hat x\in X$ (possibly at random), and pays $C=\sum_{i\le\tau}k_{e_i}(Y_i)$. It is \emph{$(\varepsilon,\delta)$-sound}
if $\Prb_\theta\{\tau<\infty,\ \Reg_\theta(\hat x)>\varepsilon\}\le\delta$ for every $\theta\in\Theta$. A learner that is
$\delta$-sound as in Definition~\ref{def:dorl} and stops only when $\Rhat\le\varepsilon$ is $(\varepsilon,\delta)$-sound,
so the lower bound below applies to the cost in Definition~\ref{def:dorl}.

\subsection{The certification radius and the decision complex}\label{sec:fw-complex}

For $S\subseteq\Theta$ and $\nu\in\Theta$ define the \emph{certification radius} and the \emph{good sets}
\begin{equation}\label{eq:fw-radius}
R(S;X)=\min_{x\in X}\ \max_{\nu\in S}\ \Reg_\nu(x),\qquad
G_\nu(\varepsilon)=\{x\in X:\Reg_\nu(x)\le\varepsilon\},
\end{equation}
with $R(\varnothing;X)=-\infty$. The radius is the regret one deployable policy can guarantee against every world in
$S$. If a set $C$ contains the truth with probability at least $1-\delta$, a minimizer of $R(C;X)$ has regret at most
$R(C;X)$ with at least that probability, which is a certificate in the sense of Definition~\ref{def:dorl}. Minimax
regret over compatible models is classical \citep{manski2004,stoye2009}; our results concern the combinatorics of the
sets that a single policy can serve.

\begin{definition}[Decision complex]\label{def:fw-complex}
For $\varepsilon\ge0$, the \emph{decision complex} is
$\cK_\varepsilon(X)=\{S\subseteq\Theta:R(S;X)\le\varepsilon\}=\{S\subseteq\Theta:\bigcap_{\nu\in S}G_\nu(\varepsilon)\ne\varnothing\}$.
Its members are \emph{faces}, its maximal faces are \emph{facets}, and its inclusion-minimal non-faces are
\emph{conflict sets}.
\end{definition}

The complex is the nerve of the convex sets $G_\nu(\varepsilon)$: it records which groups of worlds one randomized
deployable policy can serve to within $\varepsilon$, and nothing else about the payoffs. Let $d_X$ be the affine
dimension of the payoff set $J(X)=\{(J_\nu(x))_{\nu\in\Theta}:x\in X\}$; it is at most $\min\{|\Theta|,|\Pi|-1\}$ and
is often much smaller than $|\Theta|$.

\begin{proposition}[Geometry of the decision complex]\label{prop:fw-geometry}
\leavevmode
\begin{enumerate}
\item[(i)] $\cK_\varepsilon(X)$ is closed under taking subsets, is nondecreasing in $\varepsilon$ and in $X$ at a fixed
comparator, and changes only at the finitely many values $R(S;X)$, $S\subseteq\Theta$.
\item[(ii)] \textup{(Helly witness.)} Every conflict set has at most $d_X+1$ worlds, and for every nonempty
$C\subseteq\Theta$, $R(C;X)=\max\{R(S;X):S\subseteq C,\ |S|\le d_X+1\}$.
\item[(iii)] \textup{(Least-favorable worlds.)} For nonempty $C$,
\[
R(C;X)=\max_{\mu\in\Delta(C)}\Big[\sum_{\nu\in C}\mu_\nu\bar V_\nu-\max_{x\in X}\sum_{\nu\in C}\mu_\nu J_\nu(x)\Big],
\]
and a maximizing $\mu$ exists with at most $d_X+1$ support points. With the default comparator, $R(C;X)=0$ if and
only if one $x\in X$ is optimal in every world of $C$.
\item[(iv)] \textup{(Pairwise checks.)} If $d_X\le1$, a set is a face if and only if all its subsets of at most two
worlds are. For $d_X=2$ this fails: with three worlds and three actions valued $J_\nu(a)=\one\{a\ne\nu\}$, every pair
shares an optimal action, yet $R(\Theta;X)=1/3$; if the three worlds are training-equivalent, every learner that stops
almost surely outputs, in some world, a policy with regret at least $1/3$ with probability at least $1/3$.
\end{enumerate}
\end{proposition}

By (ii), a certificate must account for conflicts among up to $d_X+1$ worlds, and by (iv), pairwise certification is
unsound even with unlimited data. Part (iii) gives the least-favorable worlds that a learner reports when it cannot
certify. Proofs are in Appendix~\ref{app:fw-geometry}.

\subsection{The cost of certification}\label{sec:fw-cost}

For $A\subseteq\Theta$, define the \emph{exclusion rate}, and for the true world the \emph{certification rate} and the
\emph{characteristic cost},
\begin{gather}
\Gamma_\theta(A)=\max_{w\in\Delta(\cE)}\ \min_{\lambda\in A}\
\frac{\sum_ew_eD_e(\theta,\lambda)}{\sum_ew_e\bar k_\theta(e)},\label{eq:fw-rate}\\
\Gamma^\star_\varepsilon(\theta)=\max_{S\in\cK_\varepsilon(X):\ \theta\in S}\Gamma_\theta(\Theta\setminus S),\qquad
\cT_\varepsilon(\theta)=1/\Gamma^\star_\varepsilon(\theta),\label{eq:fw-cost}
\end{gather}
with $\min\varnothing=+\infty$, $\max\varnothing=0$, $1/0=\infty$, and $1/\infty=0$. The exclusion rate is the best
rate, in nats per unit of charge, at which a stationary mixture $w$ of experiments accumulates evidence against every
world in $A$. A learner need not identify $\theta$, only exclude every world outside some face containing $\theta$;
since $\Gamma_\theta(\Theta\setminus S)$ is nondecreasing in $S$, only facets matter. The denominator contains the
episode charge $c_0>0$, so a free verifier does not make unlimited episodes free.

\begin{lemma}[Exclusion rates]\label{lem:fw-rate}
(i) For nonempty $A$, $\Gamma_\theta(A)$ is the value of the linear program
$\max\{z:x\in\R_{\ge0}^{\cE},\ \sum_e\bar k_\theta(e)x_e=1,\ \sum_eD_e(\theta,\lambda)x_e\ge z\ \forall\lambda\in A\}$,
and $x/\sum_ex_e$ is an optimal design for any optimal $x$. (ii) For nonempty $A$,
$\Gamma_\theta(A)\le\log(1/p_{\min})/c_0$, and $\Gamma_\theta(A)>0$ if and only if $A\cap[\theta]=\varnothing$.
(iii) If $\nu\sim\theta$, then $\Gamma_\nu=\Gamma_\theta$.
\end{lemma}

\begin{theorem}[Finite-world certification cost]\label{thm:fw-cost}
Fix $\varepsilon\ge0$ and the true world $\theta$, and let $\varepsilon_{\rm free}=R(\Theta;X)$ and
$\varepsilon_{\rm imp}(\theta)=R([\theta];X)$.
\begin{enumerate}
\item[(i)] \textup{(Regimes.)} $\cT_\varepsilon(\theta)=0$ if $\varepsilon\ge\varepsilon_{\rm free}$;
$0<\cT_\varepsilon(\theta)<\infty$ if $\varepsilon_{\rm imp}(\theta)\le\varepsilon<\varepsilon_{\rm free}$; and
$\cT_\varepsilon(\theta)=\infty$ if $\varepsilon<\varepsilon_{\rm imp}(\theta)$. The cost depends on $\theta$ only
through $[\theta]$, and $\varepsilon\mapsto\cT_\varepsilon(\theta)$ is a nonincreasing step function whose jumps lie in
$\{R(S;X):S\subseteq\Theta\}$.
\item[(ii)] \textup{(Lower bound.)} Every $(\varepsilon,\delta)$-sound learner that stops $\Prb_\theta$-almost surely,
including those with adaptive experiments and randomized outputs, satisfies
$\liminf_{\delta\downarrow0}\E_\theta[C]/\log(1/\delta)\ge\cT_\varepsilon(\theta)$; a non-asymptotic form
is~\eqref{eq:fw-hp} in Appendix~\ref{app:fw-cost}.
\item[(iii)] \textup{(Matching learner.)} FaceTS (Algorithm~\ref{alg:fw-facets}) uses only transcripts and the known
laws. For every $\delta\in(0,1)$ and every true world it is $(\varepsilon,\delta)$-sound, and its output $(\hat x,\Rhat)$,
$\Rhat=R(C_\tau;X)\le\varepsilon$, is $\delta$-sound as in Definition~\ref{def:dorl}. If $\cT_\varepsilon(\theta)<\infty$,
then $\limsup_{\delta\downarrow0}\E_\theta[C]/\log(1/\delta)\le\cT_\varepsilon(\theta)$.
\end{enumerate}
\end{theorem}

\begin{algorithm}[t]
\caption{Face track-and-stop (FaceTS)}\label{alg:fw-facets}
\KwIn{$\Theta$; menu $\cE$ with known laws $P^e_\nu$ and charges $k_e$; payoffs $J_\nu(\pi)$, comparators $\bar V_\nu$;
$\varepsilon$; $\delta$}
$\ell\gets\log(1/\delta)$, $\beta\gets\log((|\Theta|-1)/\delta)$, $\epsilon_\ell\gets\min\{1/2,1/\ell\}$;
$L_n(\nu)\gets$ log-likelihood of the first $n$ transcripts under $\nu$\;
\textbf{stop} at the first $n\ge0$ (checked after every episode) with $R(C_n;X)\le\varepsilon$, where
$C_n=\{\nu:\max_vL_n(v)-L_n(\nu)\le\beta\}$; output a minimizer $\hat x$ of $R(C_n;X)$ and $\Rhat=R(C_n;X)$\;
\emph{pilot:} $\lceil\sqrt\ell/|\cE|\rceil$ cycles through $\cE$ ($n_0$ episodes); $\tilde\theta\gets\arg\max L_{n_0}$
(fixed tie-breaking)\;
\emph{face design:} a face $\tilde S\ni\tilde\theta$ and design $\tilde w$ attaining $\Gamma^\star_\varepsilon(\tilde\theta)$
(uniform $\tilde w$ if $\Gamma^\star_\varepsilon(\tilde\theta)=0$)\;
\emph{track:} draw each later experiment independently from $(1-\epsilon_\ell)\tilde w+\epsilon_\ell\,\mathrm{Unif}(\cE)$\;
\end{algorithm}

The stopping rule checks only whether the current confidence set is a face, so FaceTS never needs to identify $\theta$
or an optimal policy. The sets $C_n$ cover the truth uniformly in time (Ville's inequality), so a learner that stops
early for lack of budget still holds the minimax policy of $C_n$, the valid radius $R(C_n;X)$, and a witness of at most
$d_X+1$ worlds. The lower bound is a change-of-measure argument \citep{chernoff1959,garivier2016,kaufmann2016} in the
multiple-answer form of \citet{degenne2019}, normalized by cost as in \citet{kanarios2024}. Our contract changes the
answer set: with randomized deployment the correct answers are exactly the faces of $\cK_\varepsilon(X)$, a finite
family, and training-equivalent worlds are allowed. Proposition~\ref{prop:anchor} is an instance of the impossible
regime: without verification an outcome-reversed pair is training-equivalent, and $J_M(x)+J_{\bar M}(x)\le1$ forces
$R(\{M,\bar M\};X)\ge V^\star-\frac12$. With $\varepsilon=0$, the default comparator, and a unique optimal deterministic
policy in every world, $\cT_0(\theta)$ is the cost-normalized characteristic time of best-policy identification
(Appendix~\ref{app:fw-cost}). The theorem takes explicit inputs; the protocol menu and the set of facets can be
exponentially large in the horizon, and we give no polynomial algorithm for compact models.

\subsection{Three evidence channels in one quantity}\label{sec:fw-channels}

In Theorem~\ref{thm:fw-cost}, proxy values, audit replies, and verifier labels are transcript fields and change only
$(D,\bar k,[\theta])$. A report-adaptive audit policy is a training experiment whose replies may also change later
actions. The deployment monitor and budget change only $X$, hence $\cK_\varepsilon(X)$. The following family exhibits
all three channels.

\begin{definition}[Cue--channel family]\label{def:fw-cc}
Fix $m\ge2$ and $r\in(0,1)$, and let $\Theta=\{+1,-1\}$. In each episode a hidden bit $X\sim\mathrm{Bern}(1/2)$ and a
channel index $U\sim\mathrm{Unif}[m]$ are drawn independently, and the agent first observes $(C,U)$, where the cue
satisfies $\Prb_\theta(C=X\mid X,U)=p_\theta=(1+\theta r)/2$. It may then buy one audit $q=j\in[m]$, priced $c_a$ in
the ledger and $\kappa_a$ in utility; the reply is $Z=X$ if $j=U$ and an independent fair bit otherwise. Finally it
plays $a\in\{0,1\}$, and $Y=\one\{a=X\}$. The free proxy is $\tilde Y=\one\{a=0\}$, and the terminal verifier reveals
$Y$ at price $c_v$. Write $d_r=\KL(\mathrm{Bern}(p_+)\Vert\mathrm{Bern}(p_-))=r\log\frac{1+r}{1-r}$.
\end{definition}

\begin{proposition}[One family, three channels]\label{prop:fw-cc}
In the cue--channel family with deployment budget $B\in\{0,1\}$ and the default comparator:
\begin{enumerate}
\item[(i)] \textup{(Proxy.)} Maximizing the proxy plays $a\equiv0$ and has value $1/2$ in both worlds, whereas the
best policy without audits has value $(1+r)/2$. Every experiment that buys neither an audit nor a verification has the
same transcript law in both worlds.
\item[(ii)] \textup{(Monitor.)} Without the deployment monitor ($B=0$), $R(\Theta;X)=r/2$. With it ($B=1$) and
$\kappa_a\le(1-r)/2$, $R(\Theta;X)=0$, so $\cT_\varepsilon=0$ for every $\varepsilon\ge0$. For $B=0$ and
$\varepsilon<r/2$, $\cT_\varepsilon(\theta)=\big(\max_{e\in\cE}D_e(\theta,-\theta)/\bar k_\theta(e)\big)^{-1}$.
\item[(iii)] \textup{(Audits.)} A forced static audit, whose channel is fixed before $U$ is seen and which is bought in
every episode, has $D=d_r/m$ and charge $c_0+c_a$. The adaptive audit $j=U$ has $D=d_r$ at the same charge. For $B=0$
and $\varepsilon<r/2$, adding the adaptive audit to the menu of forced static audits divides $\cT_\varepsilon$ by
exactly $m$. Against static audits that skip payment when $U\ne j$, the factor is $(mc_0+c_a)/(c_0+c_a)$.
\item[(iv)] \textup{(Verifier.)} Every experiment of the family satisfies
$D_e(\theta,-\theta)/\bar k_\theta(e)\le d_r/(c_0+\min\{c_a,c_v\})$. The experiment that plays $a=C$ and buys
verification attains $d_r/(c_0+c_v)$, and the adaptive audit attains $d_r/(c_0+c_a)$. Hence if $c_v\le c_a$, adding
verification to the forced static menu attains the smallest characteristic cost of any menu, $(c_0+c_v)/d_r$, and the
$m$-fold advantage of adaptive audits disappears.
\end{enumerate}
\end{proposition}

\paragraph{Verification.}
Part (i) is an extreme form of the Goodhart phenomenon \citep{skalse2022,gao2023}: the proxy is free, its optimizer
is wrong in both worlds, and proxy data cannot detect this. For a general proxy, let $h$ be the free report record of an
episode (reports, proxy, own choices, audit replies), $d^\pi$ its law under $\pi$ in a world $M$, and
$\mu(h)=\E_M[Y\mid h]$, which does not depend on the policy because policies act on reports only. Let
$f(h)=\tilde Y\in[0,1]$, $\kappa(h)=\sum_t\kappa(q_t)$, $J^f(\pi)=\E^\pi[f(h)-\kappa(h)]$, $\pi_{\rm ref}$ a reference
policy with record law $d_{\rm ref}$, $\sigma^2=\operatorname{Var}_{d_{\rm ref}}(\mu-f)$, and
$\Pi_{\mathcal C}=\{\pi\in\Pi_B:\chi^2(d^\pi\Vert d_{\rm ref})\le\mathcal C\}$.

\begin{theorem}[Goodhart gap of proxy-only learning]\label{thm:fw-goodhart}
\leavevmode
\begin{enumerate}
\item[(a)] For all $\pi,\pi'\in\Pi_{\mathcal C}$,
$\big|(J(\pi)-J(\pi'))-(J^f(\pi)-J^f(\pi'))\big|\le2\sigma\sqrt{\mathcal C}$. Consequently every maximizer $\pi_f$ of
$J^f$ over $\Pi_{\mathcal C}$ satisfies $\sup_{\Pi_{\mathcal C}}J-J(\pi_f)\le2\sigma\sqrt{\mathcal C}$.
\item[(b)] Let $\mathcal C>0$ and $\sigma>0$ with $\sigma\sqrt{1+\mathcal C}\le\frac12$. There are two worlds with
identical laws of all reports and proxy values under every policy, proxy-error variance at most $\sigma^2$ under the
reference, and two policies $a,b\in\Pi_{\mathcal C}$ with $J(b)-J(a)=\pm\gamma$ in the two worlds, where
$\gamma\ge\sigma\sqrt{\mathcal C/2}$. Every learner that buys no verification and outputs $a$ or $b$ has, in one of
the worlds, expected regret relative to the better of $a$ and $b$ at least $\sigma\sqrt{\mathcal C}/(2\sqrt2)$,
however many episodes it runs.
\end{enumerate}
\end{theorem}

Part (a) uses the Cauchy--Schwarz mechanism of \citet{laidlaw2025}. In (b) the two worlds are training-equivalent for
every menu without verification, so $\varepsilon_{\rm imp}\ge\gamma/2$ on $\Delta(\{a,b\})$ and the cost of
Theorem~\ref{thm:fw-cost} is infinite below that tolerance.

With known laws, FaceTS uses verifier labels through their likelihood. When values must be estimated
(Sections~\ref{sec:theory} and~\ref{sec:deep}), the verifier enters through a doubly robust signal
\citep{robins1994,dudik2011,jiang2016}. Suppose an episode is verified with probability $p(h)\in[p_{\min},1]$ by a coin
$A$ independent of the hidden trajectory and of $Y$ given the past and $h$, that $g$ is a control variate fitted on
earlier data, and that the verifier flips $Y$ with known probability $\eta<\frac12$ (Section~\ref{sec:model}).

\begin{lemma}[Verified signal]\label{lem:fw-dr}
Let $\check Y=(Y^{\rm ver}-\eta)/(1-2\eta)$ and $\Psi=g(h)+\frac{A}{p(h)}(\check Y-g(h))$. Then
$\E[\Psi\mid\text{past},h]=\mu(h)$, and with $m_g(h)=\E[(\check Y-g(h))^2\mid\text{past},h]$,
$\operatorname{Var}(\Psi\mid\text{past})=\operatorname{Var}(\check Y\mid\text{past})
+\E\big[\tfrac{1-p(h)}{p(h)}\,m_g(h)\bigm|\text{past}\big]$.
\end{lemma}

Proposition~\ref{prop:dr} is the case $p\equiv\rho$, $\eta=0$; the lemma adds report-dependent verification and the
noisy verifier. The price of partial verification depends on the control-variate residual rather than on the proxy's
bias, and under an expected budget it is minimized by $p\propto\sqrt{m_g}$ clipped to $[p_{\min},1]$
\citep{neyman1934,zrnic2024}.

\paragraph{Audits.}
A report-adaptive audit policy is one experiment: its replies are transcript fields and its purchases enter $k_e$.
By Proposition~\ref{prop:fw-cc}(iii), adaptivity can cut the certification cost by an arbitrarily large factor, even
though every static channel separates the worlds. Parts (iii)--(iv) show that the factor depends on the comparison
menu. It shrinks to $(mc_0+c_a)/(c_0+c_a)$ against static schedules that may skip payment, and it disappears when the
menu contains a verifier no more expensive than an audit, because the verified outcome of $a=C$ reveals $X$ and every
transcript is a garbling of $(C,U,X)$. Audit comparisons must therefore price every channel. In-episode audits keep
one value that no verifier provides: at deployment they change the current action, as the monitor does in part (ii).

\paragraph{Deployment monitor.}
The monitor and budget determine $\Pi$, hence $X$. On an explicit tree they enter through the sequence form
\citep{koller1996,vonstengel1996}: a randomized perfect-recall policy is its realization plan $x$, the probability of
each sequence of own choices, with $x_\varnothing=1$ and $\sum_ax_{\sigma(I)a}=x_{\sigma(I)}$ at every information set
$I$ with own-choice prefix $\sigma(I)$, and over-budget choices removed. Values are linear, $J_\theta(x)=g_\theta^\top
x$, on a polytope $X_{\rm seq}$ with the same payoff set as $\Delta(\Pi)$ when $\Pi$ contains every deterministic
policy. Candidate sets are finite when they come from FaceTS, and polytopes when worlds are \emph{mixtures}: given a
finite set $\Xi$ of known mode models, a world is a weight vector $\theta\in\Delta(\Xi)$ and each episode first draws a
mode $\xi\sim\theta$. Then stacking the transcript probabilities of a fixed training menu gives $y=S\theta$, values are
$J_\theta(x)=\theta^\top Gx$, and unlimited training leaves the compatible polytope
$C_y=\{\theta'\in\Delta(\Xi):S\theta'=y\}$. Write $V^\star(\theta)=\max_{x\in X_{\rm seq}}J_\theta(x)$.

\begin{proposition}[Deployment certificates]\label{prop:fw-deploy}
Let $\bar V$ be the default comparator on $X_{\rm seq}$.
\begin{enumerate}
\item[(i)] For finite $C$, $R(C;X_{\rm seq})=\min\{\rho:x\in X_{\rm seq},\ \rho\ge\bar V_\nu-g_\nu^\top x\ \forall\nu\in C\}$,
a linear program of size polynomial in the tree and $|C|$.
\item[(ii)] For mixture worlds, let $C_y$ be nonempty with dimension $k$ and vertices $v_1,\dots,v_N$. Then
$R(C_y;X_{\rm seq})=\min\{\rho:x\in X_{\rm seq},\ \rho\ge V^\star(v_i)-v_i^\top Gx,\ i\le N\}$,
$N\le\binom{|\Xi|}{k}$, and
\[
R(C_y;X_{\rm seq})=\max_{\mu}\big[\E_\mu V^\star(\theta)-V^\star(\E_\mu\theta)\big],
\]
the maximum over finitely supported probability measures on $C_y$, attained by one supported on at most $k+2$ vertices.
\end{enumerate}
\end{proposition}

The right side of (ii) is the largest value of perfect information over compatible priors; with the supporting worlds
and their comparator values it certifies that no deployable policy beats the radius. With finite training data,
Theorem~\ref{thm:fw-conf} (Appendix~\ref{app:fw-deploy}) bounds the regret of the minimax policy of a Hoeffding
confidence polytope $C_n$ by $R(C_n)\le R(C_y)+\omega(2\varrho_n)$ with probability $1-\alpha$, where $\varrho_n$ is the
confidence radius and the modulus $\omega$ measures only decision-relevant contrasts between values.

\begin{corollary}[All channels in one certificate]\label{cor:fw-integration}
Consider a finite known-law instance whose menu contains any combination of proxy-only episodes, report-adaptive audit
policies, and terminal verification, and whose deployment class consists of all randomized policies of an explicit
deployment tree under the declared monitor and budget $B$. FaceTS returns a deployable policy with a valid certificate
$\Rhat\le\varepsilon$ at leading charge $\cT_\varepsilon(\theta)\log(1/\delta)$ whenever this is finite, and no sound
learner has a smaller leading charge. With probability at least $1-\delta$ its confidence sets cover the truth at all
times, so stopping at any data-dependent time yields the minimax policy of $C_n$, computed by
Proposition~\ref{prop:fw-deploy}(i), with valid radius $R(C_n;X)$ and a least-favorable witness of at most $d_X+1$
worlds. Labels used to construct the proxy are a separate setup charge.
\end{corollary}

\subsection{Consequences: hedging, memory, and monitors}\label{sec:fw-consequences}

Let $\cK_\varepsilon(\Pi)=\{S:\exists\pi\in\Pi\ \forall\nu\in S,\ \Reg_\nu(\pi)\le\varepsilon\}$ be the complex
available to learners that must output a deterministic policy, and let $\cT^{\rm det}_\varepsilon$
be~\eqref{eq:fw-cost} computed with it. The proof of Theorem~\ref{thm:fw-cost} applies verbatim to such learners, so
$\cT^{\rm det}_\varepsilon$ is their sharp constant.

\begin{proposition}[Hedging, memory, and monitors]\label{prop:fw-consequences}
\leavevmode
\begin{enumerate}
\item[(i)] \textup{(Hedging.)} $\cK_\varepsilon(\Pi)\subseteq\cK_\varepsilon(X)$, with equality at $\varepsilon=0$.
For each $\varepsilon\in(0,1/2]$ there is a two-world, two-action family with $\cT_\varepsilon=0$ in which
$\cT^{\rm det}_\varepsilon$ equals any prescribed positive value, or $+\infty$.
\item[(ii)] \textup{(Memory and monitors.)} If $X\subseteq X'$ share the comparator, then
$\cK_\varepsilon(X)\subseteq\cK_\varepsilon(X')$, and $\cT_\varepsilon(\theta)$, $\varepsilon_{\rm free}$, and
$\varepsilon_{\rm imp}(\theta)$ are no larger under $X'$. Moreover $\cT_\varepsilon(\theta;X)=\cT_\varepsilon(\theta;X')$
for every finite menu and every $\theta$ if and only if $\cK_\varepsilon(X)=\cK_\varepsilon(X')$.
\item[(iii)] \textup{(Verifiers learn, monitors act.)} $\cK_\varepsilon(X)$ depends only on the deployment class and
the comparator, and $(D,\bar k,[\theta])$ only on the training menu. If $P^e_\nu=Q\circ P^{e'}_\nu$ for a
$\nu$-independent Markov kernel $Q$ and $\bar k_\nu(e')\le\bar k_\nu(e)$ for every $\nu$, then deleting $e$ from a menu
that contains $e'$ leaves every $\Gamma_\theta(A)$ unchanged. If, for every $\pi\in\Pi$, the menu contains an
experiment that executes $\pi$ and buys verification, then with the default comparator $\varepsilon_{\rm imp}(\theta)=0$
and $\cT_\varepsilon(\theta)<\infty$ for every $\varepsilon\ge0$.
\item[(iv)] \textup{(Monitors can also learn faster.)} For every $\rho\in(0,1)$ there is a two-world family in which
an in-episode monitor and a terminal outcome verifier have equal price, both separate the worlds, and the monitor's
per-episode information exceeds the verifier's by a factor of at least $\rho^{-2}$.
\end{enumerate}
\end{proposition}

By (i), a learner that must output a deterministic policy, as in best-policy identification, can pay for verification
that hedged deployment makes unnecessary: never at exact optimality, but possibly without bound at positive tolerance.
Part (ii) is an exact criterion for when deployment memory or a monitor substitutes for training evidence, and it is
weaker than preserving every payoff vector. By (iii), a verifier changes what training can learn and at what price,
while a deployment monitor changes what can be done with what was learned. When a verifier is Blackwell-sufficient
\citep{blackwell1953} at no greater price, as in Proposition~\ref{prop:fw-cc}(iv), an in-episode audit has only
deployment value; by (iv), the monitor can also be the more informative training channel. Exact computations for the
examples are in Appendix~\ref{app:fw-checks}.

Section~\ref{sec:class} drops the assumption of known candidate laws. In a structured class, which early reports
predict which hidden events is unknown and must be learned from purchased audits and verifications, and the certified
charge depends on this structure only through the logarithm of the number of candidate cue sets.
\section{A sparse hidden-event class}\label{sec:class}

\begin{definition}[Sparse hidden-event class $\cM(L,k,s,\gamma)$]\label{def:class}
Reports $X=(X_1,\dots,X_L)$ are i.i.d.\ uniform on $\{-1,+1\}$ and arrive one per step. For $S\subseteq[L]$ let
$\chi_S(X)=\prod_{\ell\in S}X_\ell$. There are $k$ hidden events
\[
E_j=\chi_{S_j}(X)\,N_j,\qquad j\in[k],
\]
with unknown pairwise disjoint supports $1\le|S_j|\le s$ and independent signs $N_j$, independent of $X$, with
unknown means $\mu_j=\E N_j\in[\gamma,1]$. After the reports, the agent may buy audits $A\subseteq[k]$ (audit
$j$ reveals $E_j$; adaptive order allowed), then plays $a\in\{-1,+1\}$. The task outcome is
\[
Y=\one\Big\{a=\sigma\prod_{j=1}^kE_j\Big\}
\]
with unknown polarity $\sigma\in\{\pm1\}$. Deployment audits cost $\kappa_j\ge0$ in utility; the learner
knows $L$, $k$, and $s$, but not $\gamma$, the supports, the reliabilities, or $\sigma$.
\end{definition}
The class models a decision that depends on several hidden facts, each predictable from a different early cue: a
door is safe if two independent sensors agree, a patch is correct if all of its preconditions hold. For
$A\subseteq[k]$ write $T_A=\triangle_{j\notin A}S_j$ for the symmetric
difference (a disjoint union here) of the supports of the unaudited events, and
\[
v(A)=\tfrac12\Big(1+\prod_{j\notin A}\mu_j\Big)-\sum_{j\in A}\kappa_j .
\]

\begin{lemma}[Oracle structure and quotients]\label{lem:oracle}
In $\cM(L,k,s,\gamma)$ with deployment budget $B$:
\begin{enumerate}
\item[(i)] $V^\star_B=\max_{|A|\le B}v(A)$. It is attained by the controller that buys $A$ and plays
$a=\sigma\,\chi_{T_A}(X)\prod_{j\in A}E_j$; adaptive auditing cannot do better. The controller's action changes sign
whenever one audited reply is counterfactually flipped with everything else fixed.
\item[(ii)] A \emph{reply-blind} policy, whose audits and actions are measurable with respect to the reports and
internal randomness, has $J\le V^\star_0=\frac12(1+\prod_j\mu_j)$. Hence $J(\pi)>V^\star_0$ implies that
$\pi$ uses audit replies, and no training procedure can lift a policy in $\Pi_0$ above $V^\star_0$.
\item[(iii)] A \emph{report-blind} policy, measurable with respect to replies and internal randomness, has
$J\le\frac12$ if $B<k$, and $J\le\max\{\frac12,1-\sum_j\kappa_j\}$ if $B\ge k$.
\item[(iv)] Fix an optimal audit set $A^\star$. Before any audit, an optimal policy needs only the report summary
$\chi_{T_{A^\star}}(X)$: one bit, maintained by the recursive register $b_t=b_{t-1}X_t^{\one\{t\in T_{A^\star}\}}$, given the
fixed set $T_{A^\star}$ and the report index $t$, which takes two values if
$T_{A^\star}\ne\varnothing$ and is constant otherwise; after the audits the action combines it with the replies.
This summary has the minimum number of values among all report summaries on which an optimal policy can act: one
value if some optimal policy ignores the reports, and two otherwise. Optimal summaries need not be unique, since
different optimal audit sets may call for different parities. The belief abstraction (the posterior law of
$(E_1,\dots,E_k)$ given $X$) has $2^k$ classes, and the raw report history has $2^L$.
\end{enumerate}
\end{lemma}
Part (iv) compares \emph{cardinalities} of sufficient statistics: two values ($1$ bit) for the decision summary, $2^k$
values ($k$ bits) for the belief, and $2^L$ values ($L$ bits) for the history. The count for the register excludes the
learned fixed parameters and the externally supplied clock.
Algorithm~\ref{alg:dsl} recovers every cue set $S_j$ (and every $\mu_j$ when $B\ge1$) on the way to the register, for
reasons given in \S\ref{sec:theory}. DORL-PPO is \emph{trained} to predict every event, and \S\ref{sec:rev} measures what
its recurrent state encodes. By part (i), buying event $j$ replaces the factor $\mu_j$ by $1$, so cheap audits on
unreliable events are the most valuable.
\section{Certified learning of a decision state}\label{sec:theory}

\subsection{An anytime audit-supervised learner}

Let $\cS_s=\{S\subseteq[L]:1\le|S|\le\min(s,L)\}$ and $M_s=|\cS_s|$, so that $\log M_s\le s\log(eL/s)$ when
$s\le L$. Algorithm~\ref{alg:dsl} checks its statistics on a geometric schedule $n_i=\lceil n_0r^i\rceil$ ($r>1$)
with error budget $\delta_i=\delta/(2(i+1)(i+2))$ and radius
\[
t_i=\sqrt{2\log(2kM_s/\delta_i)/n_i}.
\]
Characters are orthonormal under the uniform law, so $\E[E_j\chi_S(X)]=\mu_j\one\{S=S_j\}$:
an audit reply correlates with exactly one candidate cue set, with strength $\mu_j$. Since every false candidate has
mean zero, a lead of more than $2t_i$ over the runner-up and over a zero-valued dummy competitor certifies the
selection without knowledge of $\gamma$. Once every event is audited, a single exact verification reveals the polarity,
because $(2Y-1)\,a\prod_jE_j=\sigma$.

\begin{algorithm}[t]
\caption{Audit-supervised decision-state learner (DSL)}\label{alg:dsl}
\KwIn{$L,k$; public support bound $s$; deployment budget $B$ and prices $\kappa$; target $\varepsilon\ge0$, with $\varepsilon>0$ if $B\ge1$; confidence $\delta$; verifier flip bound $\eta<\frac12$}
\For{$i=0,1,2,\dots$}{
  run episodes up to $n_i$ in total; in each, \emph{buy all $k$ audits}, play any action, buy no verification\;
  $\hat c_j(S)\gets n_i^{-1}\sum_{e\le n_i}E_{e,j}\chi_S(X_e)$ for all $j\in[k]$, $S\in\cS_s$\;
  $\hat S_j\gets\argmax_S\hat c_j(S)$, $\hat\mu_j\gets\min\{1,\max\{0,\hat c_j(\hat S_j)\}\}$, $\Rhat\gets k\,t_i\one\{B\ge1\}$\;
  \lIf{$\hat c_j(\hat S_j)-\max\{0,\max_{S\ne\hat S_j}\hat c_j(S)\}>2t_i$ for all $j$, and $\Rhat\le\varepsilon$}{\textbf{break}}
}
run $m=1$ (if $\eta=0$) or $m=\lceil2\log(2/\delta)/(1-2\eta)^2\rceil$ episodes buying all audits and a verification, with $a=+1$\;
$\hat\sigma\gets\sgn\sum_{e\le m}(2Y^{\rm ver}_e-1)\,a_e\prod_jE_{e,j}$\;
$\hat A\gets\argmax_{|A|\le B}\hat v(A)$ with $\hat v(A)=\frac12(1+\prod_{j\notin A}\hat\mu_j)-\sum_{j\in A}\kappa_j$;\quad
$\hat T\gets\triangle_{j\notin\hat A}\hat S_j$\;
\KwOut{controller $\pihat$: keep the bit $b_t=\prod_{\ell\le t,\ \ell\in\hat T}X_\ell$; buy $\hat A$; play $a=\hat\sigma\, b_L\prod_{j\in\hat A}E_j$; certificate $\Rhat$}
\end{algorithm}

\begin{theorem}[Certified decision state from audit supervision]\label{thm:upper}
Fix any world in $\cM(L,k,s,\gamma)$ and let $\mu_{\min}=\min_j\mu_j$. Set $\theta=\mu_{\min}/4$ if $B=0$ and
$\theta=\min\{\mu_{\min}/4,\varepsilon/k\}$ if $B\ge1$, and let $i^\star$ be the first check with $t_{i^\star}<\theta$ (a
deterministic index). With probability at least $1-\delta$:
\begin{enumerate}
\item[(a)] \textbf{(anytime soundness)} at every check, every support that passes its lead test is correct and
$|\hat\mu_j-\mu_j|\le t_i$; at stopping $\hat\sigma=\sigma$ and $\Reg(\pihat)\le\Rhat\le\varepsilon$. If $B=0$
the controller is exactly optimal and $\Rhat=0$;
\item[(b)] \textbf{(charge)} the learner stops no later than check $i^\star$. Either $i^\star=0$ and
$n_\tau\le\lceil n_0\rceil$, or
\[
n_\tau\le n_{i^\star}\le 1+\frac{2r}{\theta^2}\log\frac{4kM_s\,i^\star(i^\star+1)}{\delta},
\qquad i^\star\le\log_r\frac{n_{i^\star}}{n_0},
\]
so that $n_\tau=O\big(n_0+\theta^{-2}[\log M_s+\log k+\log(1/\delta)+\log\log(e^2/\theta)]\big)$ audit episodes, and
the charge is $C=(n_\tau+m)(c_0+\sum_jc_j)+m\,c_v$;
\item[(c)] \textbf{(decision state)} the deployed controller stores one bit and the audit counter; it buys the
estimated best audit set, and its action changes sign whenever an audited reply is counterfactually flipped.
\end{enumerate}
\textbf{(d) Expected charge.} Let $i^\dagger\ge i^\star$ be the first check after which every radius stays below $\theta$,
which exists because $t_i\to0$. Then $i^\dagger=i^\star$ whenever the radii are nonincreasing from $i^\star$ on; this holds for
every configuration in our experiments (checked numerically for $n_0=64$, $r=1.25$). Without conditioning on any event,
$\E[n_\tau]\le(1+C_r\delta)\,n_{i^\dagger}$ for $\delta\le\frac12$, with a constant $C_r$ that depends only on $r$. Since
$t_{i^\dagger-1}\ge\theta$ when $i^\dagger\ge1$, $n_{i^\dagger}$ satisfies the bound in (b) with $i^\dagger$ in place of $i^\star$, and so has the same
$O(\cdot)$ rate. Hence
$\E[C]\le\big((1+C_r\delta)n_{i^\dagger}+m\big)(c_0+\sum_jc_j)+mc_v$.

\emph{Audit-free route.} If $B=0$, consider the variant that buys only verifications and runs the same lead test on
$|\hat c(T)|$, with $\hat c(T)=n^{-1}\sum_e(2Y^{\rm ver}_e-1)a_e\chi_T(X_e)$, over $T\in\cS_{d_{\max}}$ with
$d_{\max}=\min(ks,L)$. It is $\delta$-sound with $\Rhat=0$. With $\lambda=(1-2\eta)\prod_j\mu_j$, it stops after
$O\big(n_0+\lambda^{-2}[\log|\cS_{d_{\max}}|+\log(1/\delta)+\log\log(e^2/\lambda)]\big)$ episodes, each charged
$c_0+c_v$, and the same expected-charge statement holds. Here $\log|\cS_{d_{\max}}|\le d_{\max}\log(eL/d_{\max})$.
\end{theorem}
The informative parity for the audit-free route is $T^\star=\bigcup_jS_j$, of degree $d=\sum_j|S_j|\le ks$, with
equality when every support has size $s$, as in our experiments. The two inequalities in (b) jointly bound
$i^\star=O(\log_r(1/\theta)+\log\log(kM_s/\delta))$. The constants are conservative: in \S\ref{sec:results} the learner
stops near the prediction with a stable proportionality factor.
\subsection{An information-per-price lower bound}

For the lower bound we reveal every parameter except the support of one event. For a training design $D=(A,v)$, an
audit set $A$ and a verification flag $v$, define its information-to-price ratio for identifying the support of
event 1 and the best such ratio,
\[
\iota(D)=\frac{\mu_1^2\,\one\{1\in A\}+\big(\prod_{j\notin A}\mu_j\big)^2\one\{1\notin A,\ v=1\}}{c_0+\sum_{j\in A}c_j+v\,c_v},
\qquad
\bar\rho=\max_D\iota(D).
\]

\begin{theorem}[Information-per-price lower bound, singleton-support subclass]\label{thm:lower}
Consider the subclass of $\cM(L,k,1,\gamma)$ with $B=0$, known $\mu\in(0,1)^k$, known $\sigma$, known singleton
supports $S_2,\dots,S_k$, an uninformative proxy, an exact verifier, and $S_1=\{J\}$ with $J$ unknown in
$\cC=[L]\setminus\bigcup_{j\ge2}S_j$, $|\cC|=L'$. Let a learner adaptively choose audits and verifications, stop
with finite expected charge, and, for every $J\in\cC$, output $\pihat$ with
$\Prb_J\{\Reg_J(\pihat)\le\frac18\prod_j\mu_j\}\ge\frac34$. Then
\[
\frac1{L'}\sum_{J\in\cC}\E_J[C]\ \ge\ \frac{\frac34\log L'-\log2}{\bar\rho}.
\]
The bound holds even if every audit reply and the verifier are noiseless and even if the learner is told every
parameter except $J$.
\end{theorem}

The proof (Appendix~\ref{app:lower}) compares every world with a reference in which event 1 is a fair coin
independent of the reports. An audit of event 1 then yields at most $\mu_1^2$ nats, a verification without that
audit at most $(\prod_{j\notin A}\mu_j)^2$, and everything else nothing. Information per unit of charge is therefore
at most $\bar\rho$ in every episode, even with adaptive designs and a random stopping time. A near-optimal policy
reveals $J$ through its unique large Fourier coefficient (Bessel's inequality), and Fano's inequality completes the
argument.

\begin{corollary}[Exponential separation and crossover price]\label{cor:separation}
Consider $s=1$, $B=0$, $\mu_j=\gamma<1$ for all $j$, all audits priced $c_a$, an exact verifier, $k\le L$, $0<\delta\le\frac14$,
the regret target $\varepsilon=\frac18\gamma^k$ of Theorem~\ref{thm:lower}, and
$L'=L-k+1\ge3$. Write
$H_a=\log(kL/\delta)+\log\log(e^2/\gamma)$ and $H_o=k\log(eL/k)+\log(1/\delta)+\log\log(e^2/\gamma^k)$; all charges below
are expected charges of $\delta$-sound learners that meet the accuracy requirement of Theorem~\ref{thm:lower}.
\begin{enumerate}
\item[(i)] \textbf{Outcome only.} Every learner that buys no audits has average expected charge at least
$(c_0+c_v)\gamma^{-2k}(\frac34\log L'-\log2)$, and the audit-free route of Theorem~\ref{thm:upper} attains
$O\big((c_0+c_v)\gamma^{-2k}H_o\big)$.
\item[(ii)] \textbf{Audit supervision.} Algorithm~\ref{alg:dsl} certifies the optimal policy with expected charge
$O\big((c_0+kc_a)\gamma^{-2}H_a+c_v\big)$. If $c_a\le(c_0+c_v)\gamma^{-2(k-1)}-c_0$, every learner, with any mix of audits
and verifications, needs average expected charge at least $(c_0+c_a)\gamma^{-2}(\frac34\log L'-\log2)$. So when the
verification term is dominated, $c_v\le(c_0+kc_a)\gamma^{-2}H_a$, Algorithm~\ref{alg:dsl} is optimal up to the factor
$(c_0+kc_a)/(c_0+c_a)\le k$ and the ratio $H_a/\log L'$.
\item[(iii)] \textbf{Separation.} The ratio of the best audit-free charge to Algorithm~\ref{alg:dsl}'s charge is at least
\[
c\,\log L'\cdot\min\Big\{\frac{c_0+c_v}{(c_0+kc_a)H_a}\,\gamma^{-2(k-1)},\ \gamma^{-2k}\Big\}
\]
for an absolute constant $c>0$. Both terms are exponential in $k$ when the prices and $H_a$ grow at most polynomially
in $k$. The second term governs when verification is
so expensive that even the audit route's single verification dominates its charge. Comparing the two upper bounds
(not the unknown optimal charges): Algorithm~\ref{alg:dsl} buys all $k$ audits per episode, and up to constants it
is the cheaper of the two routes when
$k c_a\lesssim(c_0+c_v)\gamma^{-2(k-1)}H_o/H_a-c_0$. Its per-audit crossover price is therefore of order
$(c_0+c_v)\gamma^{-2(k-1)}H_o/(kH_a)$. The ratio $H_o/H_a$, which can grow like $k$, must be kept. The
information-design threshold in (ii) concerns a single event and carries neither factor.
\end{enumerate}
\end{corollary}

\begin{remark}[The accuracy target]\label{rem:accuracy}
The separation concerns exact identification at the regret target $\varepsilon=\frac18\gamma^k$, a fixed fraction of
the advantage $V^\star_0-\frac12=\frac12\gamma^k$ that memory confers. At a fixed absolute accuracy there is no
separation for large $k$: a uniformly random action has regret $\frac12\gamma^k$, which falls below any fixed
$\varepsilon>0$.
\end{remark}

\begin{remark}[Recovering the decision state]\label{rem:existence}
A two-valued report summary suffices (Lemma~\ref{lem:oracle}(iv)), but Algorithm~\ref{alg:dsl} reaches it by
estimating every cue set $S_j$ (and, for $B\ge1$, every $\mu_j$). The detour through events pays: the only audit
statistic that targets $\chi_{T^\star}$ directly, the product $\prod_jE_j$, correlates with it at the weak
outcome-level strength $\prod_j\mu_j$, while by Theorem~\ref{thm:lower} locating even one cue costs
$\Omega(\log L'/\bar\rho)$ with $L'=L-k+1$. Hence the near-optimality in Corollary~\ref{cor:separation}(ii); with
expensive audits the event-level route can be arbitrarily worse than outcome-only learning, as the crossover price
quantifies. Linear probes in \S\ref{sec:rev} measure what DORL-PPO's recurrent state encodes.
\end{remark}

\begin{remark}[Scope of the separation]
The separation compares information channels at matched access, since Theorem~\ref{thm:lower} lets a learner mix
audits and verifications freely. A Bayes-optimal learner of the same POMDP with the same channels could reproduce
Algorithm~\ref{alg:dsl}. The result is consistent with \citet{jia2025}, whose process rewards are functions of the
observed trajectory; here, when $\mu_j<1$, the audited event $E_j$ is not a function of the report history. In every
case the action-corrected outcome $Z=(2Y-1)a=\sigma\prod_jE_j$ has exactly one nonzero Fourier coefficient on the
reports, $\E[Z\chi_T(X)]=\sigma\prod_j\mu_j\,\one\{T=T^\star\}$, a parity of degree $d=\sum_j|S_j|$.
\end{remark}

\begin{remark}[Computation]
Algorithm~\ref{alg:dsl} evaluates $kM_s$ correlations per episode, whereas the audit-free route scans
$|\cS_{d_{\max}}|\ge\binom{L}{d_{\max}}$ candidate parities. Within the statistical-query model, the SQ-dimension bound
for parity classes \citep{blum1994,kearns1998} implies that learning a parity of degree $d$ over $L$ bits needs
either $\binom{L}{d}^{\Omega(1)}$ queries or tolerance $\binom{L}{d}^{-\Omega(1)}$, in line with the learnability gains
from intermediate supervision in sequence models \citep{wies2023}.
\end{remark}
\section{From certified learning to deep RL}\label{sec:deep}

Algorithm~\ref{alg:dsl} knows the hypothesis class, whereas a deep agent must discover the relevant cues inside a
recurrent state trained by stochastic gradients. The next proposition compares the gradient signal of audit and
outcome supervision at initialization, for linear logits on a \emph{given} family of parity features; \S\ref{sec:rev}
measures what trained recurrent networks encode.

\begin{proposition}[Signal of audit supervision versus outcome supervision]\label{prop:gradient}
In $\cM(L,k,s,\gamma)$ with $B=0$, let $\Phi(X)=(\chi_S(X))_{S\in\cG}$ be a feature vector over a family
$\cG$ of report subsets, and consider linear logits initialized at $w=0$.
\begin{enumerate}
\item[(a)] For the audit-reply log-likelihood $\ell_j(w)=\log\Pr_w(E_j\mid X)$ with
$\Pr_w(E_j=e\mid X)=(1+e^{-e\,w^\top\Phi})^{-1}$: $\nabla\E\ell_j(0)=\tfrac12\mu_j\,\mathbf e_{S_j}$ if $S_j\in\cG$, and $0$ otherwise.
\item[(b)] For the policy $\pi_w(a\mid X)=(1+e^{-a\,w^\top\Phi})^{-1}$ and $J(w)=\Prb_w(Y=1)$:
$\nabla J(0)=\tfrac14\sigma\prod_j\mu_j\,\mathbf e_{T^\star}$ with $T^\star=\bigcup_jS_j$, $|T^\star|=\sum_j|S_j|$, if
$T^\star\in\cG$, and $0$ otherwise.
\item[(c)] The single-sample gradients (for (b), REINFORCE with any action-independent baseline in $[0,1]$) have every coordinate in
$[-\frac12,\frac12]$. Consequently, the sign-consistent largest coordinate of a minibatch gradient identifies the
informative feature with probability at least $1-\delta$ once the batch has
$8\log(2|\cG|/\delta)/\mu_j^2$ samples in case (a), and $32\log(2|\cG|/\delta)/(\prod_j\mu_j)^2$ samples in case (b).
\end{enumerate}
\end{proposition}
Outcome supervision thus needs a feature of degree $d=\sum_j|S_j|$ ($=ks$ when every support has size $s$) and has
signal $\prod_j\mu_j$; audit supervision needs features of degree at most $s$ and has signal $\mu_j$. This local
analogue of the leap-complexity barrier for SGD on parities \citep{abbe2023} is our reason for using purchased audit
replies in deep RL as \emph{auxiliary targets for the memory} rather than only as inputs.

\subsection{DR and DORL: what changes}\label{sec:dr-vs-dorl}

Our strongest outcome-supervised learner, called \emph{DR} throughout, is recurrent PPO whose terminal reward is the
doubly-robust verified return of Proposition~\ref{prop:dr} below, an unbiased estimate of the task outcome, so DR
does not optimize the proxy. DORL spends part of DR's verification budget on \emph{side audits},
which reveal all hidden events of an episode after it ends; their replies supervise the recurrent state through
event-belief heads at every step (Figure~\ref{fig:drdorl}a). At equal nominal charge, a verified episode yields one
label whose informative correlation with the reports has degree $\sum_j|S_j|$ and size $\prod_j\mu_j$, and an audited
episode yields $k$ labels with degree at most $s$ and size $\mu_j$. Proposition~\ref{prop:gradient} and
Corollary~\ref{cor:separation} translate this into gradient signal and certification charge. In
Figure~\ref{fig:drdorl}b the certified outcome-only route needs exponentially more episodes as $k$ grows, and DR-PPO
stays at the memory-free value for every $k\ge2$, while DORL-PPO learns the memory. When the policy may also buy audits
at deployment ($B\ge1$), the same replies enter the policy as inputs and can change the action, the only way to exceed
the reply-blind value $V^\star_0$ (Lemma~\ref{lem:oracle}).

\subsection{DORL-PPO}

\begin{figure}[t]
\centering
\includegraphics[width=\linewidth]{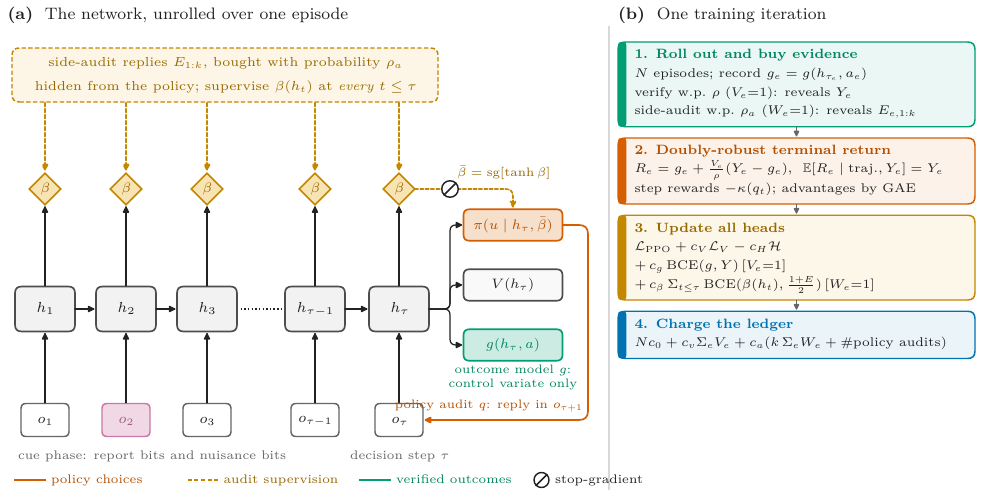}
\caption{\textbf{DORL-PPO.} \textbf{(a)} The network unrolled over one episode. Side-audit replies supervise the
event-belief heads $\beta(h_t)$ on the shared GRU state at every step $t\le\tau$ of a side-audited episode; at the
decision step $\tau$, $h_\tau$ feeds the policy, value, and outcome-model heads. The policy reads the beliefs only
through a stop-gradient, $\bar\beta=\mathrm{sg}[\tanh\beta]$, chooses $u=(q,a)$, and receives the reply to a policy
audit in the next observation. \textbf{(b)} One training iteration (Algorithm~\ref{alg:dorlppo}). The outcome model $g$
enters the return only as a control variate.}\label{fig:arch}
\end{figure}

DORL-PPO (Algorithm~\ref{alg:dorlppo}, Figure~\ref{fig:arch}) is a recurrent actor-critic over the joint space of physical actions and
audits. Its encoder and GRU state $h_t$ are shared by five heads: the policy $\pi(u\mid h_t)$, the value $V(h_t)$,
an outcome model $g(h_t,a)\approx\Prb(Y=1\mid h_t,a)$, event-belief logits $\beta_j(h_t)$ for the $k$ auditable
events, and, for one baseline only, a next-observation predictor. Each $\beta_j$ is a Bernoulli logit for
$\Prb(E_j=+1\mid h_t)$, trained by binary cross-entropy on $(1+E_j)/2$. The policy head also reads the detached
features $\tanh\beta_j(h_t)$ (the signed mean $\tanh(\beta_j/2)$ would carry the same information). Every training
episode buys a verification independently with probability $\rho$ and a full set of \emph{side audits} with
probability $\rho_a$. Side-audit replies supervise $\beta$ and are never shown to the policy. In-episode audits chosen
by the policy are ordinary actions that pay $\kappa$ in the return and $c_a$ in the charge.

\begin{algorithm}[t]
\caption{DORL-PPO (one iteration)}\label{alg:dorlppo}
roll out $N$ episodes of the current policy; record $g_e=g(h_{\tau_e},a_e)$ at each decision time $\tau_e$ before any update\;
draw $V_e\sim\mathrm{Bern}(\rho)$ (buy verification) and $W_e\sim\mathrm{Bern}(\rho_a)$ (buy side audits) independently\;
terminal return $R_e\gets g_e+\frac{V_e}{\rho}(Y_e-g_e)$, computed without gradient (detached); step rewards $-\kappa(q_t)$; advantages by GAE ($\gamma=1$)\;
\For{$K$ epochs over minibatches of whole episodes}{
  $\mathcal L\gets\mathcal L_{\rm PPO}+c_V\mathcal L_V-c_H\mathcal H
   +c_g\,\mathrm{BCE}\big(g(h_{\tau_e},a_e),Y_e\big)_{V_e=1}
   +c_\beta\,\mathrm{BCE}\big(\beta(h_t),\tfrac{1+E_e}{2}\big)_{W_e=1,\ t\le\tau_e}$\;
  gradient step with norm clipping\;
}
charge $N c_0+c_v\sum_eV_e+c_a(k\sum_eW_e+\#\text{policy audits})$\;
\end{algorithm}

\begin{proposition}[Unbiased verified return]\label{prop:dr}
Suppose $V\sim\mathrm{Bern}(\rho)$ is drawn independently of the episode, and $g$ is computed from parameters
fitted on earlier data and from the current trajectory. Then
$\E[R\mid\text{trajectory},Y]=Y$ and $\mathrm{Var}(R\mid\text{trajectory},Y)=(\rho^{-1}-1)(Y-g)^2$.
Hence the on-policy score-function (REINFORCE) estimator with return $R-\sum_t\kappa(q_t)$ and an
action-independent baseline is unbiased for the gradient of the true objective $\E Y-\E\sum_t\kappa(q_t)$, and $g$
acts only as a control variate. The plug-in return $R=g$ is biased by $\E[g-Y\mid\text{trajectory}]$, which a policy
optimizer can exploit.
\end{proposition}
Proposition~\ref{prop:dr} concerns the terminal reward only. PPO's clipped surrogate, repeated epochs, and
bootstrapped GAE are biased even with exact rewards, and the outcome-model, event-prediction, and value losses also
update the shared encoder, so the full training procedure is not an unbiased optimizer of the deployment objective.
All learners share the architecture and PPO machinery and differ in the evidence they buy, the terminal return they
use, and the auxiliary losses and policy inputs listed in Table~\ref{tab:methods}.

\begin{figure}[t]
\centering
\includegraphics[width=\linewidth]{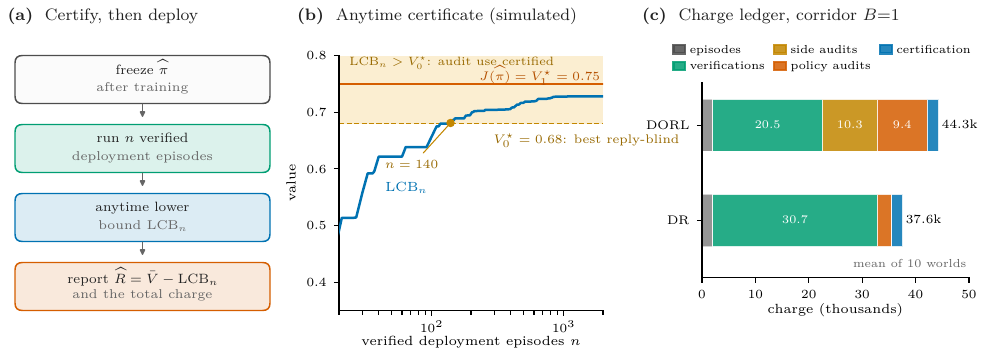}
\caption{\textbf{Certify, then deploy.} \textbf{(a)} After training, the policy is frozen and run on $n$ verified
deployment episodes, whose utilities feed the anytime lower confidence sequence of Proposition~\ref{prop:cert}; the
learner reports $\Rhat=\bar V-\mathrm{LCB}_n$ with the total charge. \textbf{(b)} The certificate on $2{,}000$ simulated
utilities of a policy that buys one audit ($\kappa=0.05$) and attains $V^\star_1=0.75$ in the corridor $B{=}1$ task.
Once $\mathrm{LCB}_n$ exceeds the reply-blind optimum $V^\star_0=0.68$, the
certificate also shows that the policy acts on its audit replies. \textbf{(c)} Mean charge ledgers of DORL-PPO and DR
in the corridor $B{=}1$ task (ten worlds each, Part~B). Verifications and side audits are matched in nominal charge;
certification and the policy audits bought during training are charged on top.}\label{fig:cert-pipeline}
\end{figure}

\begin{proposition}[Anytime value and audit-use certificates]\label{prop:cert}
Let $\pihat$ be fixed before certification, and let $U_1,U_2,\dots\in[\ell,1]$ be the utilities of independent
verified deployment episodes, with $\ell=-\kappa B$. Rescale to $x_i=(U_i-\ell)/(1-\ell)\in[0,1]$, with mean $m$, and run the
one-sided predictable plug-in empirical-Bernstein (PrPl-EB) confidence sequence of \citet[Sec.~3.2]{waudbysmith2024},
whose validity rests on the empirical-Bernstein supermartingale of \citet[Thm.~4]{howard2021}. Its exact tuning,
centring, and lower-bound formula are given in Appendix~\ref{app:cert}. Let $\mathrm{LCB}_n=\ell+(1-\ell)\max_{i\le n}L_i$ be the running
maximum of its lower bounds, mapped back to the utility scale. It satisfies
$\Prb\{J(\pihat)\ge\mathrm{LCB}_n\ \forall n\}\ge1-\alpha$. Then
\begin{enumerate}
\item[(i)] the bound remains valid at any data-dependent stopping time;
\item[(ii)] $\Rhat=\bar V-\mathrm{LCB}_n$ is an $\alpha$-sound regret certificate for any known upper bound
$\bar V\ge V^\star_B$, and $\bar V=V^\star_B$ in our synthetic tasks;
\item[(iii)] in $\cM(L,k,s,\gamma)$, the event $\{\mathrm{LCB}_n>V^\star_0\}$ certifies at level $\alpha$ that
$\pihat$ changes its actions in response to audit replies (Lemma~\ref{lem:oracle}(ii)).
\end{enumerate}
\end{proposition}
Proposition~\ref{prop:cert} certifies the trained policy (Figure~\ref{fig:cert-pipeline}) from purchased labels, which
count as certification charge; as in Definition~\ref{def:dorl}, a failed run shows up as a large certified radius.
Because the reported bound is a running maximum, $J\ge\mathrm{LCB}_n$ for all $n\le N$ holds exactly when
$J\ge\mathrm{LCB}_N$, so empirical anytime and terminal coverage of the reported sequence coincide. The running maximum
is valid because the underlying sequence $(L_i)$ is time-uniform, which Appendix~\ref{app:cert} tests by simulation.

\section{Experiments}\label{sec:experiments}

We ran five studies (Table~\ref{tab:overview}). Part A runs the certified learners of Section~\ref{sec:theory} on random
worlds and checks the scaling of Theorem~\ref{thm:upper}. Part B compares DORL-PPO with seven deep RL learners at matched
evidence charge, and Part C adds controls that isolate why DORL wins. Part D repeats the comparison with a language-model
policy. Part E studies the verification channel alone, in language-model fine-tuning with an exploitable reward model.
Parts A and B were preregistered: we fixed their hypotheses and analysis plan before analysing any deep RL result (Appendix~\ref{app:repro}). We designed Part C after seeing Part B, in response to a review. Parts D and E followed their
own pre-specified protocols.

\begin{table}[t]
\centering\small
\begin{tabular}{l>{\raggedright\arraybackslash}p{0.38\linewidth}>{\raggedright\arraybackslash}p{0.30\linewidth}r}
\toprule
part & question & setting & runs \\
\midrule
A & Do certified learners follow the predicted charge scaling? & $\cM(L,k,1,\mu)$, $2{,}200$ random worlds & $3{,}800$ \\
B & Does audit supervision let PPO learn a memory that outcome supervision misses? & corridor and grid, $7$ tasks, $8$ learners & $560$ \\
C & Which ingredient of DORL causes the gain? & $5$ controls in $8$ cells & $300$ \\
D & Does the gain hold for a language-model policy? & Qwen2.5 on rendered sensor logs & $212$ \\
E & Does sparse verification stop proxy hacking? & Qwen2.5 arithmetic, biased reward model & $39$ \\
\bottomrule
\end{tabular}
\caption{The five studies.}\label{tab:overview}
\end{table}

All studies share four conventions. Every world is synthetic or semi-synthetic, so we compute the optimal value
$V^\star_B$ and the deployed value $J(\pihat)$ exactly (or to Monte Carlo error below $0.002$) and report true regret.
Every episode, audit, and verification is charged by~\eqref{eq:charge} at $c_0=0.01$, $c_a=0.1$, and $c_v=1$; we report
training charge and certification charge separately. A seed fixes a world, and every learner runs on the same worlds,
so comparisons are paired. Baselines receive the same nominal evidence charge as DORL and spend it on verified outcomes.

\subsection{Certified learners on random worlds (Part A)}\label{sec:design-A}

We ran both certified routes of Theorem~\ref{thm:upper} with $B=0$: the audit route (Algorithm~\ref{alg:dsl}) and the
audit-free route, which buys only verifications. The grid covers $(L,k,\mu)\in\{16,32\}\times\{1,2,3,4\}\times\{0.6,0.8\}$
with $100$ worlds per cell; each world has $s=1$, random disjoint supports, equal reliabilities, and a random polarity.
We used $\delta=0.05$ and the schedule $n_0=64$, $r=1.25$. A second grid runs Algorithm~\ref{alg:dsl} with deployment
audits ($B\in\{1,2\}$, $\kappa=0.05$, $\varepsilon=0.1$, heterogeneous reliabilities, verifier flip rate
$\eta\in\{0,0.1\}$) on $200$ worlds per cell. Lemma~\ref{lem:oracle} gives exact values. We predicted that every cell
would have coverage at least $1-\delta$; that regressing $\log(n/\log(2\,\mathrm{count}/\delta))$ on
$\log(\mathrm{signal}^{-2})$ would give slope $1$ for both routes, with signal $\mu$ for audits and $\mu^k$ for outcomes;
and that the outcome-to-audit charge ratio and the crossover audit price would grow with $k$.

\subsection{Deep reinforcement learning (Parts B and C)}\label{sec:design-B}

\paragraph{Environments.}
Both environments run vectorized on a GPU and fix one world per seed (Figure~\ref{fig:envs}; details in
Appendix~\ref{app:envs}). \emph{HiddenEventCorridor} implements Definition~\ref{def:class} as a sequential task: $L$
report bits arrive one per step together with four nuisance bits, a delay corridor follows, and at the fork the agent may
buy up to $B$ audits before choosing $a=\pm1$. \emph{AuditGrid} adds navigation: a cue tile in a $7\times11$ room shows its
report bit only when the agent stands next to it, the two exits implement $a=\pm1$, audits are global query actions, and
an episode that lasts $40$ steps ends with $Y=0$. A full-history policy reaches the class optimum within $24$ steps in
every world, so grid regrets are exact. In both environments the proxy $\tilde Y=\one\{a=\mathrm{hint}\}$ rewards
following a hint shown at the decision point. The hint is independent of $Y$, so maximizing the proxy has true value
$\frac12$. We ran seven configurations: corridor with $k\in\{1,2,3\}$ ($\mu=0.8$, $B=0$), corridor long ($k=2$, $L=12$,
delay $48$), corridor $B{=}1$ ($k=2$, $\mu=0.6$, where $V^\star_1=0.75>V^\star_0=0.68$), grid ($k=2$, $\mu=0.8$, $B=0$), and
grid $B{=}1$ ($k=2$, $\mu=0.6$).

\begin{figure}[t]
\centering
\includegraphics[width=\linewidth]{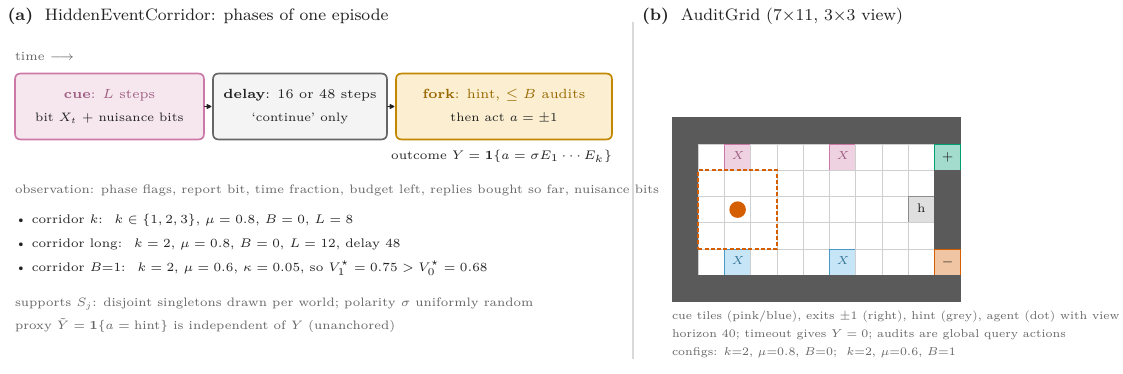}
\caption{\textbf{Deep RL environments.} \textbf{(a)} HiddenEventCorridor: a cue phase of $L$ report bits mixed with
nuisance bits, a delay corridor, and a fork where the agent may buy up to $B$ audits before acting. \textbf{(b)} AuditGrid:
cue tiles show their report bits only inside the agent's $3\times3$ view, the exits implement $a=\pm1$, and the hint
tile carries the proxy.}\label{fig:envs}
\end{figure}

\paragraph{Learners.}
The eight learners share the recurrent actor-critic of Section~\ref{sec:deep}, the joint action-and-audit space, the PPO
hyperparameters, and the number of training episodes (Appendix~\ref{app:hyper}). They differ in the evidence they buy,
their terminal return, and their auxiliary losses (Table~\ref{tab:methods}). DR is the comparison that matters most: it
has DORL's return, network, and budget, and lacks only the audit supervision of the memory.

\begin{table}[t]
\centering\small
\begin{tabular}{l>{\raggedright\arraybackslash}p{0.33\linewidth}>{\raggedright\arraybackslash}p{0.38\linewidth}}
\toprule
learner & terminal training signal & extra evidence and losses \\
\midrule
proxy & free proxy on every episode & -- \\
proxy+audit-aux & free proxy & side audits as memory targets (no verifier) \\
sparse-verified & $Y$ on verified episodes; others discarded & verification rate $\rho'$ \\
reward model & plug-in $g$ fitted to verified outcomes & verification rate $\rho'$ \\
DR & doubly-robust return (Prop.~\ref{prop:dr}) & verification rate $\rho'$ \\
DR+predictive & DR & rate $\rho'$; next-observation prediction \citep{ni2024} \\
\textbf{DORL} & DR & rate $\rho$; side-audit replies as memory targets \\
fully verified PPO & $Y$ on every episode & every episode verified ($4.0$--$7.5\times$ the charge) \\
\bottomrule
\end{tabular}
\caption{Learners in Parts B and C. Only the two learners that buy side audits feed detached event beliefs to the policy;
Part~C adds a DR learner that feeds its detached outcome-model beliefs, to match this interface.}\label{tab:methods}
\end{table}

\paragraph{Matching the evidence charge.}
DORL verifies a fraction $\rho=0.1$ of episodes and side-audits a fraction $\rho_a=0.25$. The outcome-supervised baselines
verify at $\rho'=\rho+\rho_akc_a/c_v$ instead, which equalizes the nominal charge of verifications and side audits. Their
realized evidence spending matched DORL's within $0.1\%$ in the corridors and within $8\%$ in the grid, where timed-out
episodes are not verified. Audits that a policy buys during training are charged on top
(Figure~\ref{fig:cert-pipeline}c), so Part~C also compares DORL and DR under one cap on total training charge.

\paragraph{Measurements and tests.}
The primary endpoint is the final regret $V^\star_B-J(\pihat)$ of the greedy policy, computed on $65{,}536$ fresh episodes
from evaluator-only information. We also record the training charge, the charge needed to first reach regret
$\le0.05$, the deployment audit rate, and the certified lower bound from $2{,}000$ verified deployment episodes
(Proposition~\ref{prop:cert}, $\alpha=0.05$). Each configuration has ten worlds. For each pair of learners we report the
mean paired regret difference with a bootstrap interval, the number of worlds won, and an exact one-sided paired
sign-flip test, Holm-corrected across configurations; aggregates are interquartile means (IQM) with stratified bootstrap
intervals \citep{agarwal2021}. The primary preregistered hypothesis is that DORL has lower regret than DR and than
DR+predictive. The secondary ones concern proxy hacking, plug-in reward models, certificate coverage, certified audit
use, and anchoring by audits alone. Appendix~\ref{app:stats} lists all hypotheses and test details. A pilot on held-out world
$100$ fixed one design choice: letting the policy read the detached event beliefs halved DORL's regret at $k=3$
(Table~\ref{tab:pilot}). Part~C repeats this ablation on the evaluation worlds.

\paragraph{Controls (Part C).}\label{sec:rev-design}
Part C runs the five controls of Table~\ref{tab:controls} on ten worlds per cell with Part~B's settings. Every PPO run in
Part~C also fits linear probes of the decision-time GRU state, ridge regressions trained on $4{,}096$ episodes and scored
on $4{,}096$ others, for each cue parity $\chi_{S_j}(X)$ and for the decision parity $\chi_{T^\star}(X)$.

\begin{table}[t]
\centering\small
\begin{tabular}{l>{\raggedright\arraybackslash}p{0.47\linewidth}>{\raggedright\arraybackslash}p{0.28\linewidth}}
\toprule
control & change & cells \\
\midrule
belief input & DORL without the detached event beliefs as policy input & corridor $k{=}2,3$, $B{=}1$; grid $k{=}2$ \\
matched interface & DR whose policy reads its detached outcome-model beliefs & corridor $k{=}2,3$, $B{=}1$; grid $k{=}2$ \\
longer training & DR with $4\times$ the episodes and charge & corridor $k{=}2,3$ \\
charge cap & DORL and DR stopped at the same training charge, policy audits included ($32{,}768$ corridor, $49{,}152$ grid) & corridor and grid $B{=}1$ \\
no RL & supervised learners on the same network, early-stopped at DORL's charge: $\mathrm{sup}_{\rm outcome}$ fits $(2Y-1)a$ from verified outcomes, $\mathrm{sup}_{\rm audit}$ fits each $E_j$ from audit replies & four $B{=}0$ corridors \\
\bottomrule
\end{tabular}
\caption{Part C controls. The two supervised learners see the same histories and differ only in their labels.}\label{tab:controls}
\end{table}

\subsection{A language-model policy (Part D)}\label{sec:llm-design}

Part D renders the hidden-event class as text. Each world has $L=16$ report bits, disjoint singleton supports,
$\mu=0.8$, and $k\in\{1,2,3\}$. A prompt lists the reports as sensor-log lines with an independent supervisor hint, and
Qwen2.5-0.5B-Instruct \citep{qwen2025qwen25} answers \texttt{Yes} or \texttt{No}. With $B=1$ it may first inspect one event
at price $\kappa=0.05$ and read the reply. Every learner trains rank-$32$ LoRA adapters with $600$ on-policy REINFORCE
updates and a leave-one-out baseline \citep{kool2019buy}. DORL verifies $10\%$ of training episodes, side-audits $25\%$, and
trains event-belief heads on the replies; DR verifies at the matched rate and uses the doubly-robust return. The other
learners optimize the proxy, sparse verified outcomes, a plug-in reward model, or fully verified outcomes. DR and DORL run
on ten worlds per cell and the others on five. We also ran a price sweep ($c_a\in\{0.05,0.3,1.0\}$ at $k=3$) and a
three-world scale check with Qwen2.5-1.5B-Instruct. We selected the LoRA learning rate on held-out world $100$ and tuned
nothing else. The endpoint is the exact regret of greedy decoding on $4{,}096$ fresh episodes, and the pre-specified tests
compare DR with DORL at $B=0$ and $k\in\{2,3\}$ with exact paired sign-flip tests and Holm correction. As in Part C,
linear probes decode the cue and decision parities from the decision state, and $2{,}000$ verified episodes certify the
deployed value.

\subsection{Verification in language-model fine-tuning (Part E)}\label{sec:partE-design}

Parts A--D use a proxy that is unanchored by construction. Part E keeps only the verification channel and makes the
proxy informative but exploitable, to measure how much verification an RL fine-tuning run needs to stay anchored to the
task outcome. It reuses $39$ runs from an earlier study with its own pre-specified predictions
(Appendix~\ref{app:stats}); the connection to Propositions~\ref{prop:anchor} and~\ref{prop:dr} is ours.

\paragraph{Task and proxy.}
Qwen2.5-0.5B-Instruct \citep{qwen2025qwen25} solves integer expressions with one to three operations. Each prompt adds
``A classmate says the answer is $h$'', and the suggestion $h$ is correct with probability $0.3$. A program checks the
final integer and returns $Y$; it plays the role of the verifier at price $c_v$ per response. The free proxy is
\begin{equation}\label{eq:partE-proxy}
\tilde Y=(1-\alpha)\,\hat r+\alpha\,\one\{\text{answer}=h\},\qquad \alpha=0.5,
\end{equation}
where $\hat r$ is a logistic probe on the frozen reference model, fitted to $4{,}096$ verified responses of the initial
policy (held-out AUC $0.887$). The copy term is an injected sycophancy bias \citep{sharma2024towards}: a policy that always
copies earns at least $\alpha$ in proxy value but is correct only $30\%$ of the time. This gives a controlled instance of
reward-model over-optimization \citep{gao2023} with a known exploit.

\paragraph{Learners.}
All learners train with RLOO \citep{kool2019buy,ahmadian2024back} and a KL penalty to the reference, and differ only in
the terminal return of each response (Table~\ref{tab:partE-learners}). Each response is verified independently with
probability $p$. DR and verified-only use $p=\rho$; DR-Neyman and DR-recal allocate $p$ in proportion to the square root of
an estimated residual variance in cells of (number of operations, proxy decile), as in Neyman allocation
\citep{neyman1934}, rescaled so that the mean is $\rho$. The five verifying learners run at $\rho\in\{0.05,0.2\}$ with three
seeds each, $39$ runs in total, and each run trains on $25{,}600$ responses. Unverified outcomes are masked from training,
and we checked that flipping them leaves every training signal unchanged.

\begin{table}[t]
\centering\small
\begin{tabular}{l>{\raggedright\arraybackslash}p{0.43\linewidth}l}
\toprule
learner & terminal return $R$ & verification \\
\midrule
proxy & $\tilde Y$ & none \\
proxy\,+\,10$\times$KL & $\tilde Y$, KL coefficient $0.2$ instead of $0.02$ & none \\
naive substitution & $VY+(1-V)\tilde Y$; mean $\rho Y+(1-\rho)\tilde Y$, biased & $p=\rho$ \\
verified-only (IPW) & $VY/\rho$: Proposition~\ref{prop:dr} with $g\equiv0$ & $p=\rho$ \\
DR & $\tilde Y+V(Y-\tilde Y)/\rho$: Proposition~\ref{prop:dr} with $g=\tilde Y$ & $p=\rho$ \\
DR-Neyman & $\tilde Y+V(Y-\tilde Y)/p$ & residual-adaptive $p$, mean $\rho$ \\
DR-recal & $g+V(Y-g)/p$, $g$ the proxy recalibrated on earlier verifications & residual-adaptive $p$, mean $\rho$ \\
fully verified & $Y$ & every response \\
\bottomrule
\end{tabular}
\caption{Learners in Part E. All use RLOO with eight responses per prompt and a KL coefficient of $0.02$ unless stated.}
\label{tab:partE-learners}
\end{table}

\paragraph{Measurements.}
Every $25$ steps we evaluate each run on $512$ held-out responses and record gold accuracy, proxy value, the rate at which
the final answer copies $h$, and cumulative verifications. We report final, peak, and time-averaged gold accuracy (the
area under the nine checkpoint accuracies), as means and standard deviations over three seeds, without significance
tests. The verification count, which excludes the $4{,}096$ setup labels of $\hat r$, is the verification charge in units of
$c_v$. A secondary analysis uses each run's nine checkpoints to compare proxy-only, uniform, and decision-directed
selection of a checkpoint under a verification budget (Appendix~\ref{app:partE-extra}). Appendix~\ref{app:partE} gives the
prompt format, the reward model, and the hyperparameters.
\section{Results}\label{sec:results}

\subsection{Certified learners follow the predicted scaling}\label{sec:results-A}

All $3{,}200$ certified runs of the $B=0$ grid stopped with a correct certificate (Table~\ref{tab:routes}). These
certificates claim zero regret, and the realized regret was zero in every run (per-cell Wilson lower bound on coverage
$0.963$). In the deployment-audit grid all $600$ runs were covered and chose the optimal audit set, including the cell
with a $10\%$-noisy verifier and $B=2$; their largest realized regret was $0$, against certified radii near $0.09$
(Table~\ref{tab:cover}).

\begin{figure}[t]
\centering
\includegraphics[width=\linewidth]{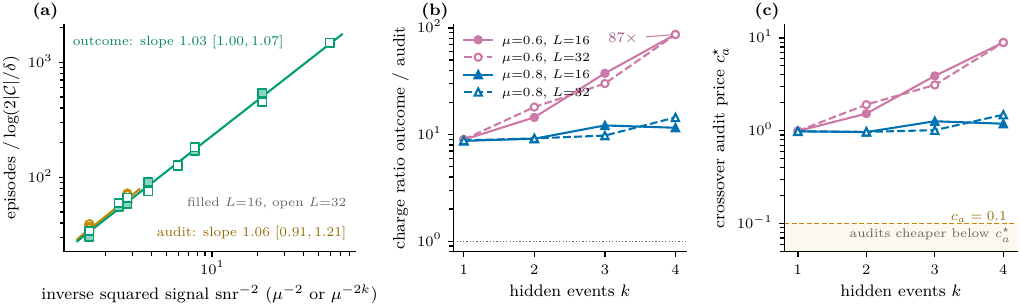}
\caption{Part A ($16$ cells, $100$ random worlds each). \textbf{(a)} Median episodes to certify, normalized by
$\log(2|\mathcal C|/\delta)$, against the inverse squared signal ($\mu^{-2}$ for the audit route, $\mu^{-2k}$ for the
outcome route). Lines are least-squares fits on the log scale, with the slope and its $95\%$ interval;
Theorem~\ref{thm:upper} predicts slope $1$. \textbf{(b)} Ratio of median charges, outcome route over audit route, against
$k$. \textbf{(c)} Crossover audit price $c_a^\star$; the shaded band lies below the price $c_a=0.1$ used throughout. Solid
lines: $L=16$; dashed: $L=32$.}\label{fig:routes}
\end{figure}

Both routes follow the predicted law (Figure~\ref{fig:routes}a). Regressing $\log\{n/\log(2\,\mathrm{count}/\delta)\}$ on
$\log(\mathrm{signal}^{-2})$ gives slope $1.03$ $[1.00,1.07]$ for the outcome route ($16$ cells, $8$ signal values) and
$1.06$ $[0.91,1.21]$ for the audit route. The audit route has only two signal values, so its slope is less precise. The two
intercepts, $3.05$ and $3.09$, are close, so a single law $n\approx e^{3.07}\log(2\,\mathrm{count}/\delta)\,\mathrm{signal}^{-2}$
describes both routes; they differ only in signal and hypothesis count.

The routes separate as $k$ grows. With $\mu=0.6$ and $L=32$, going from $k=1$ to $k=4$ multiplies the outcome route's
episodes by $44$ ($477\to21{,}176$) and the audit route's by $1.25$ ($478\to598$). Each audit-route episode buys $k$ audits,
yet at $k=4$ the outcome route still costs $87$ times more (Figure~\ref{fig:routes}b). The crossover audit price rises from
about one verifier price at $k=1$ to $8.9$ verifier prices at $k=4$ (Figure~\ref{fig:routes}c).

\begin{table}[t]
\centering\small
\begin{tabular}{rrr rrr rrr r}
\toprule
 & & & \multicolumn{3}{c}{audit route (Alg.~1)} & \multicolumn{3}{c}{outcome-only route} & \\
\cmidrule(lr){4-6}\cmidrule(lr){7-9}
$L$ & $k$ & $\mu$ & episodes & charge & cover & episodes & charge & cover & $c_a^\star$ \\
\midrule
16 & 1 & 0.6 & 383 & 43 & 100/100 & 382 & 386 & 100/100 & 0.99 \\
16 & 1 & 0.8 & 197 & 23 & 100/100 & 196 & 198 & 100/100 & 0.98 \\
16 & 2 & 0.6 & 478 & 101 & 100/100 & 1{,}456 & 1{,}471 & 100/100 & 1.53 \\
16 & 2 & 0.8 & 246 & 53 & 100/100 & 477 & 482 & 100/100 & 0.97 \\
16 & 3 & 0.6 & 478 & 149 & 100/100 & 5{,}552 & 5{,}608 & 100/100 & 3.90 \\
16 & 3 & 0.8 & 246 & 77 & 100/100 & 932 & 941 & 100/100 & 1.27 \\
16 & 4 & 0.6 & 478 & 197 & 100/100 & 16{,}941 & 17{,}110 & 100/100 & 8.93 \\
16 & 4 & 0.8 & 307 & 127 & 100/100 & 1{,}456 & 1{,}471 & 100/100 & 1.19 \\
32 & 1 & 0.6 & 478 & 54 & 100/100 & 477 & 482 & 100/100 & 0.99 \\
32 & 1 & 0.8 & 246 & 28 & 100/100 & 245 & 247 & 100/100 & 0.99 \\
32 & 2 & 0.6 & 478 & 101 & 100/100 & 1{,}819 & 1{,}837 & 100/100 & 1.91 \\
32 & 2 & 0.8 & 307 & 65 & 100/100 & 597 & 603 & 100/100 & 0.97 \\
32 & 3 & 0.6 & 598 & 186 & 100/100 & 5{,}552 & 5{,}608 & 100/100 & 3.12 \\
32 & 3 & 0.8 & 307 & 96 & 100/100 & 932 & 941 & 100/100 & 1.01 \\
32 & 4 & 0.6 & 598 & 246 & 100/100 & 21{,}176 & 21{,}388 & 100/100 & 8.92 \\
32 & 4 & 0.8 & 307 & 127 & 100/100 & 1{,}819 & 1{,}837 & 100/100 & 1.49 \\
\bottomrule
\end{tabular}
\caption{Part A: certified learning in $\cM(L,k,1,\mu)$ with $B=0$, $\delta=0.05$, and prices $c_0=0.01$, $c_a=0.1$, $c_v=1$.
Medians over $100$ random worlds, and the number of runs whose realized regret was within the certificate. $c_a^\star$ is
the audit price, in units of $c_v$, at which the two routes' median charges are equal.}\label{tab:routes}
\end{table}

\begin{table}[t]
\centering\small
\begin{tabular}{rrrl rrrrr}
\toprule
$k$ & $B$ & $\eta$ & $\mu$ & certified & covered (95\% CI) & $A^\star$ found & episodes & max regret \\
\midrule
2 & 1 & 0.0 & (0.95, 0.60) & 1.00 & 200/200 (0.981, 1.000) & 1.00 & 13{,}554 & 0.0000 \\
3 & 1 & 0.0 & (0.95, 0.70, 0.50) & 1.00 & 200/200 (0.981, 1.000) & 1.00 & 33{,}089 & 0.0000 \\
3 & 2 & 0.1 & (0.95, 0.70, 0.50) & 1.00 & 200/200 (0.981, 1.000) & 1.00 & 33{,}100 & 0.0000 \\
\bottomrule
\end{tabular}
\caption{Part A: Algorithm~\ref{alg:dsl} with deployment audits, $\kappa=0.05$, $\varepsilon=0.1$, $L=16$, and $200$ worlds
per row. ``$A^\star$ found'' is the fraction of runs whose deployed audit set is optimal.}\label{tab:cover}
\end{table}

\subsection{Only DORL learns the memory in multi-event tasks}\label{sec:results-deep}

\begin{figure}[t]
\centering
\includegraphics[width=\linewidth]{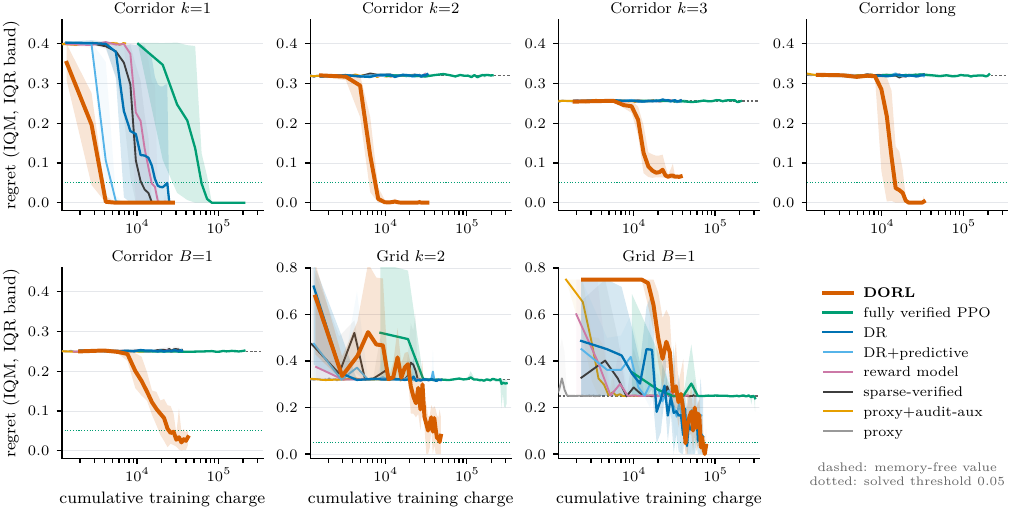}
\caption{Part B: true regret against cumulative training charge (logarithmic axis), IQM over the ten worlds with the
interquartile range shaded. Dashed lines mark the memory-free value and dotted lines the solved threshold $0.05$. Fully
verified PPO's curves extend to its larger charge.}\label{fig:curves}
\end{figure}

All $560$ Part B runs finished on two T4 GPUs in $12.0$ GPU-hours. Figure~\ref{fig:worlds} gives the final regrets (Table~\ref{tab:deep} in Appendix~\ref{app:partB-extra}
lists them with charges), and Figure~\ref{fig:curves} the learning curves against training charge.

\paragraph{Corridors.}
With one hidden event every verified learner solves the corridor, and DORL and DR do not differ ($p_{\rm Holm}=0.25$). With
two or more events, every PPO learner except DORL ends at the memory-free value: regret $0.320$, $0.256$, $0.320$, and
$0.250$ on corridors $k{=}2$, $k{=}3$, long, and $B{=}1$. This holds for sparse verification, the reward model, DR, DR with a
self-predictive loss, and fully verified PPO at $4.9$--$6.3\times$ DORL's charge. DORL reaches IQM regret $0.000$, $0.066$,
$0.003$, and $0.035$. In each of the four tasks it beats DR and DR+predictive in all ten paired worlds (sign test
$p=2^{-10}$; Holm-adjusted sign-flip $p=0.007$), with mean paired reductions against DR of $0.319$, $0.194$, $0.293$, and
$0.206$. DORL first reached regret $\le0.05$ after a median training charge of $8{,}204$ on corridor $k{=}2$ and $14{,}010$
on corridor long; on corridor $k{=}3$ its median final regret, $0.065$, sits just above that threshold. DR+predictive
matched DR within $0.001$, so predicting observations did not help the memory.

\paragraph{Grid.}
In AuditGrid DORL had the lowest IQM regret in both tasks ($0.080$ against $0.319$ for DR at $k=2$; $0.033$ against $0.075$
at $B=1$), but its results were bimodal: it solved $5/10$ worlds in each task, against $2/10$ and $6/10$ for DR, and the
tests did not reject ($p_{\rm Holm}$ between $0.09$ and $0.25$). Even fully verified PPO solved only $2/10$ grid $k{=}2$
worlds, which suggests that finding the cue tiles limits every learner there.

\paragraph{Deployment audits.}
With $B=1$, DORL bought $0.71$ (corridor) and $0.99$ (grid) audits per deployed episode and certified audit use,
$\mathrm{LCB}>V^\star_0$, in $6/10$ worlds of each task. In the corridor no other learner certified audit use or bought
audits in more than $9\%$ of episodes; in the grid DR also certified it in $6/10$ worlds. Because policy audits during
training are charged on top of the matched evidence, DORL spent $19\%$ (corridor) and $9\%$ (grid) more training charge
than DR (Figure~\ref{fig:cert-pipeline}c). Section~\ref{sec:rev} equalizes this.

\paragraph{Proxy learners.}
The proxy learner reached the shortcut value everywhere, with proxy scores of $0.90$--$1.00$. Side audits without a
verifier did not anchor the outcome: proxy+audit-aux stayed at $V^\star_B-\frac12$ in all seven tasks and lost to DORL
($p_{\rm Holm}=0.007$), as Proposition~\ref{prop:anchor} predicts. The preregistered comparison of the proxy learner with DR
was significant only at $k=1$, because elsewhere DR also failed, for lack of memory, at the same regret. Part E tests
proxy hacking directly. DR and the plug-in reward model never differed. Certification adds a charge of $2{,}020$ plus any
audits, $3$--$7\%$ on top of DORL's and DR's training charge (Table~\ref{tab:deep}).

\begin{figure}[t]
\centering
\includegraphics[width=\linewidth]{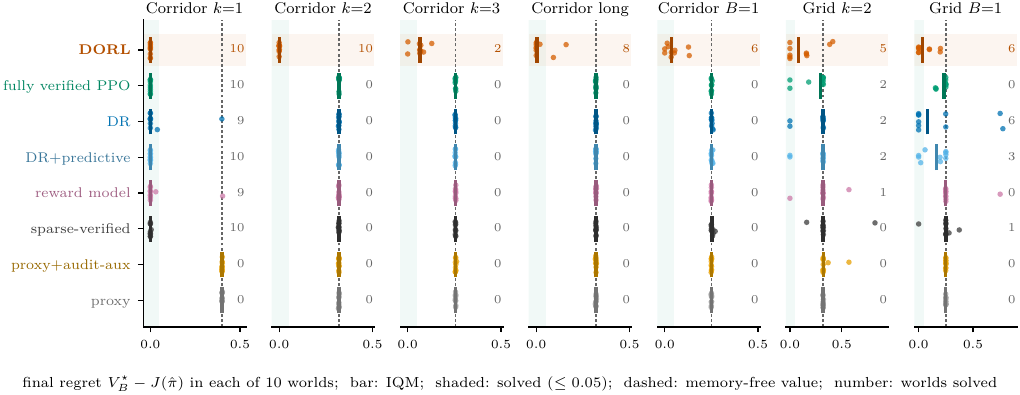}
\caption{Part B: final regret of every learner in each of the ten worlds of every configuration. Bars mark the IQM; the
shaded band is the solved region (regret $\le0.05$); dashed lines mark the memory-free value $V^\star_B-\frac12$; numbers
count worlds solved. With $k\ge2$, every learner except DORL sits on the memory-free value in every corridor
world.}\label{fig:worlds}
\end{figure}

\subsection{Audit supervision of the memory causes the gain}\label{sec:rev}

\begin{figure}[t]
\centering
\includegraphics[width=\linewidth]{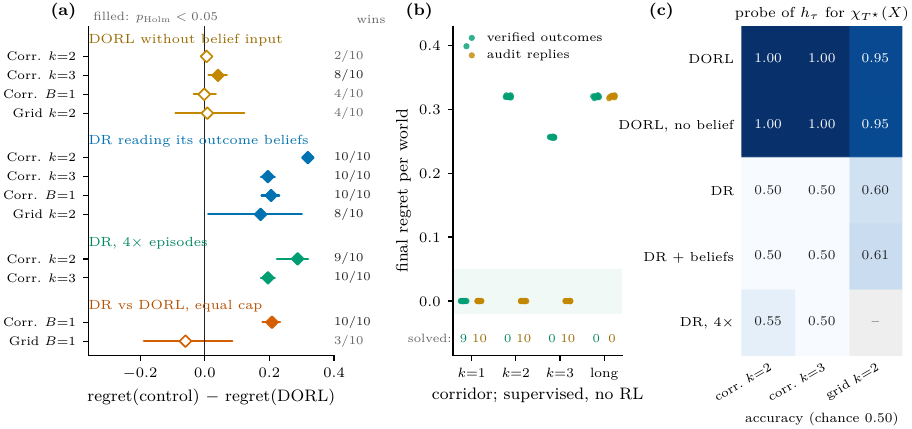}
\caption{Part C (post hoc). \textbf{(a)} Paired regret differences between each control and DORL (mean and $95\%$ bootstrap
interval over ten worlds; positive values favour DORL), with worlds won by DORL. The last group compares DR and DORL
under the same training-charge cap. \textbf{(b)} Per-world regret of the two supervised learners without RL, at equal
charge; numbers count worlds solved. \textbf{(c)} Accuracy of a linear probe of the decision-time GRU state for the
decision parity $\chi_{T^\star}(X)$ in the $B{=}0$ cells; ``--'': not run.}\label{fig:mechanism}
\end{figure}

All $300$ Part C runs finished in $8.3$ GPU-hours, and the $80$ reruns of DORL and DR reproduced Part B's final regrets
exactly. Figure~\ref{fig:mechanism} summarizes the controls, and Table~\ref{tab:rev} lists all regrets, charges, and probe scores; Table~\ref{tab:revtests} in
Appendix~\ref{app:partC-extra} gives the tests.

\paragraph{Policy input and interface.}
Removing the event beliefs from DORL's policy input changed regret by less than $0.01$ in corridor $k{=}2$, corridor $B{=}1$,
and grid $k{=}2$. Only at corridor $k{=}3$ did the beliefs help ($0.066$ versus $0.111$; paired difference $+0.041$
$[+0.012,+0.069]$, $p_{\rm Holm}=0.043$). Without belief input DORL still solved $9/10$ worlds in corridor $k{=}2$ and $8/10$
in corridor $B{=}1$, where DR solved none. In the other direction, letting DR's policy read its own outcome beliefs left
it at the memory-free value in every corridor, and DORL beat it in all ten worlds of each ($p_{\rm Holm}=0.004$) and in
$8/10$ grid worlds ($p_{\rm Holm}=0.026$).

\paragraph{Training length and charge.}
DR trained on four times as many episodes, at $4.0\times$ DORL's charge, stayed at the memory-free value (IQM regret
$0.319$ and $0.257$ at $k=2,3$), and DORL beat it in $9/10$ and $10/10$ worlds ($p_{\rm Holm}=0.002$). Under one cap on
training charge, policy audits included, DORL and DR spent $32{,}723$ and $32{,}730$ in corridor $B{=}1$. DORL kept its
performance (IQM regret $0.032$, $7/10$ solved) while DR stayed at $0.251$ ($0/10$); DORL won all ten worlds, by $0.208$
$[0.178,0.233]$ on average. In grid $B{=}1$ the cap reversed the order. DORL spends more on policy audits per iteration,
completed $386$ of $600$ iterations against DR's $426$, and reached IQM regret $0.196$ ($2/10$ solved) against DR's $0.114$
($5/10$); the mean paired difference was $-0.060$ $[-0.189,+0.085]$.

\paragraph{Supervision without RL.}
The two supervised learners see the same histories at the same charge and differ only in their labels. With one event both
succeed. With two and three events the learner trained on audit replies solved all ten worlds and the learner trained
on verified outcomes solved none, with validation cross-entropy within $0.002$ of $\ln2$ in every world
($p_{\rm Holm}=0.004$; Figure~\ref{fig:mechanism}b). At equal charge the audit learner gets $4.8\times$ ($k=2$) and $3.3\times$
($k=3$) more labelled episodes, each with $k$ labels of strength $\mu$ instead of one label of strength $\prod_j\mu_j$, which
is the mechanism of Corollary~\ref{cor:separation}. On corridor long both supervised learners failed. They apply their
loss at the decision step, $48$ steps after the cue, whereas DORL applies it at every step and solved this task in Part B.

\paragraph{Probes.}
A linear probe of DORL's decision-time state recovers the decision parity $\chi_{T^\star}$ with accuracy $0.998$--$1.000$
in the corridors and $0.95$ in the grid, with or without belief input. For DR and its variants the same probe scores
$0.50$--$0.55$ in the corridors and $0.60$--$0.61$ in the grid, close to chance (Figure~\ref{fig:mechanism}c). The cue
parities show the same pattern. Probes are uninformative in the $B{=}1$ cells, where the probe targets correlate with the
audited reply.

\subsection{The gain carries over to a language-model policy}\label{sec:llm-results}

\begin{figure}[t]
\centering
\includegraphics[width=\linewidth]{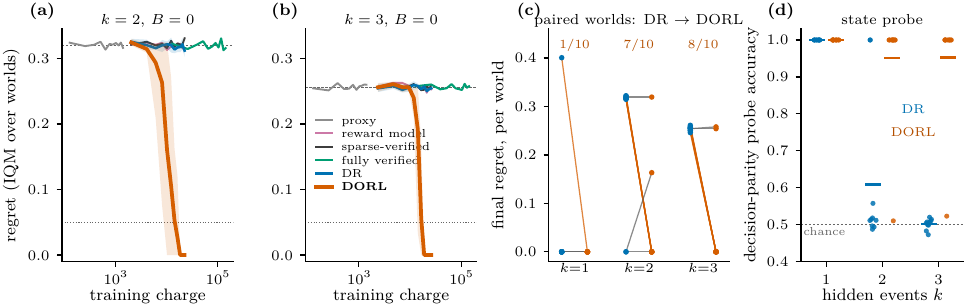}
\caption{Part D: Qwen2.5-0.5B-Instruct on rendered hidden-event logs ($B=0$, $c_a=0.1$). \textbf{(a, b)} Regret against
training charge (IQM over worlds; band: interquartile range for DR and DORL) at $k=2$ and $k=3$. Dashed: memory-free value;
dotted: solved threshold $0.05$. Only DORL leaves the memory-free value. \textbf{(c)} Final regret of DR (left) and DORL
(right) in each paired world; numbers count worlds in which DORL has strictly lower regret (the remaining $k=1$ worlds are
ties at zero). \textbf{(d)} Linear-probe accuracy for the decision parity $\chi_{T^\star}$ from the decision state, per world;
bars are means.}\label{fig:llm}
\end{figure}

All $212$ Part D runs finished (Figure~\ref{fig:llm}, Table~\ref{tab:llm-primary}). Without
deployment audits, DORL lowered mean paired regret against DR by $0.206$ $[0.094,0.317]$ at $k=2$ and by $0.203$
$[0.127,0.255]$ at $k=3$ ($p_{\rm Holm}=0.0078$ for both), and solved $8/10$ worlds in each, against $2/10$ and $0/10$ for DR.
Both learners had nearly the same total charge ($26{,}544$ and $26{,}607$ at $k=2$). At $k=1$ the learners tied at zero regret in nine
worlds, and DR failed in the tenth. The decision-parity probe scored $0.951$ and $0.952$ on DORL's state at $k=2,3$, against $0.609$ and $0.502$ on
DR's. The proxy learner stayed near value $0.5$, and fully verified REINFORCE solved none of its five worlds at $k=2$ or
$3$ despite its higher charge.

With $B=1$, DORL reduced regret by $0.227$ $[0.131,0.321]$ at $k=2$ and by $0.027$ $[-0.002,0.080]$ at $k=3$. No learner
inspected an event in its greedy evaluation, so the $B=1$ gain at $k=2$ comes from audits bought during training. In a
price sweep at $k=3$ the reduction stayed between $0.16$ and $0.21$ for audit prices from $0.05$ to $1.0$, so the sweep
found no crossover. Three-world checks with the $1.5$B model kept positive gaps ($0.109$ and $0.150$).

\begin{table}[t]
\centering\small
\resizebox{\linewidth}{!}{\begin{tabular}{crrrrr}
\toprule
$k$ & DR IQM regret [95\% CI] & DORL IQM regret [95\% CI] & solved DR/DORL & paired $\Delta$ [95\% CI] & $p$ \\
\midrule
1 & 0.000 [0.000, 0.040] & 0.000 [0.000, 0.000] & 9/10\;/\;10/10 & +0.040 [0.000, 0.120] & 0.500 \\
2 & 0.317 [0.159, 0.320] & 0.000 [0.000, 0.082] & 2/10\;/\;8/10 & +0.206 [0.094, 0.317] & 0.0078 \\
3 & 0.255 [0.251, 0.257] & 0.000 [0.000, 0.127] & 0/10\;/\;8/10 & +0.203 [0.127, 0.255] & 0.0078 \\
\bottomrule
\end{tabular}}
\caption{Part D: language-model decisions with $B=0$, $c_a=0.1$, and Qwen2.5-0.5B-Instruct on the same ten worlds per learner.
IQM regret is the interquartile mean across worlds; $\Delta$ and its bootstrap interval are based on paired \emph{mean}
regret differences. A world is solved at regret $\le0.05$. The $k=2,3$ $p$-values are one-sided exact paired sign-flip
values corrected together by Holm; $k=1$ is tested separately.}\label{tab:llm-primary}
\end{table}

\subsection{Sparse verification prevents proxy hacking}\label{sec:partE-results}

\begin{figure}[t]
\centering
\includegraphics[width=\linewidth]{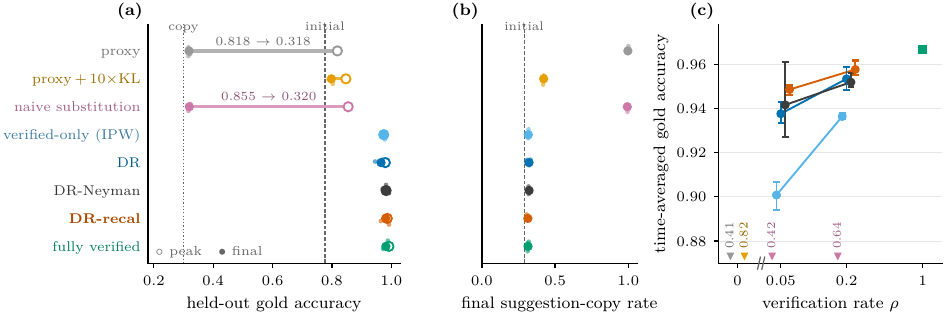}
\caption{Part E: sparse verification prevents proxy hacking in Qwen2.5-0.5B-Instruct arithmetic fine-tuning
($39$ RLOO runs, three seeds per learner and rate). \textbf{(a)} Held-out gold accuracy ($512$ responses) at the best
of nine checkpoints (open, step $0$ included) and at the end of training (filled), as means over seeds; small dots are
the individual final values. The verified learners are shown at $\rho=0.05$. Dashed line: mean step-$0$ accuracy over
all runs; dotted line: the suggestion's accuracy $0.3$, the value of always copying. \textbf{(b)} Final
suggestion-copy rate, with mean and seeds; dashed line: copy rate of the initial policy on the $4{,}096$ setup
responses. \textbf{(c)} Time-averaged gold accuracy (area under the nine checkpoint accuracies divided by $200$
steps) against the verification rate. Points are means, bars are min--max over seeds, and lines join the two rates of
the same learner. Triangles mark means below the axis, labelled with their value. The two proxy learners are plotted at
$\rho=0$, left of the axis break. The fully verified learner is plotted at $\rho=1$. Verification rates exclude the
$4{,}096$ setup labels of the reward model.}\label{fig:partE}
\end{figure}

Proxy-only training raised the proxy value to $0.959$ while gold accuracy fell from a peak of $0.818$ to $0.318\pm0.003$,
close to the $0.3$ accuracy of always copying the suggestion; the final copy rate was $0.997$ (Table~\ref{tab:partE},
Figure~\ref{fig:partE}). In two of three seeds the best checkpoint was the initial policy. A tenfold KL penalty slowed the
decline but did not stop it (final accuracy $0.798$, copy rate $0.422$). Mixing verified labels into the proxy without
correction did not help either: naive substitution ended at $0.320$ with $5\%$ verification and at $0.448$ with $20\%$.

With $5\%$ verification, every learner that uses a correctly weighted verified return stayed anchored. DR-recal ended at
$0.981\pm0.014$ with $1{,}288$ verifications, the same as fully verified training ($0.982\pm0.008$) with $25{,}600$.
Verified-only reached $0.975$, DR $0.965$, and DR-Neyman $0.982$, all with copy rates near the fully verified $0.315$. The
learners differed more in speed than in endpoint: time-averaged accuracy was $0.901$ for verified-only, $0.938$ for DR,
$0.942$ for DR-Neyman, and $0.949$ for DR-recal, against $0.967$ for full verification, and every DR-recal run
($0.946$--$0.951$) finished above every verified-only run ($0.894$--$0.906$). This ordering matches
Proposition~\ref{prop:dr}, whose conditional variance $(p^{-1}-1)(Y-g)^2$ shrinks as the control variate $g$ approaches $Y$.
Counting the $4{,}096$ setup labels of the reward model, DR-recal at $5\%$ used $5{,}384$ verifications, about as many as
verified-only at $20\%$ ($5{,}060$), and still learned faster ($0.949$ against $0.936$).

The proxy failure matches Proposition~\ref{prop:anchor}. The copy term pays the same whether the suggestion is right or
wrong, so it cannot separate a world where copying is optimal from this one, and the probe $\hat r$ was anchored only on the
initial policy's responses. Naive substitution has mean $\rho Y+(1-\rho)\tilde Y$, a biased plug-in return of the kind
Proposition~\ref{prop:dr} warns about.

In the secondary checkpoint-selection analysis, decision-directed elimination had higher regret than uniform DR at every
budget (for example $0.0319$ against $0.0283$ at $64$ verifications), and proxy-only ranking had regret $0.1288$
(Table~\ref{tab:partE-selection}). With nine checkpoints per run, concentrating verifications on the leading candidates did
not pay off.

\begin{table}[t]
\centering\small
\begin{tabular}{lrrcccc}
\toprule
learner & $\rho$ & verif. & final gold & peak gold & time-avg.\ gold & copy rate \\
\midrule
proxy & -- & 0 & $0.318\pm0.003$ & $0.818\pm0.044$ & $0.406\pm0.008$ & $0.997\pm0.006$ \\
proxy\,+\,10$\times$KL & -- & 0 & $0.798\pm0.008$ & $0.847\pm0.005$ & $0.817\pm0.002$ & $0.422\pm0.004$ \\
fully verified & 1 & 25{,}600 & $0.982\pm0.008$ & $0.992\pm0.004$ & $0.967\pm0.000$ & $0.315\pm0.005$ \\
\midrule
naive substitution & 0.05 & 1{,}242 & $0.320\pm0.003$ & $0.855\pm0.026$ & $0.423\pm0.006$ & $0.993\pm0.001$ \\
verified-only (IPW) & 0.05 & 1{,}288 & $0.975\pm0.005$ & $0.975\pm0.005$ & $0.901\pm0.006$ & $0.317\pm0.001$ \\
DR & 0.05 & 1{,}263 & $0.965\pm0.018$ & $0.980\pm0.010$ & $0.938\pm0.005$ & $0.322\pm0.012$ \\
DR-Neyman & 0.05 & 1{,}305 & $0.982\pm0.011$ & $0.982\pm0.012$ & $0.942\pm0.018$ & $0.322\pm0.005$ \\
DR-recal & 0.05 & 1{,}288 & $0.981\pm0.014$ & $0.986\pm0.006$ & $0.949\pm0.002$ & $0.314\pm0.006$ \\
\midrule
naive substitution & 0.2 & 5{,}145 & $0.448\pm0.099$ & $0.918\pm0.031$ & $0.637\pm0.043$ & $0.853\pm0.096$ \\
verified-only (IPW) & 0.2 & 5{,}060 & $0.969\pm0.020$ & $0.984\pm0.002$ & $0.936\pm0.001$ & $0.318\pm0.002$ \\
DR & 0.2 & 5{,}141 & $0.986\pm0.006$ & $0.986\pm0.006$ & $0.953\pm0.005$ & $0.318\pm0.005$ \\
DR-Neyman & 0.2 & 5{,}202 & $0.980\pm0.007$ & $0.988\pm0.005$ & $0.952\pm0.004$ & $0.317\pm0.001$ \\
DR-recal & 0.2 & 5{,}224 & $0.983\pm0.004$ & $0.988\pm0.002$ & $0.958\pm0.004$ & $0.318\pm0.002$ \\
\bottomrule
\end{tabular}
\caption{Part E endpoints: mean $\pm$ standard deviation over three seeds of held-out gold accuracy at the final and
best checkpoints, time-averaged gold accuracy, and final suggestion-copy rate. ``verif.'' is the mean number of
on-policy verifications, the verification charge in units of $c_v$. It excludes the $4{,}096$ verified setup
responses of the reward model, which all learners except verified-only and fully verified rely on.}\label{tab:partE}
\end{table}

\subsection{Certificates hold for trained policies}\label{sec:results-cert}

\begin{figure}[t]
\centering
\includegraphics[width=\linewidth]{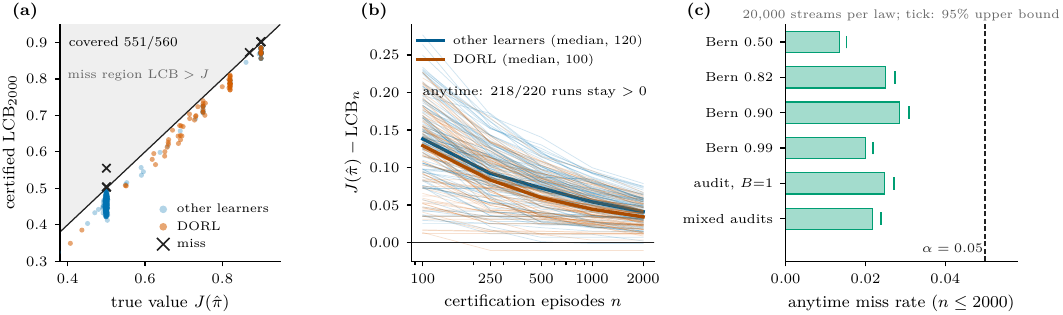}
\caption{Certificate soundness. \textbf{(a)} Certified lower bound after $2{,}000$ verified deployment episodes against the
true value, for all $560$ Part~B runs; crosses mark the nine misses. \textbf{(b)} Gap $J(\hat\pi)-\mathrm{LCB}_n$ along the
certification sequence for the $220$ Part~C PPO runs, with medians for DORL and the other learners; a path below zero
would be an anytime miss. \textbf{(c)} Simulated time-uniform miss rates of the underlying PrPl-EB sequence for six
utility laws (Appendix~\ref{app:cert}).}\label{fig:cert}
\end{figure}

Every Part A certificate held. In Part B the lower bound after $2{,}000$ verified episodes covered the true value in
$551$ of $560$ runs ($0.984$; world-cluster bootstrap $[0.963,0.998]$), and every learner--task cell covered at least $9$
of its $10$ worlds (Figure~\ref{fig:cert}a, Table~\ref{tab:coverage}). Learners that reach the same policy in the same world
share a certification sample; counting such runs once leaves $404$ distinct certificates, of which $399$ were covered. In
Part C we checked the bound at every one of the $2{,}000$ steps: it held throughout in $218$ of $220$ runs
(Figure~\ref{fig:cert}b). In Part D it covered the deployed value in $210$ of $212$ runs. In simulation the underlying
sequence missed at any time in $1.3$--$2.9\%$ of streams at $\alpha=0.05$ (Figure~\ref{fig:cert}c). The certified radius
also separates success from failure: DORL's radii were $0.03$--$0.10$, while the learners stuck at the memory-free value had
radii of $0.30$--$0.37$.

\section{Discussion}\label{sec:discussion}

\paragraph{Summary.}
We measured reinforcement learners by the charge they need to certify a deployed decision. Under this accounting the
three evidence channels play distinct roles. Verification anchors the objective (Proposition~\ref{prop:anchor}), and in
language-model fine-tuning a $5\%$ verification rate with a doubly-robust return prevented proxy hacking. Audits bought
during training teach the memory: when the outcome is a product of $k$ hidden events, they replace one signal of order
$\gamma^k$ on a degree-$k$ parity by $k$ signals of order $\gamma$ on single cues, an exponential saving in sample
complexity (Corollary~\ref{cor:separation}) and in gradient signal (Proposition~\ref{prop:gradient}). DORL-PPO turns this
saving into a training method that solved every multi-event corridor task at matched charge, and audits bought at
deployment, certified by Proposition~\ref{prop:cert}, are the only route above the reply-blind value
(Lemma~\ref{lem:oracle}). The finite-world theory of Section~\ref{sec:foundations} places all three channels in one
characteristic cost.

\paragraph{Application scenarios.}
The framework fits tasks in which the outcome is checkable but costly to check and intermediate checks reveal hidden
facts that bear on a later decision. In tool-using language-model agents, unit tests and static checks act as audits, an
end-to-end acceptance test as the verifier, and a self-assessment as the proxy. Inspection and maintenance robots can
query a sensor before acting; clinical decision support can order a laboratory test before treatment; experimental
science can run a cheap assay before an expensive validation. In each case the charge ledger tells the practitioner how
to split a budget between audits and verification, and the certificate states what the trained policy guarantees.

\paragraph{Future directions.}
The main theoretical direction is a learner for unknown kernels whose charge is polynomial in a checkable
decision-continuation dimension, with a matching lower bound; it would join the finite-world characterization and the
hidden-event mechanism in one theory. Further directions are noisy or strategic audits, audit menus learned from data,
and deployment monitors that change over time. Empirically, the next step is a tool-using agent trained and certified
with real audit and verification prices.

\clearpage
\appendix
\etocdepthtag.toc{mtappendix}
\phantomsection
\section*{Guide to the appendices}\label{app:guide}
\addcontentsline{toc}{section}{Guide to the appendices}

Appendices~\ref{app:foundations} and~\ref{app:proofs} contain the proofs, one subsection per result, each opening with a
short roadmap. Appendix~\ref{app:fw-checks} checks the finite-world theory on small exact examples.
Appendices~\ref{app:exp} and~\ref{app:extra} give the experimental protocols and the full result tables. This guide
lists the appendices, the dependencies among the results, where each result is stated and proved, and the notation.

\begingroup
\etocsettagdepth{mtmain}{none}
\etocsettagdepth{mtappendix}{subsection}
\etocsettocstyle{\subsection*{Contents of the appendices}\small}{}
\tableofcontents
\endgroup

\subsection*{How the results depend on each other}
Two tools carry the finite-world theory. The change-of-measure identity~\eqref{eq:fw-com} gives every lower bound of
Appendix~\ref{app:foundations} (Theorem~\ref{thm:fw-cost}(ii), the Goodhart gap of Theorem~\ref{thm:fw-goodhart}(b), and the
channel comparisons of Proposition~\ref{prop:fw-cc}). The convex geometry of Proposition~\ref{prop:fw-geometry} gives every
certificate (Proposition~\ref{prop:fw-deploy}, Theorem~\ref{thm:fw-conf}, and Corollary~\ref{cor:fw-integration}). For the
hidden-event class, the orthonormality of parity characters is the common fact: it yields the oracle of
Lemma~\ref{lem:oracle}, the lead test of Theorem~\ref{thm:upper}, the single informative coefficient in
Theorem~\ref{thm:lower}, and the gradient signals of Proposition~\ref{prop:gradient}. Corollary~\ref{cor:separation} combines
Theorems~\ref{thm:upper} and~\ref{thm:lower}, and Proposition~\ref{prop:anchor} is a special case of the equal-laws
argument that also underlies Theorem~\ref{thm:fw-cost}.

\subsection*{Where each result is stated and proved}
\newcommand{\resrow}[4]{#1~\ref{#2} & #3 & \pageref{#2} & \pageref{pf:#2} & #4\\}
\begingroup\small
\setlength{\tabcolsep}{4pt}
\begin{longtable}{@{}p{1.55cm}>{\raggedright\arraybackslash}p{3.6cm}p{1.0cm}p{0.9cm}>{\raggedright\arraybackslash}p{5.6cm}@{}}
\toprule
Result & Content & Stated & Proof & Main tools \\
\midrule
\endfirsthead
\toprule
Result & Content & Stated & Proof & Main tools \\
\midrule
\endhead
\bottomrule
\endfoot
\multicolumn{5}{@{}l}{\emph{Appendix~\ref{app:foundations}: finite-world foundations (Section~\ref{sec:foundations})}}\\
\resrow{Prop.}{prop:fw-geometry}{geometry of the decision complex}{Helly's theorem; minimax theorem; Carath\'eodory's theorem}
\resrow{Lemma}{lem:fw-rate}{exclusion rates}{linear-programming form of a max--min ratio}
\resrow{Thm.}{thm:fw-cost}{finite-world certification cost}{change of measure; Wald's identity; Azuma--Hoeffding; Ville's inequality; tracking of a design}
\resrow{Thm.}{thm:fw-goodhart}{Goodhart gap of proxy-only learning}{Cauchy--Schwarz in $L^2(d_{\rm ref})$; two-point construction}
\resrow{Lemma}{lem:fw-dr}{verified signal}{conditional expectation over the verification coin}
\resrow{Prop.}{prop:fw-cc}{one family, three channels}{Bayes' rule; explicit exclusion rates; data processing}
\resrow{Prop.}{prop:fw-deploy}{deployment certificates}{sequence form; Kuhn's theorem; minimax theorem; Carath\'eodory's theorem}
\resrow{Thm.}{thm:fw-conf}{confidence-set deployment bound}{Hoeffding's inequality; union bound; decision-contrast modulus}
\resrow{Cor.}{cor:fw-integration}{all channels in one certificate}{Theorem~\ref{thm:fw-cost}; Propositions~\ref{prop:fw-geometry} and~\ref{prop:fw-deploy}}
\resrow{Prop.}{prop:fw-consequences}{hedging, memory, and monitors}{proof of Theorem~\ref{thm:fw-cost}; Blackwell ordering; data processing}
\midrule
\multicolumn{5}{@{}l}{\emph{Appendix~\ref{app:proofs}: boundaries, hidden-event class, and deep RL (Sections~\ref{sec:model}, \ref{sec:class}--\ref{sec:deep})}}\\
\resrow{Prop.}{prop:anchor}{no anchor without verification}{equal transcript laws; $J_M+J_{\bar M}\le1$}
\resrow{Lemma}{lem:oracle}{oracle structure and quotients}{orthonormality of parity characters; conditioning on replies}
\resrow{Thm.}{thm:upper}{certified decision state}{Hoeffding's inequality; union bound over the checking schedule}
\resrow{Thm.}{thm:lower}{information-per-price lower bound}{change of measure to a reference world; Bessel's inequality; Fano's inequality}
\resrow{Cor.}{cor:separation}{exponential separation and crossover price}{Theorems~\ref{thm:upper} and~\ref{thm:lower}}
\resrow{Prop.}{prop:gradient}{gradient signal of the two channels}{differentiation at $w=0$; Hoeffding's inequality}
\resrow{Prop.}{prop:dr}{unbiased verified return}{conditional expectation over the verification coin}
\resrow{Prop.}{prop:cert}{anytime certificates}{empirical-Bernstein supermartingale; Ville's inequality}
\end{longtable}
\endgroup

\subsection*{Notation}
\begingroup\small
\setlength{\tabcolsep}{4pt}
\begin{longtable}{@{}>{\raggedright\arraybackslash}p{3.1cm}>{\raggedright\arraybackslash}p{8.8cm}p{1.6cm}@{}}
\toprule
Symbol & Meaning & Defined \\
\midrule
\endfirsthead
\toprule
Symbol & Meaning & Defined \\
\midrule
\endhead
\bottomrule
\endfoot
\multicolumn{3}{@{}l}{\emph{The decision-observable contract}}\\
$o_t,\ h_t,\ u_t=(q_t,a_t)$ & report, report history, joint choice of audit and physical action & Def.~\ref{def:hedp}\\
$Y,\ \tilde Y,\ Y^{\rm ver},\ \eta$ & task outcome, free proxy, verifier label, verifier flip probability & Def.~\ref{def:hedp}\\
$c_0,\ c(q),\ c_a,\ c_v,\ \kappa$ & training prices of an episode, an audit, a verification; deployment audit price & \eqref{eq:charge}\\
$C$ & training charge, the sum of all prices paid & \eqref{eq:charge}\\
$B,\ \Pi_B$ & deployment audit budget; report-history policies that respect it & \S\ref{sec:model}\\
$J_M,\ V^\star_{M,B},\ \Reg_M$ & value, optimal value, and regret in world $M$ & \S\ref{sec:model}\\
$\pihat,\ \Rhat,\ \delta$ & deployed policy, regret certificate, failure probability & Def.~\ref{def:dorl}\\
\midrule
\multicolumn{3}{@{}l}{\emph{Finite-world foundations}}\\
$\Theta,\ \theta,\ M_\theta$ & finite set of candidate worlds and the process indexed by $\theta$ & \S\ref{sec:fw-contract}\\
$e,\ \cE,\ P^e_\theta,\ k_e,\ D_e$ & training experiment, menu, transcript law, charge, and KL divergence & \S\ref{sec:fw-contract}\\
$X=\Delta(\Pi),\ d_X$ & randomized deployment class; affine dimension of its payoff set & \S\ref{sec:fw-contract}\\
$R(S;X),\ G_\nu(\varepsilon)$ & certification radius of a world set; $\varepsilon$-good policies of world $\nu$ & \eqref{eq:fw-radius}\\
$\cK_\varepsilon(X)$ & decision complex: world sets that one policy serves within $\varepsilon$ & Def.~\ref{def:fw-complex}\\
$\Gamma_\theta,\ \cT_\varepsilon(\theta)$ & charge-normalized exclusion rate; characteristic cost & \eqref{eq:fw-rate}, \eqref{eq:fw-cost}\\
\midrule
\multicolumn{3}{@{}l}{\emph{Sparse hidden-event class}}\\
$L,\ k,\ s,\ \gamma$ & number of reports, of hidden events, support bound, reliability lower bound & Def.~\ref{def:class}\\
$X_\ell,\ \chi_S$ & report bits; parity $\prod_{\ell\in S}X_\ell$ & Def.~\ref{def:class}\\
$E_j,\ S_j,\ N_j,\ \mu_j,\ \sigma$ & hidden event, its cue set, noise sign, reliability; outcome polarity & Def.~\ref{def:class}\\
$T_A,\ T^\star,\ v(A)$ & parity of the unaudited cues; $\bigcup_jS_j$; value of auditing the set $A$ & Lemma~\ref{lem:oracle}\\
$V^\star_0$ & reply-blind optimum $\frac12(1+\prod_j\mu_j)$ & Lemma~\ref{lem:oracle}\\
$\mu_{\min},\ \theta$ & smallest reliability; accuracy threshold of the lead test (Appendix~\ref{app:proofs} only) & Thm.~\ref{thm:upper}\\
$\iota(D),\ \bar\rho$ & information per unit price of a training design; its maximum & Thm.~\ref{thm:lower}\\
\midrule
\multicolumn{3}{@{}l}{\emph{DORL-PPO and certificates}}\\
$\rho,\ \rho_a,\ \rho'$ & verification rate, side-audit rate, charge-matched verification rate of baselines & \S\ref{sec:deep}\\
$g,\ R,\ \beta_j$ & outcome model, doubly-robust return, event-belief logits & Alg.~\ref{alg:dorlppo}\\
$\mathrm{LCB}_n,\ \alpha$ & anytime lower confidence bound on the deployed value; its error level & Prop.~\ref{prop:cert}\\
\end{longtable}
\endgroup
\section{Proofs for Section~\ref{sec:foundations}}\label{app:foundations}

\begin{roadmap}[: how Appendix~\ref{app:foundations} is organized]
Every result of Section~\ref{sec:foundations} is proved here with two tools: the change-of-measure identity
\eqref{eq:fw-com} for lower bounds, and the convex geometry of the deployment class (Helly's theorem, the minimax
theorem, Carath\'eodory's theorem) for certificates. Appendix~\ref{app:fw-geometry} proves the geometry of the decision
complex, which every later proof uses. Appendices~\ref{app:fw-rate} and~\ref{app:fw-cost} prove the cost theorem: the
exclusion rate is a linear program, its reciprocal lower-bounds every sound learner, and FaceTS attains it. The remaining
subsections treat the three channels in turn (verification, audits, deployment) and the consequences of
Proposition~\ref{prop:fw-consequences}.
\end{roadmap}

\subsection{Notation and two standard facts}\label{app:fw-notation}

Throughout, $a_0=\log(1/p_{\min})$, which we may take positive. Because the candidate laws of an experiment share one
support on which they are at least $p_{\min}$, every log-likelihood ratio $\log(P^e_\theta(y)/P^e_\lambda(y))$ lies in
$[-a_0,a_0]$, and $0\le D_e(\theta,\lambda)\le a_0$. A learner's information after $n$ episodes is
$\cF_n=\sigma(\xi,e_1,Y_1,\dots,e_n,Y_n)$, where $\xi$ is its internal randomness, whose law does not depend on the
world; later internal randomization can be absorbed into $\xi$. Experiment choices are $\cF_{i-1}$-measurable, $\tau$
is an $(\cF_n)$-stopping time, and $\hat x$ is $\cF_\tau$-measurable. We write
$L_n(\nu)=\sum_{i\le n}\log P^{e_i}_\nu(Y_i)$, $\Lambda_n(\lambda)=L_n(\theta)-L_n(\lambda)$, $N_e(n)$ for the number
of uses of $e$ among the first $n$ episodes, and $\hat C_n=\sum_{i\le n}\bar k_\theta(e_i)$.

Two standard facts are used repeatedly. \emph{Change of measure:} for every $E\in\cF_\tau$ with
$E\subseteq\{\tau<\infty\}$ and every $\lambda$,
\begin{equation}\label{eq:fw-com}
\Prb_\lambda(E)=\E_\theta\big[\one_Ee^{-\Lambda_\tau(\lambda)}\big].
\end{equation}
Indeed, on $\{\tau=n\}$ the density of $(\xi,e_1,Y_1,\dots,e_n,Y_n)$ under $\nu$ is the law of $\xi$ times the
world-independent choice kernels times $\prod_{i\le n}P^{e_i}_\nu(Y_i)$, so the ratio of the $\lambda$- and
$\theta$-densities is $e^{-\Lambda_n(\lambda)}$; common supports exclude a singular part, and summing over $n$ gives
\eqref{eq:fw-com}. \emph{Equal laws:} if $P^e_\nu=P^e_\lambda$ for every $e$ in the menu, then by induction over
episodes the joint law of $(\xi,e_1,Y_1,e_2,\dots)$, and hence of $(\tau,\hat x)$, is the same under $\nu$ and
$\lambda$.

\subsection{Proof of Proposition~\ref{prop:fw-geometry} (geometry of the decision complex)}\label{app:fw-geometry}\label{pf:prop:fw-geometry}

\begin{proof}
Each $J_\nu$ is affine on the simplex $X$, so $x\mapsto\max_{\nu\in S}\Reg_\nu(x)$ is continuous and attains its
minimum. Hence $R(S;X)\le\varepsilon$ if and only if some $x$ has $\Reg_\nu(x)\le\varepsilon$ for all $\nu\in S$, which
proves the two descriptions of $\cK_\varepsilon(X)$. Each $G_\nu(\varepsilon)$ is the intersection of $X$ with a closed
half-space, hence compact and convex.

(i) A witness $x$ for $S$ serves every subset of $S$. The sets $G_\nu(\varepsilon)$ grow with $\varepsilon$, and with
$X$ at a fixed comparator. Membership $S\in\cK_\varepsilon(X)$ is the threshold event $R(S;X)\le\varepsilon$, so the
complex is constant on every interval that contains no value $R(S;X)$.

(ii) Regret depends on $x$ only through $p=J(x)\in J(X)$. Let $A$ be the affine hull of $J(X)$, of dimension $d_X$,
and $K_\nu=\{p\in J(X):\bar V_\nu-p_\nu\le\varepsilon\}$, a convex subset of $A$. If $x\in\bigcap_{\nu\in S}G_\nu$,
then $J(x)\in\bigcap_\nu K_\nu$; conversely, if $p\in\bigcap_\nu K_\nu$, any $x$ with $J(x)=p$ lies in every
$G_\nu$. So $S$ is a face if and only if $\bigcap_{\nu\in S}K_\nu\ne\varnothing$. Let $S$ be a conflict set with
$|S|\ge d_X+2$. All its subsets of size $d_X+1$ are faces, so every $d_X+1$ of the sets $K_\nu$ intersect, and Helly's
theorem in $A\cong\R^{d_X}$ \citep{helly1923,barvinok2002} gives $\bigcap_{\nu\in S}K_\nu\ne\varnothing$, a
contradiction. For the formula, monotonicity gives $R(S;X)\le R(C;X)$ for $S\subseteq C$. Conversely, for each real
$\varepsilon<R(C;X)$ the set $C$ is not in $\cK_\varepsilon(X)$ (the argument above applies verbatim to every real
$\varepsilon$); since $R(\varnothing;X)=-\infty$, the empty set is a face, so $C$ contains a conflict set
$S_\varepsilon$, which has at most $d_X+1$ worlds and satisfies $R(S_\varepsilon;X)>\varepsilon$. Only finitely many subsets exist, so one subset
$S$ occurs for a sequence $\varepsilon\uparrow R(C;X)$, and $R(S;X)\ge R(C;X)$.

(iii) The function $(x,\mu)\mapsto\sum_\nu\mu_\nu\Reg_\nu(x)$ is affine in each argument on the compact convex sets $X$
and $\Delta(C)$. The minimax theorem \citep{sion1958} gives
$R(C;X)=\min_x\max_\mu\sum_\nu\mu_\nu\Reg_\nu(x)=\max_\mu\min_x\sum_\nu\mu_\nu\Reg_\nu(x)$, which is the displayed
formula; the outer maximum is attained by continuity of the inner minimum in $\mu$. By (ii) choose $S\subseteq C$ with $|S|\le d_X+1$ and
$R(S;X)=R(C;X)$, and apply the formula to $S$: some $\mu_S\in\Delta(S)\subseteq\Delta(C)$ has value $R(C;X)$, so it is
optimal for $C$. With the default comparator every regret is nonnegative, so a minimizer of the attained minimum has
zero maximal regret if and only if it is optimal in every world of $C$.

(iv) Suppose every conflict set has at most two worlds, which holds when $d_X\le1$ by (ii). A non-face $S$ contains a
conflict set, hence a non-face of size at most two. The converse direction is downward closure. In the example,
$X=\Delta(\{1,2,3\})$ and $\Reg_\nu(x)=1-(1-x_\nu)=x_\nu$. The pair $\{\nu,\nu'\}$ is served exactly by the third
action, while $R(\Theta;X)=\min_x\max_\nu x_\nu=1/3$; the payoff set is the image of a two-dimensional simplex under an
injective affine map, so $d_X=2$. If the worlds are training-equivalent, the law of $(\tau,\hat x)$ is the same in all
three. Since $\sum_\nu\Reg_\nu(\hat x)=1$, at least one world has $\Reg_\nu(\hat x)\ge1/3$ on every stopped path, so
$\sum_\nu\Prb_\nu\{\tau<\infty,\ \Reg_\nu(\hat x)\ge1/3\}\ge\Prb\{\tau<\infty\}=1$, and some world has probability at
least $1/3$.
\end{proof}

\subsection{Proof of Lemma~\ref{lem:fw-rate} (exclusion rates)}\label{app:fw-rate}\label{pf:lem:fw-rate}

\begin{proof}
(i) For a design $w$, set $x_e=w_e/\sum_fw_f\bar k_\theta(f)$; then $\sum_e\bar k_\theta(e)x_e=1$ and
$\min_{\lambda\in A}\sum_eD_e(\theta,\lambda)x_e$ is the ratio in~\eqref{eq:fw-rate}. Conversely, a feasible $x$ is
nonzero because every $\bar k_\theta(e)\ge c_0>0$, and $w=x/\sum_ex_e$ has ratio $\min_\lambda\sum_eD_e(\theta,\lambda)x_e$.
The two problems have the same values. The maximum in~\eqref{eq:fw-rate} is attained because the ratio is continuous
on the compact simplex.
(ii) Numerators are at most $a_0$ and denominators at least $c_0$. If $\lambda\in A\cap[\theta]$, every
$D_e(\theta,\lambda)=0$ and every design has ratio zero. If $A\cap[\theta]=\varnothing$, every $\lambda\in A$ has some
$e$ with $P^e_\lambda\ne P^e_\theta$, so the uniform design has a positive numerator for each of the finitely many
$\lambda\in A$.
(iii) If $\nu\sim\theta$, then $D_e(\nu,\cdot)=D_e(\theta,\cdot)$ and $\bar k_\nu=\bar k_\theta$, because the charge is
a function of the transcript.
\end{proof}

\subsection{Proof of Theorem~\ref{thm:fw-cost} (finite-world certification cost)}\label{app:fw-cost}\label{pf:thm:fw-cost}

\paragraph{(i) Regimes.} By Lemma~\ref{lem:fw-rate}(ii), $\Gamma_\theta(\Theta\setminus S)=\infty$ exactly when
$S=\Theta$, so $\cT_\varepsilon(\theta)=0$ if and only if $\Theta\in\cK_\varepsilon(X)$, that is,
$\varepsilon\ge\varepsilon_{\rm free}$. Otherwise every candidate rate is finite, and $\Gamma^\star_\varepsilon(\theta)>0$
if and only if some face $S\ni\theta$ has $(\Theta\setminus S)\cap[\theta]=\varnothing$, that is, $S\supseteq[\theta]$.
By downward closure such a face exists if and only if $[\theta]$ is a face, that is,
$\varepsilon\ge\varepsilon_{\rm imp}(\theta)$; note $\varepsilon_{\rm imp}\le\varepsilon_{\rm free}$ by monotonicity.
If $\nu\sim\theta$, every face of positive rate for $\theta$ contains $[\theta]\ni\nu$, and Lemma~\ref{lem:fw-rate}(iii)
gives the same rates, so $\Gamma^\star_\varepsilon(\nu)=\Gamma^\star_\varepsilon(\theta)$ (both vanish together).
Finally, $\cT_\varepsilon(\theta)$ depends on $\varepsilon$ only through $\cK_\varepsilon(X)$, which is nondecreasing in
$\varepsilon$ and constant between consecutive values $R(S;X)$ by Proposition~\ref{prop:fw-geometry}(i).

\paragraph{(ii) Lower bound.} Fix a sound learner that stops $\Prb_\theta$-a.s. If $\E_\theta[C]=\infty$ there is
nothing to prove. Otherwise $C\ge c_0\tau$ gives $\E_\theta\tau<\infty$. The process
$\sum_{i\le n}(k_{e_i}(Y_i)-\bar k_\theta(e_i))$ is a martingale with increments bounded by $\bar c$, so optional
stopping \citep{wald1944} gives $\E_\theta[C]=\E_\theta[\hat C_\tau]$. If $\cT_\varepsilon(\theta)=0$ the claim is
trivial, so assume $\Theta\notin\cK_\varepsilon(X)$, and hence $\Gamma^\star:=\Gamma^\star_\varepsilon(\theta)<\infty$.
Fix a finite $T<\cT_\varepsilon(\theta)$, put $\rho=1-T\Gamma^\star\in(0,1]$, $\ell=\log(1/\delta)$, and
$t_\ell=\lfloor T\ell/c_0\rfloor+1$. We show
\begin{equation}\label{eq:fw-hp}
\Prb_\theta\{\hat C_\tau\le T\ell\}\ \le\ \delta+|\Theta|e^{-\rho\ell/2}
+|\Theta|\exp\Big(-\frac{\rho^2\ell^2}{32a_0^2(T\ell/c_0+1)}\Big).
\end{equation}

Let $B=\{\tau<\infty,\ \hat C_\tau\le T\ell,\ \Reg_\theta(\hat x)\le\varepsilon\}$. Since $\hat C_n\ge c_0n$, $B$
forces $\tau<t_\ell$. On $B$ the set $S(\hat x)=\{\nu:\Reg_\nu(\hat x)\le\varepsilon\}$ is a face (with witness
$\hat x$) containing $\theta$, and $S(\hat x)\ne\Theta$. If $\tau\ge1$, the design $w=N(\tau)/\tau$ satisfies
$\min_{\lambda\notin S(\hat x)}\langle w,D(\theta,\lambda)\rangle/\langle w,\bar k_\theta\rangle\le
\Gamma_\theta(\Theta\setminus S(\hat x))\le\Gamma^\star$. Multiplying by $\tau\langle w,\bar k_\theta\rangle=\hat C_\tau$,
some $\lambda$ with $\Reg_\lambda(\hat x)>\varepsilon$ has
\begin{equation}\label{eq:fw-lowinfo}
\textstyle\sum_eN_e(\tau)D_e(\theta,\lambda)\le\Gamma^\star\hat C_\tau\le(1-\rho)\ell,
\end{equation}
which holds trivially if $\tau=0$. For each $\lambda$, the process
$M_n(\lambda)=\Lambda_{n\wedge\tau}(\lambda)-\sum_{i\le n\wedge\tau}D_{e_i}(\theta,\lambda)$ is a $\Prb_\theta$-martingale
with increments in $[-2a_0,a_0]$. The maximal Azuma--Hoeffding inequality \citep{azuma1967,boucheron2013} gives
$\Prb_\theta\{\max_{n\le t_\ell}M_n(\lambda)\ge\rho\ell/2\}\le\exp(-\rho^2\ell^2/(32a_0^2t_\ell))$. Let
$\mathrm{Dev}$ be the union of these events over $\lambda\ne\theta$; its probability is at most the last term
of~\eqref{eq:fw-hp}. Outside $\mathrm{Dev}$, on $B$, \eqref{eq:fw-lowinfo} gives
$\Lambda_\tau(\lambda)<(1-\rho/2)\ell$. Let
$B_\lambda=B\cap\mathrm{Dev}^c\cap\{\Reg_\lambda(\hat x)>\varepsilon,\ \Lambda_\tau(\lambda)\le(1-\rho/2)\ell\}\in\cF_\tau$.
By soundness at $\lambda$ and~\eqref{eq:fw-com},
\[
\delta\ \ge\ \Prb_\lambda\{\tau<\infty,\ \Reg_\lambda(\hat x)>\varepsilon\}\ \ge\ \Prb_\lambda(B_\lambda)
=\E_\theta\big[\one_{B_\lambda}e^{-\Lambda_\tau(\lambda)}\big]\ \ge\ e^{-(1-\rho/2)\ell}\,\Prb_\theta(B_\lambda),
\]
so $\Prb_\theta(B_\lambda)\le e^{-\rho\ell/2}$. Since $B\setminus\mathrm{Dev}\subseteq\bigcup_\lambda B_\lambda$ and
$\Prb_\theta\{\tau<\infty,\Reg_\theta(\hat x)>\varepsilon\}\le\delta$, \eqref{eq:fw-hp} follows. Its right side tends
to $0$ as $\delta\downarrow0$, so $\E_\theta[\hat C_\tau]\ge T\ell(1-o(1))$. Letting $T\uparrow\cT_\varepsilon(\theta)$,
or $T\to\infty$ if $\cT_\varepsilon(\theta)=\infty$, proves the claim. Randomized experiment choices and outputs enter
only through $\xi$ and the choice kernels, which cancel in~\eqref{eq:fw-com}.

\paragraph{(iii) Soundness of FaceTS.} If $R(\Theta;X)\le\varepsilon$, FaceTS stops at $n=0$ with a policy whose regret
is at most $\varepsilon$ in every world. Otherwise fix any true world $\nu$. For $v\ne\nu$, $\exp(L_n(v)-L_n(\nu))$ is a
nonnegative $\Prb_\nu$-martingale with mean one, because each choice is predictable and
$\sum_yP^e_v(y)=1$ on the common support. Ville's inequality \citep{ville1939} and a union bound over the $|\Theta|-1$
worlds $v\ne\nu$ give $\Prb_\nu\{\exists n:\nu\notin C_n\}\le\delta$. On the complement, if FaceTS stops at $\tau$,
then $\Reg_\nu(\hat x)\le\max_{\nu'\in C_\tau}\Reg_{\nu'}(\hat x)=R(C_\tau;X)=\Rhat\le\varepsilon$. This proves both
soundness statements and the anytime coverage used in Corollary~\ref{cor:fw-integration}.

\paragraph{(iii) Cost of FaceTS.} Assume $0<\cT_\varepsilon(\theta)<\infty$ (if it is $0$, FaceTS stops at $n=0$), so
$\Theta\notin\cK_\varepsilon(X)$ and $[\theta]\in\cK_\varepsilon(X)$. Two observations drive the argument. First, if
$S$ is a face and $\Lambda_n(\lambda)>\beta$ for every $\lambda\notin S$, then $\max_vL_n(v)-L_n(\lambda)\ge
\Lambda_n(\lambda)>\beta$, so $C_n\subseteq S$ and $R(C_n;X)\le R(S;X)\le\varepsilon$: FaceTS has stopped by episode
$n$. Second, conditional on the pilot, the post-pilot pairs $(e_j,Y_j)$ are i.i.d.: $e_j$ is drawn from the fixed
mixture $w_\ell=(1-\epsilon_\ell)\tilde w+\epsilon_\ell u$, where $u$ is uniform. For $\lambda\in\Theta$ the increments
$X_j(\lambda)=\log(P^{e_j}_\theta(Y_j)/P^{e_j}_\lambda(Y_j))\in[-a_0,a_0]$ then have mean
$\langle w_\ell,D(\theta,\lambda)\rangle$, and the pilot contributes at least $-a_0n_0$ to each $\Lambda_n(\lambda)$.
Write $h_\ell=\beta+a_0n_0$ and let $N$ be the number of post-pilot episodes before stopping.

\emph{Hoeffding tail.} Let $A\subseteq\Theta$ and suppose every $\lambda\in A$ has post-pilot mean at least $\bar\mu>0$,
and FaceTS stops once $\sum_{j\le t}X_j(\lambda)>h_\ell$ for all $\lambda\in A$. For $\zeta>0$ and
$t\ge t_0=(1+\zeta)h_\ell/\bar\mu$ we have $t\bar\mu-h_\ell\ge t\bar\mu\zeta/(1+\zeta)$, and Hoeffding's inequality
\citep{hoeffding1963} gives
\[
\Prb\{N>t\mid\text{pilot}\}\le\sum_{\lambda\in A}\Prb\Big\{\sum_{j\le t}X_j(\lambda)\le h_\ell\Bigm|\text{pilot}\Big\}
\le|\Theta|e^{-tb},\qquad b=\frac{\bar\mu^2\zeta^2}{2a_0^2(1+\zeta)^2}.
\]
Hence $\E[N\mid\text{pilot}]\le t_0+1+|\Theta|/(1-e^{-b})\le t_0+2+|\Theta|(1+1/b)$. Since $N$ is a stopping time for
the i.i.d.\ post-pilot sequence, Wald's identity \citep{wald1944} gives post-pilot expected charge
$\E[N\mid\text{pilot}]\langle w_\ell,\bar k_\theta\rangle$.

\emph{Correct pilot.} Let $G=\{\tilde\theta\in[\theta]\}$. For $\lambda\notin[\theta]$, Markov's inequality applied
to $\exp(\frac12(L_{n_0}(\lambda)-L_{n_0}(\theta)))$ and independence across the deterministic pilot give
$\Prb_\theta\{L_{n_0}(\lambda)\ge L_{n_0}(\theta)\}\le\prod_{e}\mathrm{BC}(P^e_\theta,P^e_\lambda)^{n_0/|\cE|}$, where
$\mathrm{BC}(P,Q)=\sum_y\sqrt{P(y)Q(y)}$ is strictly less than one when $P\ne Q$. Thus $\Prb_\theta(G^c)\le
|\Theta|e^{-a\sqrt\ell/|\cE|}$ for a constant $a>0$ that does not depend on $\delta$. On $\{\tilde\theta=\nu\}$ with
$\nu\in[\theta]$, Lemma~\ref{lem:fw-rate}(iii) and part (i) make the face design an optimal pair for $\theta$ with
$\Gamma_\theta(\Theta\setminus\tilde S)=\Gamma^\star_\varepsilon(\theta)>0$, so $\tilde S\supseteq[\theta]\ni\theta$.
Put $D_*=\min_{\lambda\notin\tilde S}\langle\tilde w,D(\theta,\lambda)\rangle$ and $c_*=\langle\tilde w,\bar k_\theta\rangle$;
then $c_*/D_*=\cT_\varepsilon(\theta)$. These constants depend on $\nu$, but $\nu$ ranges over a finite set. Apply the
tail bound with $A=\Theta\setminus\tilde S$ and $\bar\mu=(1-\epsilon_\ell)D_*\ge D_*/2$:
\[
\E[C\mid\text{pilot}]\le n_0\bar c+\Big(\frac{(1+\zeta)h_\ell}{(1-\epsilon_\ell)D_*}+K_\zeta\Big)
\big(c_*+\epsilon_\ell\bar c\big)
=\frac{(1+\zeta)\,\cT_\varepsilon(\theta)\,h_\ell}{1-\epsilon_\ell}+O(\sqrt\ell),
\]
where $K_\zeta$ does not depend on $\delta$, and we used $h_\ell\epsilon_\ell=O(1)$ and $n_0\le\sqrt\ell+|\cE|$.

\emph{Wrong pilot.} On $G^c$ use the uniform component. Every $\lambda\notin[\theta]$ has post-pilot mean at least
$\epsilon_\ell a'$ with $a'=\min_{\lambda\notin[\theta]}\langle u,D(\theta,\lambda)\rangle>0$. Once all of them are
excluded, $C_n\subseteq[\theta]$, a face, and FaceTS stops. The tail bound with $\zeta=1$ gives
$\E[N\mid\text{pilot}]\le2h_\ell/(\epsilon_\ell a')+2+|\Theta|(1+8a_0^2/(\epsilon_\ell a')^2)=O(\ell^2)$, so
$\E[C\mid\text{pilot}]=O(\ell^2)$.

Combining, $\E_\theta[C]\le(1+\zeta)\cT_\varepsilon(\theta)\ell(1+o(1))+O(\sqrt\ell)+O(\ell^2)e^{-a\sqrt\ell/|\cE|}$,
because $h_\ell=\ell+\log(|\Theta|-1)+a_0n_0=\ell(1+o(1))$ and $\epsilon_\ell\to0$. Dividing by $\ell$ and letting
$\zeta\downarrow0$ proves the upper bound. \qed

\paragraph{Reduction to best-policy identification.} Suppose $\varepsilon=0$, the comparator is the default, and each
$\nu$ has a unique optimal deterministic policy $\pi_\nu\in\Pi$. A distribution $x\in\Delta(\Pi)$ attains
$\max_XJ_\nu$ if and only if it is supported on $\arg\max_\Pi J_\nu=\{\pi_\nu\}$, so $G_\nu(0)=\{\delta_{\pi_\nu}\}$.
Hence $S$ is a face if and only if all $\pi_\nu$, $\nu\in S$, coincide; the unique facet containing $\theta$ is
$\{\nu:\pi_\nu=\pi_\theta\}$, and $\cT_0(\theta)=1/\Gamma_\theta(\{\lambda:\pi_\lambda\ne\pi_\theta\})$, the
cost-normalized characteristic time for identifying the optimal policy.

\subsection{Proof of Theorem~\ref{thm:fw-goodhart} (Goodhart gap)}\label{app:fw-verify}\label{pf:thm:fw-goodhart}

Policy independence of $\mu$: under a report-measurable policy whose randomization is independent of the hidden
trajectory $z$, the joint probability of $(h,z)$ factors into the policy's choice probabilities, which depend on $h$
only, and the world's probability of $z$ and of the reports and replies in $h$ given the choices in $h$. The first
factor cancels in $\Prb(z\mid h)$, so $\E[Y\mid h]$ is the same for every policy.

\begin{proof}
Let $e=\mu-f$, so that $(J-J^f)(\pi)=\E_{d^\pi}[e(h)]$ for every $\pi$, because the audit prices $\kappa(h)$ cancel.
Write $w_\pi=\mathrm dd^\pi/\mathrm dd_{\rm ref}$, which exists when the $\chi^2$-divergence is finite, and
$\bar e=\E_{d_{\rm ref}}e$.

(a) Since $\E_{d_{\rm ref}}w_\pi=\E_{d_{\rm ref}}w_{\pi'}=1$, the Cauchy--Schwarz and triangle inequalities give
\[
(J-J^f)(\pi)-(J-J^f)(\pi')=\E_{d_{\rm ref}}[(w_\pi-w_{\pi'})(e-\bar e)]
\le\big(\Vert w_\pi-1\Vert_2+\Vert w_{\pi'}-1\Vert_2\big)\sigma\le2\sigma\sqrt{\mathcal C},
\]
with norms in $L^2(d_{\rm ref})$, and symmetrically. For any $\pi^\star\in\Pi_{\mathcal C}$,
$J(\pi^\star)-J(\pi_f)\le J^f(\pi^\star)-J^f(\pi_f)+2\sigma\sqrt{\mathcal C}\le2\sigma\sqrt{\mathcal C}$; take the
supremum over $\pi^\star$.

(b) Take one-step worlds with no audits. The report is the action $h\in\{0,1\}$, the proxy is $f\equiv\frac12$, and
$Y\mid h\sim\mathrm{Bern}(\mu_\pm(h))$ in world $\pm$. A policy is a law on $\{0,1\}$; the report and proxy laws do not
depend on the world, so by the equal-laws fact a learner without verification has the same output law in both worlds.
If $\mathcal C\le1$, let $d_{\rm ref}=(\frac12,\frac12)$, $t=\sqrt{\mathcal C}/2$, $d_a=(\frac12+t,\frac12-t)$,
$d_b=(\frac12-t,\frac12+t)$, and $\mu_\pm(1)=\frac12\pm\sigma$, $\mu_\pm(0)=\frac12\mp\sigma$. Then
$\chi^2(d_a\Vert d_{\rm ref})=\chi^2(d_b\Vert d_{\rm ref})=4t^2=\mathcal C$, $\operatorname{Var}_{d_{\rm ref}}(e)=\sigma^2$,
and $J(b)-J(a)=\pm4t\sigma=\pm2\sigma\sqrt{\mathcal C}$. If $\mathcal C>1$, let $q=1/(1+\mathcal C)$,
$d_{\rm ref}=d_a=(1-q,q)$, $d_b=(0,1)$, $\mu_\pm(0)=\frac12$, and $\mu_\pm(1)=\frac12\pm s$ with
$s=\sigma\sqrt{1+\mathcal C}\le\frac12$. Then $\chi^2(d_b\Vert d_{\rm ref})=(1-q)/q=\mathcal C$,
$\operatorname{Var}_{d_{\rm ref}}(e)=q(1-q)s^2\le\sigma^2$, and $J(b)-J(a)=\pm s(1-q)=\pm\sigma\mathcal C/\sqrt{1+\mathcal C}$.
In both cases $\gamma\ge\sigma\sqrt{\mathcal C/2}$. If the learner outputs $b$ with probability $\varrho$ in both worlds,
its expected regrets are $\gamma(1-\varrho)$ and $\gamma\varrho$, the larger of which is at least
$\gamma/2\ge\sigma\sqrt{\mathcal C}/(2\sqrt2)$.
\end{proof}

\subsection{Proof of Lemma~\ref{lem:fw-dr} (verified signal)}\label{pf:lem:fw-dr}

\begin{proof}
Let $\mathcal G$ be generated by the past, $h$, $Y$, and $Y^{\rm ver}$. Since $Y^{\rm ver}$ equals $Y$ with probability
$1-\eta$ and $1-Y$ otherwise, $\E[Y^{\rm ver}\mid Y,h]=\eta+(1-2\eta)Y$, so $\E[\check Y\mid\text{past},h,Y]=Y$. The coin
satisfies $\E[A\mid\mathcal G]=p(h)$ and $\E[(A/p-1)^2\mid\mathcal G]=(1-p)/p$, while $p$ and $g$ are measurable with
respect to the past and $h$. Write $\Psi=\check Y+(A/p-1)(\check Y-g)$. Then $\E[\Psi\mid\mathcal G]=\check Y$, so
$\E[\Psi\mid\text{past},h]=\E[Y\mid\text{past},h]=\mu(h)$. For the variance, the cross term
$\E[(\check Y-\E[\check Y\mid\text{past}])(A/p-1)(\check Y-g)\mid\text{past}]$ vanishes after conditioning on
$\mathcal G$, and the square of the second term has conditional mean $\E[(1-p)/p\cdot(\check Y-g)^2\mid\text{past}]$;
conditioning on $h$ gives $m_g(h)$.
\end{proof}

\subsection{Proof of Proposition~\ref{prop:fw-cc} (one family, three channels)}\label{app:fw-cc}\label{pf:prop:fw-cc}

Let $W=(C,U,\xi)$, where $\xi$ is the experiment's internal randomness. Since $X$ is fair and
$\Prb_\theta(C=X\mid X,U)=p_\theta$ is symmetric, $C$ is a fair bit independent of $U$ in both worlds, so $W$ has the
same law in both worlds, and by Bayes' rule $\Prb_\theta(X=C\mid W)=p_\theta$.

\begin{proof}
(i) The proxy value is $\Prb(a=0)$, maximized by $a\equiv0$, whose task value is $\Prb(X=0)=\frac12$. A policy without
audits chooses $a$ from $W$, so $\Prb_\theta(a=X)=sp_\theta+(1-s)(1-p_\theta)$ with $s=\Prb(a=C)$; this is at most
$(1+r)/2$ and attains it with $s=\one\{\theta=+1\}$. An experiment that buys neither an audit nor a verification has a
transcript that is a function of $W$, whose law is world-independent.

(ii) For $B=0$, the computation in (i) gives $J_\theta=\frac12+\theta r(s-\frac12)$ and $\bar V_\theta=(1+r)/2$, so the
regrets are $r(1-s)$ and $rs$, and $R(\Theta;X)=\min_s\max\{r(1-s),rs\}=r/2$. For $B=1$, the policy that audits $j=U$
and plays $a=Z$ has utility $1-\kappa_a$ in both worlds. Any policy earns at most $1-\kappa_a$ on paths with an audit
and, by (i), at most $(1+r)/2\le1-\kappa_a$ in conditional expectation given $W$ on paths without one. So this policy
is optimal in both worlds and $R(\Theta;X)=0$, whence $\cT_\varepsilon=0$ by Theorem~\ref{thm:fw-cost}(i). For $B=0$ and
$\varepsilon<r/2$, the faces are $\varnothing$, $\{+1\}$, and $\{-1\}$. If some experiment separates the worlds, then
$[\theta]=\{\theta\}$ and $\Gamma^\star_\varepsilon(\theta)=\Gamma_\theta(\{-\theta\})$. With one alternative,
$\langle w,D\rangle=\sum_ew_e\bar k_\theta(e)\,(D_e/\bar k_\theta(e))\le\max_e(D_e/\bar k_\theta(e))\langle w,\bar k_\theta\rangle$,
with equality at the best vertex, which proves the formula; if no experiment separates them, both sides are infinite.

(iii) Given $W$, the transcript of the adaptive audit adds $Z=X$, which given $W$ is the indicator $\one\{X=C\}$ up to
a $W$-measurable relabeling, distributed as $\mathrm{Bern}(p_\theta)$; by the chain rule its divergence is $d_r$ in
either direction. A forced static audit on channel $j$ reveals $X$ on $\{U=j\}$, which has probability $1/m$ and is
$W$-measurable, and an independent fair bit otherwise; its divergence is $d_r/m$. Both cost $c_0+c_a$ per episode.
With one alternative, the characteristic cost of a menu is $(\max_eD_e/\bar k_\theta(e))^{-1}$ by (ii), which equals
$m(c_0+c_a)/d_r$ for the forced static menu and $(c_0+c_a)/d_r$ after adding the adaptive audit. A static audit that
skips payment on $\{U\ne j\}$ has the same divergence $d_r/m$ and expected charge $c_0+c_a/m$, giving
$(mc_0+c_a)/d_r$ and the stated factor.

(iv) Consider any experiment, with at most one audit. Its audit decision is a function of $W$. If it audits, the rest of
the transcript is a function of $(W,X,\zeta)$, where $\zeta$ is a fair bit independent of everything (the reply on a
mismatched channel) and the verification decision may depend on the reply. If it does not audit, its verification
decision is a function of $W$, and the rest of the transcript is a function of $W$ alone if it does not verify, and of
$(W,X)$ if it does, because $Y=\one\{a=X\}$ with $a$ a function of $W$. Let $b(W)$ indicate that the experiment audits
or, failing that, verifies. By the chain rule, the data-processing inequality \citep{cover2006}, and
$\KL(\Prb_\theta(X\mid W)\Vert\Prb_{-\theta}(X\mid W))=d_r$, its divergence is at most $d_r\Prb(b(W)=1)$, while its
expected charge is at least $c_0+\min\{c_a,c_v\}\Prb(b(W)=1)$. Since $x\mapsto d_rx/(c_0+c'x)$ is increasing, the ratio
is at most $d_r/(c_0+\min\{c_a,c_v\})$. The verified experiment with $a=C$ observes $Y=\one\{C=X\}$, so it has
divergence $d_r$ and charge $c_0+c_v$; the adaptive audit was treated in (iii). If $c_v\le c_a$, the verified experiment
therefore attains the largest ratio of any experiment, and by (ii) adding it to the forced static menu yields the
smallest characteristic cost $(c_0+c_v)/d_r$, equal to that of the adaptive audit when $c_v=c_a$.
\end{proof}

\subsection{Proof of Proposition~\ref{prop:fw-deploy} (deployment certificates)}\label{app:fw-deploy}\label{pf:prop:fw-deploy}

\begin{proof}
\emph{Sequence form.} On a finite perfect-recall tree, a vector $x\ge0$ with $x_\varnothing=1$ and
$\sum_ax_{\sigma(I)a}=x_{\sigma(I)}$ at every information set defines a behavior policy by
$x_{\sigma(I)a}/x_{\sigma(I)}$ when $x_{\sigma(I)}>0$ (and arbitrarily otherwise), and every behavior policy arises in
this way \citep{koller1996,vonstengel1996}. Removing choices that exceed the remaining budget, which is visible at each
history, imposes the pathwise budget. The probability of each terminal history is the product of the plan entry of its
own-choice sequence and of chance and report probabilities that do not depend on the policy, so
$J_\theta(x)=g_\theta^\top x$, and for mixture worlds $J_\theta(x)=\theta^\top Gx$. The realization plans of the
deterministic policies satisfy the constraints, and by Kuhn's theorem \citep{kuhn1953} every plan is the plan of a
mixture of deterministic policies, so $X_{\rm seq}$ and $\Delta(\Pi)$ have the same payoff set.

(i) Each $\bar V_\nu$ is a linear program over $X_{\rm seq}$, and the epigraph form of $\min_x\max_{\nu\in C}$ is the
stated program, whose numbers of variables and constraints are linear in the tree size and $|C|$.

(ii) For fixed $x$, $\theta\mapsto V^\star(\theta)-\theta^\top Gx=\max_{u\in X_{\rm seq}}\theta^\top G(u-x)$ is a maximum
of linear functions, hence convex, so its maximum over the polytope $C_y$ is attained at a vertex; this gives the vertex
program. Inside the affine hull of $C_y$ the only inequalities are coordinate constraints $\theta'_\xi\ge0$. At a
vertex, the active ones must cut out a point, so some $k$ active coordinates have independent restrictions to the
hull; otherwise a nonzero direction in the hull would preserve all active constraints and both small moves along it
would stay feasible. Choosing such $k$ coordinates determines the vertex, so $N\le\binom{|\Xi|}{k}$. For the dual form,
let $\mu$ range over the distributions on $\{v_1,\dots,v_N\}$. The function
$H(x,\mu)=\sum_i\mu_iV^\star(v_i)-\sum_i\mu_iv_i^\top Gx$ is affine in each argument on compact convex sets, so the
minimax theorem gives $R(C_y;X_{\rm seq})=\max_\mu\min_xH(x,\mu)=\max_\mu[\sum_i\mu_iV^\star(v_i)-V^\star(\sum_i\mu_iv_i)]$.
A finitely supported measure on all of $C_y$ cannot do better: writing a nonvertex point as a convex combination of
vertices and moving its mass accordingly preserves the mean and, by convexity of $V^\star$, does not decrease
$\E_\mu V^\star$. The objective depends on $\mu$ only through $(\sum_i\mu_iv_i,\sum_i\mu_iV^\star(v_i))$, a point in
the convex hull of $N$ points of an affine space of dimension at most $k+1$; by Carath\'eodory's theorem
\citep{barvinok2002} it is a convex combination of at most $k+2$ of them, which preserves the objective.
\end{proof}

\subsection{A finite-data deployment bound}\label{app:fw-conf}\label{pf:thm:fw-conf}

Suppose each experiment $e$ of a fixed training menu is repeated $n_e\ge1$ times
independently, the stacked vector $y=S\theta$ has $N_y$ coordinates, $\hat y$ is the vector of empirical frequencies,
$n_{\min}=\min_en_e$, $\varrho_n=\sqrt{\log(2N_y/\alpha)/(2n_{\min})}$, and
$C_n=\{\theta'\in\Delta(\Xi):\Vert S\theta'-\hat y\Vert_\infty\le\varrho_n\}$. With $C=C_{S\theta}$ the population
compatible set, define the \emph{decision-contrast modulus}
\[
\omega_C(t)=\sup_{\substack{\theta'\in\Delta(\Xi)\\ \Vert S\theta'-S\theta\Vert_\infty\le t}}\
\inf_{\bar\theta\in C}\ \sup_{u,x\in X_{\rm seq}}
\big|[J_{\theta'}(u)-J_{\theta'}(x)]-[J_{\bar\theta}(u)-J_{\bar\theta}(x)]\big|.
\]

\begin{theorem}[Confidence-set deployment bound]\label{thm:fw-conf}
Let $\hat x_n$ minimize $R(C_n;X_{\rm seq})$. With probability at least $1-\alpha$,
\[
\Reg_\theta(\hat x_n)\ \le\ R(C_n;X_{\rm seq})\ \le\ R(C;X_{\rm seq})+\omega_C(2\varrho_n).
\]
\end{theorem}

The modulus can vanish along directions in which values are unidentified but the ranking of policies is unaffected.
The bound is an upper bound, not a matching rate, and the inequality-defined polytope $C_n$ does not inherit the vertex
count of Proposition~\ref{prop:fw-deploy}(ii).

\begin{proof}
Each coordinate of $\hat y$ is an average of $n_e\ge n_{\min}$ independent indicators with mean $y_j=(S\theta)_j$, so
Hoeffding's inequality gives $\Prb\{|\hat y_j-y_j|>\varrho_n\}\le2e^{-2n_{\min}\varrho_n^2}=\alpha/N_y$. On the union-bound
event $\{\Vert\hat y-S\theta\Vert_\infty\le\varrho_n\}$, of probability at least $1-\alpha$, the truth lies in $C_n$, every
$\theta'\in C_n$ satisfies $\Vert S\theta'-S\theta\Vert_\infty\le2\varrho_n$, and $C\subseteq C_n$. Fix $x$ and
$\theta'\in C_n$, and for $\gamma>0$ let $\bar\theta\in C$ be within $\gamma$ of the infimum in the definition of
$\omega_C(2\varrho_n)$. Then
\begin{align*}
V^\star(\theta')-J_{\theta'}(x)&=\sup_u[J_{\theta'}(u)-J_{\theta'}(x)]
\le\sup_u[J_{\bar\theta}(u)-J_{\bar\theta}(x)]+\omega_C(2\varrho_n)+\gamma\\
&=V^\star(\bar\theta)-J_{\bar\theta}(x)+\omega_C(2\varrho_n)+\gamma
\le\max_{\theta''\in C}\Reg_{\theta''}(x)+\omega_C(2\varrho_n)+\gamma.
\end{align*}
Letting $\gamma\downarrow0$, maximizing over $\theta'\in C_n$, and minimizing over $x$ gives
$R(C_n)\le R(C)+\omega_C(2\varrho_n)$. Because the truth lies in $C_n$, the minimizer $\hat x_n$ has regret at most
$R(C_n)$.
\end{proof}

\subsection{Proof of Corollary~\ref{cor:fw-integration} (all channels in one certificate)}\label{pf:cor:fw-integration}

\begin{proof}
A proxy-only episode, a report-adaptive audit policy, and a terminal verification rule are each a one-episode protocol
whose transcript has a known law on a finite set and whose charge is a function of the transcript, so they are
experiments in the sense of Section~\ref{sec:fw-contract}. The deployment class of an explicit tree is $\Delta(\Pi)$
with $\Pi$ the deterministic policies, and by Kuhn's theorem each $x\in\Delta(\Pi)$ is implemented by a policy of
$\Pi_B$ with the same values. Theorem~\ref{thm:fw-cost}(ii)--(iii) gives the cost statements, the soundness proof gives
anytime coverage, Proposition~\ref{prop:fw-deploy}(i) computes the minimax policy of a finite $C_n$, and
Proposition~\ref{prop:fw-geometry}(iii) gives the witness. Proxy-construction labels are not part of any experiment
and are charged separately.
\end{proof}

\subsection{Proof of Proposition~\ref{prop:fw-consequences} (hedging, memory, and monitors)}\label{app:fw-cons}\label{pf:prop:fw-consequences}

\begin{proof}
\emph{Deterministic outputs.} The complex $\cK_\varepsilon(\Pi)$ is closed under subsets and is nondecreasing in
$\varepsilon$. These are the only properties of the complex used in the proof of Theorem~\ref{thm:fw-cost}: in the
lower bound $S(\hat x)$ is a face of $\cK_\varepsilon(\Pi)$ when $\hat x\in\Pi$, and FaceTS with
$\min_{\pi\in\Pi}\max_{\nu\in C_n}\Reg_\nu(\pi)$ in place of $R(C_n;X)$ is analyzed verbatim. Hence
$\cT^{\rm det}_\varepsilon$ is the sharp constant for deterministic outputs.

(i) Pure policies are points of $X$. If $\bar V_\nu>\max_\Pi J_\nu$ then $G_\nu(0)=\varnothing$; otherwise
$G_\nu(0)=\Delta(A_\nu)$ with $A_\nu=\arg\max_{\pi\in\Pi}J_\nu(\pi)$, because $J_\nu(x)$ is an average of
$J_\nu(\pi)$ over the support of $x$. So $\bigcap_{\nu\in S}G_\nu(0)=\Delta(\bigcap_\nu A_\nu)$ is nonempty exactly
when some pure policy serves $S$. For the separation, fix $\varepsilon\in(0,\frac12]$ and $g\in(\varepsilon,2\varepsilon]$,
and let two worlds have actions $\alpha,\beta$ with values $(g,0)$ and $(0,g)$ and default comparators $g$. The
half--half mixture has regret $g/2\le\varepsilon$ in both worlds, so $\Theta\in\cK_\varepsilon(X)$ and
$\cT_\varepsilon=0$. Each pure action has regret $g>\varepsilon$ in one world, so the facets of $\cK_\varepsilon(\Pi)$
are the singletons and $\cT^{\rm det}_\varepsilon(\theta)=1/\Gamma_\theta(\{\lambda\})$. With a menu consisting of one
experiment that returns $\mathrm{Bern}(p)$ or $\mathrm{Bern}(q)$ in the two worlds at charge $c_0$, this equals
$c_0/\KL(\mathrm{Bern}(p)\Vert\mathrm{Bern}(q))$, which takes every value in $(0,\infty)$ as $p,q$ range over $(0,1)$,
and is $+\infty$ when $p=q$.

(ii) The inclusion of complexes is Proposition~\ref{prop:fw-geometry}(i), and a larger complex enlarges the set of
candidate faces in~\eqref{eq:fw-cost}; the thresholds $R(\Theta;\cdot)$ and $R([\theta];\cdot)$ decrease because the
same comparator is minimized over a larger class. If the complexes coincide, so do the costs. Conversely, suppose
$S\in\cK_\varepsilon(X')\setminus\cK_\varepsilon(X)$; then $S\ne\varnothing$. Index $\Theta$ by $1,\dots,|\Theta|$ and
take one experiment with alphabet $\{0,1,\dots,|\Theta|\}$, charge $c_0$, and
$P_\nu(y)=(1-\alpha)\one\{y=\gamma(\nu)\}+\alpha/(|\Theta|+1)$ for some $\alpha\in(0,1)$, where $\gamma(\nu)=0$ for
$\nu\in S$ and $\gamma(\nu)$ is the index of $\nu$ otherwise. Then $[\theta]=S$ for $\theta\in S$, and
Theorem~\ref{thm:fw-cost}(i) gives $\cT_\varepsilon(\theta;X')<\infty=\cT_\varepsilon(\theta;X)$.

(iii) The first sentence holds by the definitions~\eqref{eq:fw-radius} and~\eqref{eq:fw-rate}. If
$P^e_\nu=Q\circ P^{e'}_\nu$, the data-processing inequality gives $D_e(\theta,\lambda)\le D_{e'}(\theta,\lambda)$ for all
$\lambda$. Moving the mass of $e$ in any design onto $e'$ does not decrease any numerator in~\eqref{eq:fw-rate} and does
not increase the denominator, so the optimum over the menu without $e$ is at least the optimum with it; the reverse
inequality is trivial. For the last claim, let $\nu\sim\theta$. The experiment that executes $\pi$ and verifies has the
same transcript law under $\nu$ and $\theta$. Its transcript determines the audit prices of the episode and, through
$\E[Y^{\rm ver}]=\eta+(1-2\eta)\E Y$, the mean outcome, so $J_\nu(\pi)=J_\theta(\pi)$ for all $\pi\in\Pi$, hence
$J_\nu=J_\theta$ on $X$ and $\bar V_\nu=\bar V_\theta$. A policy optimal for $\theta$ is then optimal in all of $[\theta]$,
so $R([\theta];X)=0\le\varepsilon$ and Theorem~\ref{thm:fw-cost}(i) applies.

(iv) Fix $\rho\in(0,1)$ and $\eta'\in(0,1)$. In world $\pm$, draw independent hidden bits
$X_1\sim\mathrm{Bern}((1\pm\eta')/2)$, $\Psi\sim\mathrm{Bern}(\rho)$, and $N\sim\mathrm{Bern}(1/2)$, and score a final
action $a\in\{0,1\}$ by $Y=\one\{a=X_1\}$ if $\Psi=1$ and $Y=\one\{a=N\}$ otherwise; the optimal actions differ between
the worlds. An in-episode monitor revealing $X_1$ has divergence $f(\eta')$, where
$f(x)=x\log\frac{1+x}{1-x}$. For any fixed action, the verified outcome is
$\mathrm{Bern}((1\pm\rho\eta')/2)$, with divergence $f(\rho\eta')$. Both are positive. Since
$\log\frac{1+x}{1-x}=2\sum_{j\ge0}x^{2j+1}/(2j+1)$, the function $f(x)/x^2$ is increasing on $(0,1)$, so
$f(\eta')/f(\rho\eta')\ge\rho^{-2}$. At equal prices the information per unit charge has the same ratio.
\end{proof}
\section{Proofs for Sections~\ref{sec:model} and~\ref{sec:class}--\ref{sec:deep}}\label{app:proofs}

\begin{roadmap}[: how Appendix~\ref{app:proofs} is organized]
The proofs for the hidden-event class rest on one fact: parity characters are orthonormal, so every audit reply and the
action-corrected outcome each have a single nonzero Fourier coefficient on the reports. Lemma~\ref{lem:oracle} uses it to
compute the oracle; Theorem~\ref{thm:upper} turns it into an anytime lead test with Hoeffding's inequality and a union
bound over the checking schedule; Theorem~\ref{thm:lower} combines a change of measure against a reference world with
Bessel's and Fano's inequalities; and Corollary~\ref{cor:separation} compares the two. The last three subsections prove
the deep RL propositions, the certificate by Ville's inequality.
\end{roadmap}

\subsection{Notation and standard facts}\label{app:notation-b}

Throughout, $\langle f,g\rangle=\E[f(X)g(X)]$ under the uniform law on $\{-1,+1\}^L$. The characters
$\{\chi_S\}_{S\subseteq[L]}$ are orthonormal: $\langle\chi_S,\chi_T\rangle=\one\{S=T\}$ \citep{odonnell2014}.
Hoeffding's inequality for $n$ i.i.d.\ variables with values in an interval of length $\ell$ gives
$\Prb\{|\bar Z-\E Z|\ge t\}\le2\exp(-2nt^2/\ell^2)$ \citep{hoeffding1963}.

\subsection{Proof of Proposition~\ref{prop:anchor} (no anchor without verification)}\label{app:boundaries}\label{pf:prop:anchor}

\begin{proof}
A learner that buys no verification observes only reports, the proxy, and audit replies. Their joint law is the
same in $M$ and $\bar M$ under every policy, so by induction over episodes the law of the whole transcript,
and hence of the output $\pihat$, is the same distribution $\mathsf Q$ in both worlds. For any fixed
$\pi\in\Pi_B$, the audit charges have the same law in both worlds and $Y_{\bar M}\le1-Y_M$, so
$J_M(\pi)+J_{\bar M}(\pi)\le1-2\E^\pi\sum_t\kappa(q_t)\le1$. Hence
$\Reg_M(\pi)+\Reg_{\bar M}(\pi)\ge2V^\star-1$, and at least one of the two regrets is at least $V^\star-\frac12$.
Write $\mathcal E_W=\{\pi:\Reg_W(\pi)\ge V^\star-\frac12\}$. Then $\mathcal E_M\cup\mathcal E_{\bar M}=\Pi_B$, so
$\Prb_M\{\pihat\in\mathcal E_M\}+\Prb_{\bar M}\{\pihat\in\mathcal E_{\bar M}\}=\mathsf Q(\mathcal E_M)+\mathsf Q(\mathcal E_{\bar M})\ge1$.
\end{proof}

In $\cM(L,k,s,\gamma)$, changing $\sigma$ to $-\sigma$ complements $Y$ and leaves reports and replies unchanged. By
Lemma~\ref{lem:oracle}(i), $V^\star$ does not depend on $\sigma$, so Proposition~\ref{prop:anchor} applies with
$V^\star-\frac12\ge\frac12\prod_j\mu_j$, provided the proxy law does not depend on $\sigma$.

\subsection{Proof of Lemma~\ref{lem:oracle} (oracle structure and quotients)}\label{pf:lem:oracle}

\begin{proof}
Let $\xi$ denote the policy's internal randomness and $A$ the (random) set of audits bought before the
action. Let $\cF$ be the $\sigma$-field generated by $X$, $\xi$, and all replies observed before the action.
\emph{Claim:} on $\{A=a\}$, the noises $(N_j)_{j\notin a}$ are independent of $\cF$ and have their prior law. Each audit
decision is a function of $X$, $\xi$, and earlier replies, and a reply of event $j$ is a function of $X$ and $N_j$.
By induction over the audit steps, $\{A=a\}$ and $(E_j)_{j\in a}$ are measurable with respect to
$\sigma(X,\xi,(N_j)_{j\in a})$, which is independent of $(N_j)_{j\notin a}$.

(i) On $\{A=a\}$, $\prod_{j}E_j=\prod_{j\in a}E_j\cdot\chi_{T_a}(X)\prod_{j\notin a}N_j$, so by the claim
\[
\Prb(Y=1\mid\cF,\,a_{\rm act})=\tfrac12\Big(1+a_{\rm act}\,\sigma\prod_{j\in a}E_j\,\chi_{T_a}(X)\prod_{j\notin a}\mu_j\Big)
\le\tfrac12\Big(1+\prod_{j\notin a}\mu_j\Big).
\]
Subtracting the audit charges and taking expectations gives $J(\pi)\le\E[v(A)]\le\max_{|a|\le B}v(a)$. The
controller in (i) makes the product inside the parentheses equal to $\prod_{j\notin a}\mu_j$ and attains $v(a)$.

(ii) If audits and actions are measurable with respect to $(X,\xi)$, then $N$ is independent of $(X,\xi,A,a_{\rm act})$
and $\Prb(Y=1\mid X,\xi)=\frac12(1+a_{\rm act}\sigma\chi_{T_\varnothing}(X)\prod_j\mu_j)\le V^\star_0$. Audit charges
are nonnegative.

(iii) If audits and actions are measurable with respect to $\xi$ and replies, then on $\{A=a\}$ with $a\ne[k]$ the
variable $\chi_{T_a}(X)$ depends only on report coordinates in $\bigcup_{j\notin a}S_j$. These coordinates are
disjoint from $\bigcup_{j\in a}S_j$, so $\chi_{T_a}(X)$ is a uniform sign independent of $(\xi,(E_j)_{j\in a},N)$, and the
conditional success probability is $\frac12$. On $\{A=[k]\}$, which requires $B\ge k$, the success probability is at
most $1$ and the charge is $\sum_j\kappa_j$.

(iv) By (i), an optimal policy with audit set $A^\star$ plays $\sigma\chi_{T_{A^\star}}(X)\prod_{j\in A^\star}E_j$, which depends on
the reports only through $\chi_{T_{A^\star}}(X)=\prod_{\ell\le L,\ \ell\in T_{A^\star}}X_\ell$. This product is updated recursively,
one bit per report, and is constant when $T_{A^\star}=\varnothing$. So at most two values always suffice. If no optimal
policy ignores the reports, a one-valued summary cannot support an optimal policy, so two is the minimum number.
For example, if $T_{A^\star}\ne\varnothing$ for every optimal $A^\star$, then by (iii) any policy whose audits and actions
ignore the reports has value at most $\max\{\frac12,\,1-\sum_j\kappa_j\}$. Both terms are below $V^\star_B$: the first
because $\frac12<v(A^\star)$, and the second because $[k]$ is not an optimal audit set. Given $X$, the events are independent with
$\Prb(E_j=e\mid X)=\frac12(1+e\mu_j\chi_{S_j}(X))$. Since $\mu_j>0$, the posterior determines the vector
$(\chi_{S_j}(X))_{j\le k}$. Because the supports are disjoint and nonempty, this vector is uniform on $\{\pm1\}^k$
and takes all $2^k$ values.
\end{proof}

\subsection{Proof of Theorem~\ref{thm:upper} (certified decision state)}\label{app:upper}\label{pf:thm:upper}

\begin{proof}
\emph{Good event.} In a Stage-I episode, $E_j\chi_S(X)\in\{-1,1\}$ and
$\E[E_j\chi_S(X)]=\E[N_j]\langle\chi_{S_j},\chi_S\rangle=\mu_j\one\{S=S_j\}$, because $N_j$ is independent of $X$. The
statistics at check $i$ are means of the first $n_i$ episodes, a deterministic sample size. Hoeffding's inequality
(interval length $2$) and a union bound over the $kM_s$ pairs $(j,S)$ give
$\Prb\{\exists(j,S):|\hat c_{j,i}(S)-\mu_j\one\{S=S_j\}|>t_i\}\le2kM_s\exp(-n_it_i^2/2)=\delta_i$.
A union bound over checks gives total failure probability $\sum_{i\ge0}\delta_i=\frac\delta2\sum_{i\ge0}\frac1{(i+1)(i+2)}=\frac\delta2$.
This is valid for the data-dependent stopping check because it holds simultaneously for all checks. In Stage II,
which uses fresh episodes, $(2Y^{\rm ver}-1)a\prod_jE_j=\sigma\zeta$ with i.i.d.\ signs $\zeta$ of mean at least
$1-2\eta>0$. The majority vote errs with probability at most $\exp(-m(1-2\eta)^2/2)\le\delta/2$, and it is exact if
$\eta=0$. Let $\mathcal G$ be the intersection of these events, so $\Prb(\mathcal G)\ge1-\delta$.

\emph{(a) Soundness on $\mathcal G$.} Suppose the lead test passes for $j$ at check $i$ but $\hat S_j\ne S_j$. Then
$\hat c_j(\hat S_j)\le t_i$ and $\hat c_j(S_j)\ge\mu_j-t_i\ge-t_i$. Since $S_j$ is among the competitors, the lead of
$\hat S_j$ over $\max\{0,\max_{S\ne\hat S_j}\hat c_j(S)\}$ is at most $\hat c_j(\hat S_j)-\hat c_j(S_j)\le2t_i$, a contradiction.
(If $\cS_s$ is a singleton, $\hat S_j=S_j$ trivially.) So every certified support is correct and
$|\hat c_j(\hat S_j)-\mu_j|\le t_i$; clipping to $[0,1]\ni\mu_j$ cannot increase the error. With correct supports and
$\hat\sigma=\sigma$, the deployed controller is exactly the oracle controller for $\hat A$, so $J(\pihat)=v(\hat A)$. If
$B=0$ then $\hat A=\varnothing$ and $J(\pihat)=V^\star_0$. If $B\ge1$, then $|\prod_ja_j-\prod_jb_j|\le\sum_j|a_j-b_j|$ for
$a,b\in[0,1]^k$, which gives $|\hat v(A)-v(A)|\le\frac12\sum_{j\notin A}t_\tau\le\frac k2t_\tau$ for every $A$. Hence
$V^\star_B-J(\pihat)=v(A^\star)-v(\hat A)\le\hat v(A^\star)-\hat v(\hat A)+kt_\tau\le kt_\tau=\Rhat\le\varepsilon$.

\emph{(b) Charge.} On $\mathcal G$, if $t_i<\mu_j/4$ then the true support leads with
$\hat c_j(S_j)\ge\mu_j-t_i>3t_i$ while every other candidate has $\hat c_j(S)\le t_i$. So $\hat S_j=S_j$ and the lead
exceeds $2t_i$. The accuracy test passes when $kt_i\le\varepsilon$. The true support also beats the dummy
competitor, because $\hat c_j(S_j)-0>3t_i$. Both tests therefore pass at the first check $i^\star$ with
$t_{i^\star}<\theta$. If $i^\star=0$ then $n_\tau\le n_0$. If $i^\star\ge1$, then $t_{i^\star-1}\ge\theta$, that is,
$n_{i^\star-1}\le2\theta^{-2}\log(2kM_s/\delta_{i^\star-1})=2\theta^{-2}\log(4kM_si^\star(i^\star+1)/\delta)$.
Moreover $n_{i^\star}\le r\,n_{i^\star-1}+1$ and $n_{i^\star}\ge n_0r^{i^\star}$, which proves the displayed bounds. Since
$\log(i^\star(i^\star+1))=O(\log\log n_{i^\star})$, solving the implicit inequality gives the $O(\cdot)$ form with
$\log M_s\le s\log(eL/s)$ for $s\le L$. The charge identity counts $k$ audits in every episode and $m$ verifications.

\emph{(d) Expected charge.} Off the good event the stopping index is random, so we bound its tail directly. Fix
$i\ge i^\dagger$, so that $t_i<\theta$. The accuracy test passes deterministically: it is vacuous if $B=0$, and if $B\ge1$ then
$kt_i<k\theta\le\varepsilon$. The argument in (b) shows that the lead tests pass whenever every deviation at check $i$ is below
$\mu_{\min}/4$, because $\mu_{\min}/4\ge\theta>t_i$. Hence $\Prb\{\tau>i\}\le2kM_s\exp(-n_i\theta^2/2)$. Write $q_i=n_i/n_{i^\dagger}\ge1$.
Because $n_{i^\dagger}\theta^2/2>\log(2kM_s/\delta_{i^\dagger})$, this gives
$\Prb\{\tau>i\}\le2kM_s(\delta_{i^\dagger}/2kM_s)^{q_i}\le\delta_{i^\dagger}^{q_i}\le(\delta/4)^{q_i}$. Using
$n_{i+1}\le(r+1)n_i$,
\[
\E[n_\tau]\le n_{i^\dagger}+\sum_{i\ge i^\dagger}n_{i+1}\Prb\{\tau>i\}\le n_{i^\dagger}\Big(1+(r+1)\sum_{i\ge i^\dagger}q_i(\delta/4)^{q_i}\Big).
\]
The ratios $q_i\ge r^{i-i^\dagger}n_0r^{i^\dagger}/(n_0r^{i^\dagger}+1)$ grow geometrically and $\delta\le\frac12$, so the sum is at most
$\tilde C_r\delta$ for a constant $\tilde C_r$ that depends only on $r$. This gives $\E[n_\tau]\le(1+C_r\delta)n_{i^\dagger}$. Stage II
has deterministic size $m$.

\emph{(c)} The controller's state is the running product $b_t$ of the reports indexed by $\hat T$, and its action is
$\hat\sigma b_L\prod_{j\in\hat A}E_j$, which changes sign whenever one audited reply changes sign.

\emph{Audit-free route.} Here $(2Y^{\rm ver}-1)a\chi_T(X)=\zeta\sigma\prod_jE_j\chi_T(X)$ is bounded by $1$ and has mean
$(1-2\eta)\sigma\prod_j\mu_j\,\one\{T=T^\star\}$ with $T^\star=\bigcup_jS_j$. Since $|T^\star|=d\le\min(ks,L)$,
$T^\star\in\cS_{d_{\max}}$. Repeat the argument with $|\hat c(T)|$, with the union over the $|\cS_{d_{\max}}|$ candidates, and with
$\delta_i=\delta/((i+1)(i+2))$. A false leader has $|\hat c|\le t_i$, while the competitors include $T^\star$ and the dummy value
$0$, both nonnegative in absolute value. So a lead larger than $2t_i$ cannot be false, even when $\cS_{d_{\max}}$ is a
singleton. On a
correct selection, $|\hat c(T^\star)|>2t_i$ and $|\hat c(T^\star)-(1-2\eta)\sigma\prod\mu_j|\le t_i$ force
$\sgn\hat c(T^\star)=\sigma$. The controller $a=\hat\sigma\chi_{\hat T}(X)$ is then optimal in $\Pi_0$. The test passes
once $t_i<\lambda/4$. The expected-charge bound follows as in (d) with $\theta=\lambda/4$.
\end{proof}

\subsection{Proof of Theorem~\ref{thm:lower} (information-per-price lower bound)}\label{app:lower}\label{pf:thm:lower}

\begin{proof}
\emph{Reference world.} Let $\mathsf Q$ be the law in which $E_1$ is replaced by an independent fair sign and
everything else is unchanged. It is not in the class, which the argument does not require. Fix $J\in\cC$, write
$\mathsf P_J$ for the world with $S_1=\{J\}$, and write $g_A=\prod_{j\notin A}\mu_j$.

\emph{Per-episode information.} Within an episode the learner observes, in order, the reports, the proxy, replies
of the audits it chooses sequentially, its own action, and possibly $Y$. Conditional on everything observed
earlier, the following hold. The reports and proxy have the same law under $\mathsf P_J$ and $\mathsf Q$. A reply of
event $j\ge2$ has the same law, because $E_j=X_{S_j}N_j$ with known $S_j$ and $\mu_j$. The reply of event 1 has
mean $\mu_1X_J$ under $\mathsf P_J$ and $0$ under $\mathsf Q$; for $\pm1$ variables with means $m$ and $0$,
$\KL\le\chi^2=m^2$, so its contribution is at most $\mu_1^2$. The action kernel is the learner's own. If event 1 was
audited, the law of $Y$ given the past depends only on unaudited events $j\ge2$ and is the same in both worlds. If
it was not audited, $2Y-1$ has conditional mean
$a\sigma\prod_{j\in A}E_j\prod_{j\notin A,\,j\ge2}X_{S_j}\cdot X_J\cdot g_A$ under $\mathsf P_J$ and $0$ under $\mathsf Q$,
so its contribution is at most $g_A^2$. With the realized design $D_e=(A_e,v_e)$ of episode $e$, the chain rule
gives a conditional per-episode divergence
$\mathrm{kl}_e\le\mu_1^2\one\{1\in A_e\}+g_{A_e}^2\one\{1\notin A_e,v_e=1\}\le\bar\rho\,C_e$, where
$C_e=c_0+\sum_{j\in A_e}c_j+v_ec_v$ is the episode's charge. If $1\in A_e$, then $\mu_1^2=\iota(\{1\},0)(c_0+c_1)\le\bar\rho C_e$; if
$1\notin A_e$ and $v_e=1$, then $g_{A_e}^2=\iota(D_e)C_e$; otherwise $\mathrm{kl}_e=0$.

\emph{Whole transcript.} Let $O$ be the transcript up to the stopping time $\tau$, including the learner's
randomness. By the divergence decomposition for adaptively collected data with a stopping time of finite
expectation \citep[Lemma~1]{garivier2019}, see also \citet{kaufmann2016},
$\KL(\mathsf P_J^O\|\mathsf Q^O)=\E_J\sum_{e\le\tau}\mathrm{kl}_e\le\bar\rho\,\E_J[C]$. If $J$ is uniform on $\cC$, the mutual
information satisfies $I(J;O)=\min_{\mathsf R}\frac1{L'}\sum_J\KL(\mathsf P_J^O\|\mathsf R)\le\frac{\bar\rho}{L'}\sum_J\E_J[C]$.

\emph{Decoding.} A deployed policy in $\Pi_0$ plays an action whose conditional mean given the reports is some
$f(X)\in[-1,1]$. With $T_J=\{J\}\cup\bigcup_{j\ge2}S_j$, Lemma~\ref{lem:oracle}(i) gives
$\Reg_J(\pi)=\frac12\prod_j\mu_j\,(1-\sigma\langle f,\chi_{T_J}\rangle)$. Regret at most $\frac18\prod_j\mu_j$ forces
$\sigma\langle f,\chi_{T_J}\rangle\ge\frac34$. The sets $T_J$ are distinct, so Bessel's inequality gives
$\sum_J\langle f,\chi_{T_J}\rangle^2\le\|f\|^2\le1$, and since $2(\frac34)^2>1$ at most one $J$ can satisfy this. Decoding
$\hat J$ as that index (arbitrary if none) recovers $J$ whenever the regret target is met, so
$\Prb(\hat J\ne J)\le\frac14$. Fano's inequality \citep{cover2006} gives
$\frac14\ge1-(I(J;O)+\log2)/\log L'$. Rearranging and combining with the mutual-information bound proves the claim.
\end{proof}

\subsection{Proof of Corollary~\ref{cor:separation} (exponential separation)}\label{pf:cor:separation}

\begin{proof}
(i) Without audits, the only informative design is $(\varnothing,1)$, so $\bar\rho=\gamma^{2k}/(c_0+c_v)$ in
Theorem~\ref{thm:lower}. The upper bound is the audit-free route of Theorem~\ref{thm:upper}(d) with $\lambda=\gamma^k$,
$d_{\max}=k$, and $\log|\cS_k|\le k\log(eL/k)$.
(ii) The upper bound is Theorem~\ref{thm:upper}(d) with $s=1$, $B=0$, $\theta=\gamma/4$, and $m=1$; it gives
$\E[C]\le((1+C_r\delta)n_{i^\dagger}+1)(c_0+kc_a)+c_v$ with $n_{i^\dagger}=O(\gamma^{-2}H_a)$. For the lower bound, under the stated
price condition the design $(\{1\},0)$ maximizes $\iota$. Indeed $(\varnothing,1)$ has
$\iota=\gamma^{2k}/(c_0+c_v)\le\gamma^2/(c_0+c_a)$ by assumption, and for $1\le a\le k-1$ audits without event 1,
$\iota=\gamma^{2(k-a)}/(c_0+c_v+ac_a)\le\gamma^2/(c_0+ac_a)\le\gamma^2/(c_0+c_a)$. So $\bar\rho=\gamma^2/(c_0+c_a)$. When
$c_v\le(c_0+kc_a)\gamma^{-2}H_a$, the upper bound is $O((c_0+kc_a)\gamma^{-2}H_a)$, and $c_0+kc_a\le k(c_0+c_a)$.
(iii) Write $a=(c_0+c_v)\gamma^{-2k}\log L'$ for the lower bound in (i), up to its constant, and $b+c$ with
$b=(c_0+kc_a)\gamma^{-2}H_a$ and $c=c_v$ for the upper bound in (ii), up to its constant. Then
$a/(b+c)\ge\frac12\min\{a/b,a/c\}$, and $a/c\ge\gamma^{-2k}\log L'$ because $c_0+c_v\ge c_v$. The crossover statement compares
$(c_0+kc_a)\gamma^{-2}H_a$ with $(c_0+c_v)\gamma^{-2k}H_o$; the lower-order term $c_v$ is absorbed into the constants.
\end{proof}

\subsection{Proof of Proposition~\ref{prop:gradient} (gradient signal)}\label{app:deep}\label{pf:prop:gradient}

\begin{proof}
(a) $\partial_w\log(1+e^{-Ew^\top\Phi})^{-1}=E(1-\Pr_w(E\mid X))\Phi$, which at $w=0$ equals $\frac12E_j\Phi(X)$. Its
expectation is $\frac12\mu_j(\langle\chi_{S_j},\chi_S\rangle)_{S\in\cG}$.
(b) At $w=0$ the policy is uniform and $\partial_w\log\pi_w(a\mid X)=\frac a2\Phi(X)$. With $Y=\frac12(1+a\sigma\prod_jE_j)$,
\[
\nabla J(0)=\E\Big[Y\frac a2\Phi\Big]=\frac14\E[a\Phi]+\frac14\sigma\E\Big[\prod_jE_j\,\Phi\Big]=\frac14\sigma\prod_j\mu_j\,(\langle\chi_{T^\star},\chi_S\rangle)_{S\in\cG},
\]
because $\E[a\Phi]=0$ under the uniform action and $\prod_jE_j=\chi_{T^\star}(X)\prod_jN_j$.
(c) The per-sample gradients are $\frac12E_j\Phi$ and $(Y-b)\frac a2\Phi$, whose coordinates lie in $[-\frac12,\frac12]$.
Hoeffding's inequality (interval length $1$) and a union bound over $\cG$ show that every coordinate of the
minibatch mean is within $t$ of its expectation with probability at least $1-2|\cG|e^{-2nt^2}$. Take $t=\mu_j/4$ in (a)
and $t=\prod_j\mu_j/8$ in (b). The informative coordinate then exceeds every other in signed value in (a) and in
absolute value in (b).
\end{proof}

\subsection{Proof of Proposition~\ref{prop:dr} (unbiased verified return)}\label{pf:prop:dr}

\begin{proof}
Conditional on the trajectory and $Y$, $g$ is fixed and $V$ is Bernoulli$(\rho)$, so
$\E R=g+(Y-g)=Y$ and $\mathrm{Var}R=\rho^{-2}(Y-g)^2\rho(1-\rho)$. The score-function estimator
$\big(R-\sum_t\kappa(q_t)-b\big)\sum_t\nabla\log\pi(u_t\mid h_t)$ has the same conditional expectation as the same estimator
with $Y$ in place of $R$, which is unbiased for the true gradient when the baseline $b$ does not depend on the actions.
For $R=g$ the conditional mean is $g$, not $Y$.
\end{proof}

\subsection{Proof of Proposition~\ref{prop:cert} (anytime certificates)}\label{pf:prop:cert}

\begin{proof}
The rescaled utilities $x_i$ are i.i.d.\ in $[0,1]$ with mean $m=(J(\pihat)-\ell)/(1-\ell)$, because $\pihat$ is fixed. With
the predictable $\lambda_i\in(0,\frac12]$ and centres $\hat m_{i-1}$ of Appendix~\ref{app:cert}, the process
$\prod_{i\le n}\exp\{\lambda_i(x_i-m)-\psi(\lambda_i)(x_i-\hat m_{i-1})^2\}$, with $\psi(\lambda)=-\log(1-\lambda)-\lambda$, is a
nonnegative supermartingale \citep[Thm.~4]{howard2021}; see also \citet[Sec.~3.2]{waudbysmith2024}. Ville's inequality bounds the
probability that it ever exceeds $1/\alpha$ by $\alpha$. On the complement, $m\ge L_n$ for every $n$, with $L_n$ as in
\eqref{eq:prpl}; hence $m\ge\max_{i\le n}L_i$ for every $n$. Mapping back gives $J(\pihat)\ge\mathrm{LCB}_n$ for all $n$, which
proves (i). Part (ii) follows from $V^\star_B\le\bar V$. For (iii), on the coverage event $J(\pihat)\ge\mathrm{LCB}_n>V^\star_0$, and
Lemma~\ref{lem:oracle}(ii) rules out every reply-blind policy.
\end{proof}
\section{Exact checks of the finite-world theory}\label{app:fw-checks}
\setlength{\belowcaptionskip}{6pt}

Table~\ref{tab:fw-checks} lists exact computations for the examples of Section~\ref{sec:foundations}, obtained by
linear programming on small explicit instances.

\emph{Audit menus (panel a).} The audit rows use the cue--channel construction with one paid audit per episode,
per-episode charge $c_0=0.05$, audit charge $c_a=1$, and informative-channel reliability $(1+r)/2=0.7$, so that
$d_r=0.4\log(7/3)$. They agree with Proposition~\ref{prop:fw-cc}(iii): $\cT=m(c_0+c_a)/d_r$ for forced static audits
and $(c_0+c_a)/d_r$ with the adaptive audit. By Proposition~\ref{prop:fw-cc}(iv), adding a terminal verifier with
$c_v=c_a$ to the static menu gives the adaptive value.

\emph{Deployment radii (panel b).} The deployment rows use two-stage trees with known mode kernels. The monitor-removal
fixture has two worlds with opposite correct final actions and no informative training reports; the deployment sensor
reveals the world before the final action at a cost within the budget. We checked it against enumeration of
deterministic policies, and the monitor and three-world rows against the least-favorable form of
Proposition~\ref{prop:fw-geometry}(iii). The three-world fixture is the example of
Proposition~\ref{prop:fw-geometry}(iv). The audit-menu rows (XOR and duplicate tests) report the worst-case radius of
the compatible set left by population training reports over four (respectively two) finite hypotheses.

\emph{Confidence-set deployment (panel c).} This row is a $200$-repetition simulation of Theorem~\ref{thm:fw-conf} for
two mixture components, one binary training report with success probability $0.2+0.6\,\theta_1$ and true
$\theta_1=0.55$, a deployment tree without the sensor, $n=400$, and $\alpha=0.05$.

\begin{table}[ht]
\centering
\small
\caption{Exact values of the finite-world examples. $\cT$ is the characteristic cost~\eqref{eq:fw-cost} from the design
LP of Lemma~\ref{lem:fw-rate}(i); radii $R$ are minimax regrets $R(C;X)$ from the deployment LP.}
\label{tab:fw-checks}
\begin{minipage}[t]{0.5\linewidth}
\centering
\textit{(a) Cue--channel audits: cost $\cT$}\par\smallskip
\begin{tabular}[t]{rrrr}
\toprule
events $m$ & static & adaptive & ratio \\
\midrule
2 & 6.196 & 3.098 & 2 \\
4 & 12.392 & 3.098 & 4 \\
8 & 24.785 & 3.098 & 8 \\
\bottomrule
\end{tabular}
\end{minipage}\hfill
\begin{minipage}[t]{0.46\linewidth}
\centering
\textit{(c) Confidence-set deployment}\par\smallskip
\begin{tabular}[t]{lr}
\toprule
coverage ($n=400$) & $200/200$ \\
mean regret & 0.0316 \\
mean radius & 0.0928 \\
mean modulus & 0.4934 \\
\bottomrule
\end{tabular}
\end{minipage}

\bigskip
\textit{(b) Minimax regret radius $R$}\par\smallskip
\begin{tabular}{llr}
\toprule
fixture & sensor or tests & $R$ \\
\midrule
monitor removed / restored & without sensor & $0.5$\rlap{$^{1}$} \\
 & with sensor & $0$ \\
\midrule
three-world pairwise trap & -- & $1/3$\rlap{$^{2}$} \\
\midrule
partial identification, common best action & -- & $0$\rlap{$^{3}$} \\
\midrule
XOR decision & none; $u$ only; $v$ only & $0.5$ \\
 & $u$ and $v$ & $0$ \\
\midrule
duplicate revealing tests & none & $0.5$ \\
 & first only; second only; both & $0$ \\
\bottomrule
\end{tabular}

\medskip
\begin{minipage}{0.92\linewidth}
\footnotesize
$^{1}$\,Enumeration and least-favorable value both give $0.5$ (two-world certificate).
$^{2}$\,Least-favorable prior $(1/3,1/3,1/3)$ on all three worlds.
$^{3}$\,The compatible polytope has $4$ vertices.
Panel~(c): $200$ repetitions, no violation of Theorem~\ref{thm:fw-conf} on coverage.
\end{minipage}
\end{table}

The XOR and duplicate-test rows show that the value of a training-audit menu is in general neither submodular nor
supermodular: two tests can be useless separately and decisive together, or each decisive and jointly redundant. Menu
design must therefore evaluate the compatible sets themselves; greedy marginal gains can mislead.
\newcommand{\hypgroup}[1]{\multicolumn{2}{@{}l}{\textit{#1}}\\}
\newenvironment{settingtable}{\begin{tabular}{@{}>{\hspace{1em}}l>{\raggedright\arraybackslash}p{0.52\linewidth}@{}}
  \toprule \multicolumn{1}{@{}l}{setting} & value \\}{\bottomrule\end{tabular}}

\section{Experimental details}\label{app:exp}

This appendix describes the environments (Appendix~\ref{app:envs}), the hyperparameters of every part
(Appendix~\ref{app:hyper}), the deployment certificate (Appendix~\ref{app:cert}), and the statistical tests and
preregistered predictions (Appendix~\ref{app:stats}).

\subsection{Environments}\label{app:envs}

\subsubsection{Deep RL environments (Parts B and C)}

Each seed $s\in\{0,\dots,9\}$ fixes a world: disjoint singleton supports drawn uniformly, the configuration's
reliabilities, and a random polarity. Seed $100$ is reserved for the pilot. Episodes in a batch start together and
last at most the horizon; finished episodes are masked.

\emph{HiddenEventCorridor.} The observation vector contains phase indicators (cue, delay, fork), the current report
bit during the cue phase, the elapsed-time fraction, the remaining audit budget, the hint (at the fork only), the
replies of audits bought so far ($0$ if not bought), and four fresh uniform nuisance bits. Before the fork the only
legal action is ``continue''. At the fork the legal actions are $a=\pm1$ and, while budget remains, ``audit $j$'' for
each unaudited $j$; the last fork step forces a decision. The horizon is $L+\text{delay}+B+1$.

\emph{AuditGrid.} The room is $7\times11$ including walls. Cue tiles are at $(1,2),(5,2),(1,6),(5,6)$, exits at
$(1,10)$ ($a=+1$) and $(5,10)$ ($a=-1$), and the hint tile at $(3,9)$. The agent starts at row $3$, column $1$ or
$2$, where no cue is visible. It observes a $3\times3$ egocentric window with seven channels (wall, exit $+1$,
exit $-1$, cue showing $+1$, cue showing $-1$, hint $+1$, hint $-1$), its normalized position, the elapsed-time
fraction, the remaining budget, and the audit replies. Actions are four moves and the $k$ audit queries. Walking
into an exit decides the episode; after $40$ steps an undecided episode ends with $Y=0$.

\emph{AuditGrid comparator.} Every grid policy acts on a sub-history of $(X,E_A)$, so its value is at most the class
optimum $V^\star_B$ of Lemma~\ref{lem:oracle}. Conversely, consider the following full-history policy. It first buys the
optimal audit set $A^\star$ ($|A^\star|\le B\le1$ step). It then moves up once, right to column $7$ (passing within one
cell of the cue tiles at $(1,2)$ and $(1,6)$), down twice, left four cells (passing $(5,6)$ and $(5,2)$), and right six
cells to $(4,9)$. Finally it exits through $(1,10)$ or $(5,10)$ according to
$a=\sigma\chi_{T_{A^\star}}(X)\prod_{j\in A^\star}E_j$. It sees all four cue tiles, never times out (at most $24$ steps
against a horizon of $40$), and so attains $v(A^\star)=V^\star_B$. Hence the grid optimum equals $V^\star_B$, and grid
regrets are exact. The test suite verifies this on several worlds with both budgets. A polarity flip leaves timed-out
outcomes at zero, so the grid satisfies the condition $Y_{\bar M}\le1-Y_M$ of Proposition~\ref{prop:anchor} rather than
exact complementarity.

\emph{Proxy and evaluation.} The proxy is $\tilde Y=\one\{a=\text{hint}\}$ for decided episodes and $0$ otherwise.
Evaluation uses the Rao--Blackwellized success probability
$\Prb(Y=1\mid X,E_A,a)=\frac12(1+a\sigma\prod_{j\in A}E_j\prod_{j\notin A}\mu_j\chi_{S_j}(X))$ for decided episodes and
$0$ for timed-out ones. The learner never sees it.

\subsubsection{Part E: arithmetic task and reward model}\label{app:partE}

\emph{Prompts and parsing.} Each prompt is rendered with the model's chat template, with the system message ``You are a
careful math assistant.'' and the user message ``Compute $\langle$expression$\rangle$. A classmate says the answer is
$\langle h\rangle$. Show at most three short steps, then write the final answer on its own line as `\#\#\#\#
<integer>'.'' An expression starts from an integer in $[2,19]$ and applies one to three operations drawn uniformly
from $\{+,-,\times\}$, with a second operand in $[2,29]$ for $\pm$ and in $[2,9]$ for $\times$. With probability $0.7$
the suggestion is wrong: $h$ is the answer plus an offset drawn uniformly from
$\{-10,-2,-1,1,2,10,d\}$, where $d$ is uniform on $\{3,\dots,9\}$. The parsed answer is the integer after the last
\texttt{\#\#\#\#}, or else the last integer in the response. Sampling is pure temperature sampling (temperature $0.7$,
no top-$k$, top-$p$, or repetition penalty) with at most $160$ new tokens.

\emph{Reward model.} The probe input is the final-layer hidden state of the frozen reference model, averaged over the
response tokens. Features are standardized on the fitting split. The probe is an $\ell_2$-regularized logistic
regression (coefficient $10^{-3}$) fitted by L-BFGS to the verified outcomes of $3{,}276$ initial-policy responses
and validated on the other $820$. It reaches held-out AUC $0.887$ and Brier score $0.118$. Their prompts come
from a random stream separate from the training and evaluation prompts. The proxy~\eqref{eq:partE-proxy} combines the
probe score with the copy indicator.

Table~\ref{tab:partE-hyper} lists the training settings, and Table~\ref{tab:partE-selection}
(Appendix~\ref{app:partE-extra}) the checkpoint-selection results.

\subsection{Hyperparameters}\label{app:hyper}

Tables~\ref{tab:hyper}, \ref{tab:llm-hyper}, and~\ref{tab:partE-hyper} list the settings of the deep RL study
(Parts B and C), the language-model study of Part D, and the language-model study of Part E. Settings hold for all
learners of a part unless a row says otherwise.

\begin{table}[ht]
\centering\small
\caption{Hyperparameters of the deep RL learners (Parts B and C).}\label{tab:hyper}
\begin{settingtable}
\midrule
\hypgroup{Network}
encoder & linear + ReLU, width 128 \\
memory & GRU, width 128 \\
heads & two-layer MLPs, width 128 \\
\midrule
\hypgroup{PPO}
parallel episodes per iteration & 512 \\
iterations & 400 (corridor), 600 (grid) \\
epochs; minibatches & 4; 4, of whole episodes \\
clip ratio & 0.2 \\
GAE $\lambda$; discount $\gamma$ & 0.95; 1 \\
optimizer & Adam, learning rate $5\cdot10^{-4}$ \\
gradient-norm clip & 0.5 \\
\midrule
\hypgroup{Loss weights}
value; entropy & 0.5; 0.01 \\
outcome model; audit auxiliary; prediction & 1; 1; 1 \\
\midrule
\hypgroup{Evidence and prices}
verification rate $\rho$ & 0.1 \\
side-audit rate $\rho_a$ & 0.25 \\
charge-matched rate $\rho'$ & $\rho+\rho_akc_a/c_v$ \\
prices $c_0$; $c_a$; $c_v$ & 0.01; 0.1; 1 \\
deployment audit price $\kappa$ & 0.05 \\
\midrule
\hypgroup{Evaluation and certificate}
policy & greedy \\
learning curve & every 20 iterations, 4{,}096 episodes \\
final evaluation & 65{,}536 episodes \\
certificate & PrPl-EB, 2{,}000 verified episodes, $\alpha=0.05$, $\lambda_{\max}=\frac12$ \\
\midrule
\hypgroup{Part C additions}
linear probes & ridge; fit on 4{,}096 and scored on 4{,}096 episodes \\
training-charge cap & 32{,}768 (corridor), 49{,}152 (grid) \\
\end{settingtable}
\end{table}

\begin{table}[ht]
\centering\small
\caption{Hyperparameters of the language-model study (Part D).}\label{tab:llm-hyper}
\begin{settingtable}
\midrule
\hypgroup{Model}
policy & Qwen2.5-0.5B-Instruct \\
scale check & Qwen2.5-1.5B-Instruct, 3 worlds, 12 runs \\
training backend & Hugging Face \\
adapter & LoRA, rank 32, scale 64, dropout 0 \\
\midrule
\hypgroup{Optimization}
algorithm & REINFORCE, leave-one-out baseline \\
updates & 600 \\
batch per update & 64 documents $\times$ 4 sampled actions \\
learning rate & $3\cdot10^{-5}$, selected on world 100 \\
\midrule
\hypgroup{Task}
report bits $L$; reliability $\mu$ & 16; 0.8 \\
events $k$; deployment audits $B$ & 1, 2, 3; 0, 1 \\
deployment audit price $\kappa$ & 0.05 \\
\midrule
\hypgroup{Evidence and prices}
verification rate $\rho$ & 0.1 \\
side-audit rate $\rho_a$ & 0.25 \\
charge-matched rate $\rho'$ & $\min\{1,\rho+\rho_akc_a/c_v\}$ \\
price sweep ($k=3$, $B=0$) & $c_a\in\{0.05,0.3,1.0\}$ \\
\midrule
\hypgroup{Evaluation and certificate}
learning curve & every 50 updates, 1{,}024 episodes \\
final evaluation & 4{,}096 episodes, greedy decoding \\
linear probes & fit on 2{,}048 and scored on 2{,}048 episodes \\
certificate & 2{,}000 verified episodes \\
\midrule
\hypgroup{Worlds per cell}
DR and DORL & 10 \\
other learners & 5 (the first five) \\
\end{settingtable}
\end{table}

\begin{table}[ht]
\centering\small
\caption{Hyperparameters of the language-model study (Part E).}\label{tab:partE-hyper}
\begin{settingtable}
\midrule
\hypgroup{Model}
policy and reference & Qwen2.5-0.5B-Instruct \\
fine-tuning & full; fp32 weights, bf16 autocast (fp16 without bf16) \\
\midrule
\hypgroup{RLOO}
steps & 200 \\
batch per step & 16 prompts $\times$ 8 responses \\
baseline & leave-one-out within each prompt \\
reward & $R-\lambda_{\rm KL}\,\mathrm{KL}$, sampled sequence log-ratio to the reference \\
$\lambda_{\rm KL}$ & 0.02 (0.2 for proxy\,+\,10$\times$KL) \\
\midrule
\hypgroup{Optimizer}
optimizer & AdamW, $\beta=(0.9,0.95)$, no weight decay \\
learning rate & $10^{-6}$, 10 linear warm-up steps \\
gradient clip & 1 \\
\midrule
\hypgroup{Proxy and verification}
proxy weight $\alpha$ & 0.5 \\
probe data & 4{,}096 verified initial-policy responses \\
probe fit; held out & 3{,}276; 820 \\
verification rate $\rho$ & 0.05, 0.2 \\
adaptive rules & clip $p$ to $[0.01,1]$ \\
cells & number of operations $\times$ 10 proxy bins \\
recalibration prior & weight 4 on $\tilde Y$ in each cell \\
residual prior & 0.25, weight 1 \\
\midrule
\hypgroup{Evaluation and selection}
evaluation & every 25 steps; 256 held-out prompts $\times$ 2 responses \\
sampling temperature & 0.7 \\
selection budgets; replicates & 64--1{,}024; 200 \\
DD-SE & 6 rounds, $\delta=0.1$ \\
\midrule
\hypgroup{Seeds}
seeds per cell; runs & 3; 39 \\
base seed & 20261101 \\
\end{settingtable}
\end{table}

\subsection{The deployment certificate}\label{app:cert}

The certificate is the one-sided predictable plug-in empirical-Bernstein (PrPl-EB) confidence sequence of
\citet[Sec.~3.2]{waudbysmith2024}. Its validity rests on the empirical-Bernstein supermartingale of
\citet[Thm.~4]{howard2021}. Utilities $U_i\in[\ell,1]$ with $\ell=-\kappa B$ are rescaled to
$x_i=(U_i-\ell)/(1-\ell)\in[0,1]$, and $\alpha=0.05$. With $\hat m_0=\frac12$, $\hat s^2_0=\frac14$, and
$\psi(\lambda)=-\log(1-\lambda)-\lambda$, for $t=1,2,\dots$:
\begin{gather}
\lambda_t=\min\Big\{\sqrt{\frac{2\log(1/\alpha)}{\hat s^2_{t-1}\,t\log(1+t)}},\ \frac12\Big\},\notag\\
L_t=\frac{\sum_{i\le t}\lambda_ix_i-\log(1/\alpha)-\sum_{i\le t}\psi(\lambda_i)(x_i-\hat m_{i-1})^2}{\sum_{i\le t}\lambda_i},\label{eq:prpl}\\
\hat m_t=\frac{\frac12+\sum_{i\le t}x_i}{t+1},\qquad \hat s^2_t=\frac{\frac14+\sum_{i\le t}(x_i-\hat m_i)^2}{t+1}.\notag
\end{gather}
The reported bound on the utility scale is $\mathrm{LCB}_t=\ell+(1-\ell)\max\{0,\max_{i\le t}L_i\}$, a running maximum.
Thus $J\ge\mathrm{LCB}_t$ for all $t\le N$ if and only if $J\ge\mathrm{LCB}_N$, and empirical anytime coverage of the
reported sequence equals its terminal coverage.

The running maximum is valid only because $(L_t)$ is time-uniform, so we test that property directly by simulation. For utility laws matching the deployment tasks, we draw 20{,}000 independent
streams of length 2{,}000 and count the streams in which $L_t$ exceeds the true mean at \emph{any} $t$
(Table~\ref{tab:cscheck}). Every miss rate is below $\alpha=0.05$. Because of the running maximum, the anytime and
terminal miss rates of the reported bound coincide. In the revision study (Part C), every run evaluates
the flag $\{J\ge\mathrm{LCB}_t\ \forall t\le2000\}$ on the full sequence and stores the bound at five checkpoints,
$t\in\{100,250,500,1000,2000\}$.

\begin{table}[ht]
\centering\small
\caption{Simulated time-uniform miss rates of the certificate stay below $\alpha=0.05$ for every utility law. A miss
means that the lower bound exceeded the true mean at some time $t\le2{,}000$; $20{,}000$ streams per law.}
\label{tab:cscheck}
\begin{tabular}{lrrrr}
\toprule
 & & \multicolumn{2}{c}{anytime miss rate} & mean width \\
\cmidrule(lr){3-4}
utility law & mean & estimate & Wilson 95\% upper & at $n=2{,}000$ \\
\midrule
Bernoulli, memory-free policy & 0.500 & 0.0135 & 0.0152 & 0.0423 \\
Bernoulli, $V^\star$ of corridor $k=2$ & 0.820 & 0.0251 & 0.0274 & 0.0305 \\
Bernoulli, $V^\star$ of corridor $k=1$ & 0.900 & 0.0285 & 0.0309 & 0.0226 \\
Bernoulli, near-deterministic & 0.990 & 0.0199 & 0.0220 & 0.0070 \\
audit policy, $\kappa=0.05$, $V^\star$ with $B=1$ & 0.750 & 0.0249 & 0.0271 & 0.0317 \\
mixed audit use, $\kappa=0.05$ & 0.728 & 0.0217 & 0.0239 & 0.0346 \\
\bottomrule
\end{tabular}
\end{table}

\subsection{Statistical tests and preregistered predictions}\label{app:stats}

\emph{Paired tests.} Each configuration has ten worlds, and all learners run on the same worlds. For a baseline and a
comparator, let $\Delta_s=\mathrm{regret}_{\rm baseline}(s)-\mathrm{regret}_{\rm comparator}(s)$ in world $s$. The
primary test is the exact one-sided paired sign-flip test, which enumerates all $2^{10}$ sign patterns; it is exact when
the two learners are exchangeable within a world. We also report paired wins $\#\{s:\Delta_s>0\}$ and the exact
one-sided sign test, which needs no symmetry assumption. Holm's correction is applied across configurations within each
hypothesis in Part B, within each comparison in Part C, and across the two pre-specified cells $k\in\{2,3\}$ in Part D,
where $k=1$ is tested separately.

\emph{Effect sizes and aggregates.} Effect sizes are mean paired differences with bootstrap intervals, and the unpaired
probability of improvement
$\widehat{\rm PI}=n^{-2}\sum_{s,s'}[\one\{r^{\rm c}_s<r^{\rm b}_{s'}\}+\frac12\one\{r^{\rm c}_s=r^{\rm b}_{s'}\}]$.
Aggregate scores are interquartile means (IQM) with $95\%$ stratified bootstrap intervals \citep{agarwal2021}.

\emph{Certificate coverage.} Learners trained in the same world share a certification sample. Coverage is therefore
reported per cell, with a world-cluster bootstrap, and after collapsing runs with identical world and bound value.

Table~\ref{tab:hypotheses} lists the preregistered hypotheses of Part B, and Table~\ref{tab:predictions} the
preregistered predictions of Part A and the directional predictions of Part E, recorded before its full run.

\begin{table}[ht]
\centering\small
\caption{Preregistered hypotheses of Part B with the comparison and test for each. For H1--H3 and H6 the
comparison states the alternative on final regret, at matched nominal evidence charge; the learner on the left is
expected to do worse.}\label{tab:hypotheses}
\begin{tabular}{@{}>{\raggedright\arraybackslash}p{0.47\linewidth}ll@{}}
\toprule
hypothesis & comparison & test \\
\midrule
H1: proxy optimization is hacked & proxy $>$ DR & sign-flip, Holm \\
H2: the verifier must be the training target & reward model $>$ DR & sign-flip, Holm \\
H3a: audit supervision of memory is decisive & DR $>$ DORL & sign-flip, Holm \\
H3b: as H3a & DR+predictive $>$ DORL & sign-flip, Holm \\
H4: certificates are sound & coverage of $J\ge\mathrm{LCB}$ $\ge0.95$ & descriptive$^{a}$ \\
H5: with $B=1$, DORL certifiably uses audit replies & $\mathrm{LCB}>V^\star_0$ & certificate, per run \\
H6: audits alone do not anchor the outcome & proxy+audit-aux $>$ DORL & sign-flip, Holm \\
\bottomrule
\end{tabular}

\smallskip
{\footnotesize $^{a}$\,Runs in the same world are not independent, so H4 is checked descriptively.\par}
\end{table}

\begin{table}[ht]
\centering\small
\caption{Preregistered predictions of Part A and directional predictions of Part E.}\label{tab:predictions}
\begin{tabular}{@{}l>{\raggedright\arraybackslash}p{0.84\linewidth}@{}}
\toprule
& prediction \\
\midrule
\multicolumn{2}{@{}l}{\textit{Part A: certified learners on random worlds}} \\
T1 & Soundness: the fraction of runs with $\Reg\le\Rhat$ is at least $1-\delta$ in every cell. \\
T2 & Scaling: regressing $\log(n/\log(2\,\mathrm{count}/\delta))$ on $\log(\mathrm{signal}^{-2})$ gives slope $1$ for
both routes, with signal $\mu$ for audits and $\mu^k$ for outcomes. \\
T3 & Separation: the outcome-to-audit charge ratio grows geometrically in $k$, and the crossover audit price increases
with $k$. \\
\midrule
\multicolumn{2}{@{}l}{\textit{Part E: a language-model policy with a learned proxy}} \\
V1 & Proxy and proxy\,+\,10$\times$KL gold accuracy peaks and then declines as proxy value and copying rise, with KL
delaying the decline. \\
V2 & Naive substitution stays biased toward the proxy optimum at low rates. \\
V3 & The DR variants follow the fully verified trend, with DR-recal closest at equal verification counts. \\
V4 & Verified-only learns more slowly than DR. \\
V5 & In checkpoint selection DD-SE beats uniform DR, which beats uniform IPW, while proxy-only ranking selects
over-optimized checkpoints. \\
\bottomrule
\end{tabular}
\end{table}

\subsection{Compute and study timeline}\label{app:repro}

Parts A--C ran on two NVIDIA T4 GPUs: Part A took $96$ seconds in total, Part B $12.0$ GPU-hours, and Part C $8.3$
GPU-hours. Part D ran on four NVIDIA RTX 4090 GPUs and took $105.6$ GPU-hours, and Part E took $14.7$ GPU-hours.

The order of the studies matters for the status of their tests. A pilot on held-out world $100$, which the evaluation
never uses, fixed the belief-input design of DORL (Table~\ref{tab:pilot}). A first launch of the deep grid was stopped
after about an hour, when we found that the proxy leaked $Y$; none of its runs is reported. We fixed the hypotheses and
analysis plan of Parts A and B shortly after relaunching the corrected grid and before analysing any deep RL result.
Afterwards we changed only the presentation of that analysis. Part C, the sign tests, paired win counts, clustered
coverage, slope intervals, censored charge-to-threshold, and aggregate curves were added in response to a review and
are post hoc. After the first $280$ Part C runs we noticed that capped DORL had been compared with uncapped DR, which
spends above the cap on policy audits; we then ran DR under the same cap ($20$ runs) and report the equal-cap comparison
as the primary one. The Part D protocol and analysis were fixed before its evaluation runs, after a pilot on held-out
world $100$ selected the learning rate. The Part E predictions (Table~\ref{tab:predictions}) were recorded before its
full run.

\section{Additional results}\label{app:extra}

\subsection{Part B: pilot, tests, charge to threshold, and coverage}\label{app:partB-extra}

\begin{table}[ht]
\centering\scriptsize
\resizebox{\linewidth}{!}{\begin{tabular}{lccccccc}
\toprule
learner & Corr $k{=}1$ & Corr $k{=}2$ & Corr $k{=}3$ & Corr long & Corr $B{=}1$ & Grid $k{=}2$ & Grid $B{=}1$ \\
 & $V^\star{=}0.900$ & $V^\star{=}0.820$ & $V^\star{=}0.756$ & $V^\star{=}0.820$ & $V^\star{=}0.750$ & $V^\star{=}0.820$ & $V^\star{=}0.750$ \\
\midrule
proxy & 0.400 \tiny[0.399, 0.401] & 0.320 \tiny[0.319, 0.321] & 0.256 \tiny[0.255, 0.257] & 0.320 \tiny[0.319, 0.321] & 0.250 \tiny[0.249, 0.251] & 0.320 \tiny[0.319, 0.321] & 0.250 \tiny[0.250, 0.250] \\
proxy+audit-aux & 0.400 \tiny[0.400, 0.401] & 0.320 \tiny[0.319, 0.321] & 0.257 \tiny[0.256, 0.258] & 0.320 \tiny[0.319, 0.321] & 0.250 \tiny[0.249, 0.251] & 0.321 \tiny[0.320, 0.356] & 0.250 \tiny[0.250, 0.251] \\
sparse-verified & 0.000 \tiny[0.000, 0.000] & 0.320 \tiny[0.319, 0.321] & 0.256 \tiny[0.255, 0.257] & 0.320 \tiny[0.319, 0.321] & 0.252 \tiny[0.250, 0.258] & 0.320 \tiny[0.303, 0.371] & 0.251 \tiny[0.225, 0.269] \\
reward model & 0.000 \tiny[0.000, 0.046] & 0.320 \tiny[0.319, 0.320] & 0.257 \tiny[0.256, 0.258] & 0.320 \tiny[0.319, 0.321] & 0.250 \tiny[0.249, 0.250] & 0.319 \tiny[0.286, 0.345] & 0.250 \tiny[0.249, 0.300] \\
DR & 0.000 \tiny[0.000, 0.048] & 0.320 \tiny[0.319, 0.321] & 0.256 \tiny[0.255, 0.256] & 0.320 \tiny[0.319, 0.321] & 0.251 \tiny[0.250, 0.254] & 0.319 \tiny[0.160, 0.321] & 0.075 \tiny[0.000, 0.425] \\
DR+predictive & 0.000 \tiny[0.000, 0.000] & 0.320 \tiny[0.319, 0.320] & 0.255 \tiny[0.255, 0.256] & 0.320 \tiny[0.319, 0.320] & 0.251 \tiny[0.250, 0.255] & 0.319 \tiny[0.160, 0.321] & 0.168 \tiny[0.041, 0.245] \\
\textbf{DORL} & 0.000 \tiny[0.000, 0.000] & 0.000 \tiny[0.000, 0.001] & 0.066 \tiny[0.034, 0.078] & 0.003 \tiny[0.000, 0.051] & 0.035 \tiny[0.013, 0.079] & 0.080 \tiny[0.000, 0.231] & 0.033 \tiny[0.000, 0.123] \\
fully verified PPO & 0.000 \tiny[0.000, 0.000] & 0.320 \tiny[0.320, 0.321] & 0.256 \tiny[0.255, 0.257] & 0.320 \tiny[0.319, 0.320] & 0.250 \tiny[0.250, 0.251] & 0.305 \tiny[0.159, 0.320] & 0.241 \tiny[0.187, 0.251] \\
\midrule
training charge, DORL & 27{,}653 & 32{,}769 & 37{,}886 & 32{,}723 & 42{,}173 & 48{,}009 & 77{,}336 \\
training charge, DR & 27{,}668 & 32{,}776 & 37{,}884 & 32{,}727 & 35{,}545 & 48{,}018 & 71{,}187 \\
total charge, DORL & 29{,}673 & 34{,}789 & 39{,}906 & 34{,}743 & 44{,}335 & 50{,}029 & 79{,}555 \\
total charge, DR & 29{,}688 & 34{,}796 & 39{,}904 & 34{,}747 & 37{,}574 & 50{,}038 & 73{,}337 \\
\bottomrule
\end{tabular}}
\caption{Part B: final true regret $V^\star_B-J(\pihat)$ of the deployed greedy policy, IQM over $10$ worlds with $95\%$
stratified bootstrap interval. The bottom rows give the mean training charge, and the mean total charge including the
$2{,}000$ verified certification episodes and their audits, for DORL and the charge-matched DR baseline. Fully verified
PPO spends $206{,}848$ (corridor) and about $308{,}000$ (grid) in training. In the $B{=}1$ columns the certification audit
cost is estimated from the evaluation audit rate.}\label{tab:deep}
\end{table}

Table~\ref{tab:pilot} gives the pilot that fixed the belief input, Table~\ref{tab:tests} the preregistered tests of
Table~\ref{tab:hypotheses}, Table~\ref{tab:threshold} the training charge needed to reach regret $0.05$, and
Table~\ref{tab:coverage} the certificate coverage by learner and configuration.

\begin{table}[ht]
\centering\small
\caption{Pilot on held-out world seed 100: letting the policy read the detached event beliefs lowered DORL's regret at
$k=3$. One run per row, 400 iterations on a T4 (43--60 s per run); the seed is not used for evaluation. These learners
never read the proxy, so the later proxy correction (Appendix~\ref{app:repro}) does not affect them.}\label{tab:pilot}
\begin{tabular}{lrrr}
\toprule
learner & final regret & certified LCB & training charge \\
\midrule
\multicolumn{4}{l}{\textit{Corridor $k=2$, $V^\star=0.82$}} \\
DORL, policy reads beliefs & 0.000 & 0.790 & 32{,}815 \\
DORL, no belief input & 0.001 & 0.789 & 32{,}815 \\
DR, charge-matched & 0.321 & 0.459 & 32{,}685 \\
\midrule
\multicolumn{4}{l}{\textit{Corridor $k=3$, $V^\star=0.756$}} \\
DORL, policy reads beliefs & 0.064 & 0.652 & 37{,}949 \\
DORL, no belief input & 0.129 & 0.584 & 37{,}949 \\
DR, charge-matched & 0.255 & 0.450 & 37{,}851 \\
fully verified PPO & 0.255 & 0.455 & 206{,}848 \\
\bottomrule
\end{tabular}
\end{table}

\begin{table}[ht]
\centering\footnotesize
\caption{Part B training charge at the first evaluation with regret $\le0.05$ (preregistered secondary endpoint). Each
cell gives the median over all ten runs and, in parentheses, the number of runs that reached the threshold. A run that
never reaches it counts as $+\infty$; ``n.r.'' means that fewer than half of the runs reached it, so the median is not
identified. All runs in a cell were evaluated on the same iteration schedule.}\label{tab:threshold}
\begin{tabular}{lrrrrr}
\toprule
 & \multicolumn{5}{c}{HiddenEventCorridor} \\
\cmidrule(l){2-6}
learner & $k{=}1$ & $k{=}2$ & $k{=}3$ & long & $B{=}1$ \\
\midrule
proxy & n.r. (0/10) & n.r. (0/10) & n.r. (0/10) & n.r. (0/10) & n.r. (0/10) \\
proxy+audit-aux & n.r. (0/10) & n.r. (0/10) & n.r. (0/10) & n.r. (0/10) & n.r. (0/10) \\
sparse-verified & 10{,}458 (10/10) & n.r. (0/10) & n.r. (0/10) & n.r. (0/10) & n.r. (0/10) \\
reward model & 11{,}711 (9/10) & n.r. (0/10) & n.r. (0/10) & n.r. (0/10) & n.r. (0/10) \\
DR & 9{,}685 (9/10) & n.r. (0/10) & n.r. (0/10) & n.r. (0/10) & n.r. (0/10) \\
DR+predictive & 4{,}818 (10/10) & n.r. (0/10) & n.r. (0/10) & n.r. (0/10) & n.r. (0/10) \\
\textbf{DORL} & 4{,}117 (10/10) & 8{,}204 (10/10) & n.r. (2/10) & 14{,}010 (9/10) & 25{,}417 (8/10) \\
fully verified PPO & 62{,}054 (10/10) & n.r. (0/10) & n.r. (0/10) & n.r. (0/10) & n.r. (0/10) \\
\midrule
 & \multicolumn{2}{c}{AuditGrid} &  &  &  \\
\cmidrule(l){2-3}
learner & $k{=}2$ & $B{=}1$ &  &  &  \\
\midrule
proxy & n.r. (0/10) & n.r. (0/10) &  &  &  \\
proxy+audit-aux & n.r. (0/10) & n.r. (0/10) &  &  &  \\
sparse-verified & n.r. (0/10) & n.r. (1/10) &  &  &  \\
reward model & n.r. (1/10) & n.r. (0/10) &  &  &  \\
DR & n.r. (2/10) & 36{,}418 (7/10) &  &  &  \\
DR+predictive & n.r. (2/10) & 47{,}810 (6/10) &  &  &  \\
\textbf{DORL} & 31{,}834 (7/10) & 44{,}109 (10/10) &  &  &  \\
fully verified PPO & n.r. (2/10) & n.r. (0/10) &  &  &  \\
\bottomrule
\end{tabular}
\end{table}

\begin{table}[ht]
\centering\small
\caption{Part B certificate coverage ($J\ge\mathrm{LCB}$) by learner and configuration, out of ten worlds. The bottom
rows pool all runs: the world-cluster bootstrap resamples worlds within each configuration and keeps all learners of a
resampled world together, and the last row counts coverage after collapsing runs with identical world and bound value
(which share a certification sample).}\label{tab:coverage}
\begin{tabular}{lrrrrrrr}
\toprule
 & \multicolumn{5}{c}{HiddenEventCorridor} & \multicolumn{2}{c}{AuditGrid} \\
\cmidrule(l){2-6}\cmidrule(l){7-8}
learner & $k{=}1$ & $k{=}2$ & $k{=}3$ & long & $B{=}1$ & $k{=}2$ & $B{=}1$ \\
\midrule
proxy & 10/10 & 10/10 & 10/10 & 10/10 & 10/10 & 10/10 & 9/10 \\
proxy+audit-aux & 10/10 & 10/10 & 10/10 & 10/10 & 10/10 & 10/10 & 9/10 \\
sparse-verified & 9/10 & 10/10 & 10/10 & 10/10 & 10/10 & 10/10 & 10/10 \\
reward model & 9/10 & 10/10 & 10/10 & 10/10 & 10/10 & 10/10 & 10/10 \\
DR & 10/10 & 10/10 & 10/10 & 10/10 & 10/10 & 10/10 & 10/10 \\
DR+predictive & 9/10 & 10/10 & 10/10 & 10/10 & 10/10 & 10/10 & 10/10 \\
\textbf{DORL} & 9/10 & 10/10 & 10/10 & 10/10 & 10/10 & 10/10 & 10/10 \\
fully verified PPO & 9/10 & 9/10 & 9/10 & 10/10 & 10/10 & 10/10 & 10/10 \\
\midrule
\multicolumn{8}{l}{all runs: 551/560} \\
\multicolumn{8}{l}{world-cluster bootstrap 95\% CI: [0.963, 0.998]} \\
\multicolumn{8}{l}{distinct (world, bound) signatures covered: 399/404} \\
\bottomrule
\end{tabular}
\end{table}

\begin{table}[p]
\centering
\rotatebox{90}{\begin{minipage}{\textheight}
\centering
\caption{Part B: preregistered one-sided tests of $\mathrm{regret(baseline)}>\mathrm{regret(comparator)}$, paired by world.
$\Delta$ is the mean paired difference; ``wins'' counts worlds where the comparator has strictly lower regret;
$p_{\rm sign}$ is the exact one-sided sign test and $\widehat{\rm PI}$ the unpaired probability of improvement. $p_{\rm Holm}$
is the Holm-adjusted exact sign-flip $p$-value; we reject $H_0$ when $p_{\rm Holm}<0.05$.}\label{tab:tests}
\scriptsize\setlength{\tabcolsep}{3pt}
\begin{tabular}{llrrrrrrl}
\toprule
hypothesis & config & $n$ & mean $\Delta$ regret (95\% CI) & wins & $p_{\rm sign}$ & $\widehat{\rm PI}$ & $p_{\rm Holm}$ & decision \\
\midrule
H1 proxy is hacked & Corr $k{=}1$ & 10 & +0.356 (+0.273, +0.400) & 10/10 & 0.0010 & 1.00 & 0.0068 & reject $H_0$ \\
H1 proxy is hacked & Corr $k{=}2$ & 10 & -0.000 (-0.001, +0.001) & 4/10 & 0.8281 & 0.44 & 1.0000 & fail to reject \\
H1 proxy is hacked & Corr $k{=}3$ & 10 & +0.001 (-0.000, +0.002) & 6/10 & 0.3770 & 0.66 & 0.7568 & fail to reject \\
H1 proxy is hacked & Corr long & 10 & -0.001 (-0.002, +0.001) & 5/10 & 0.6230 & 0.40 & 1.0000 & fail to reject \\
H1 proxy is hacked & Corr $B{=}1$ & 10 & -0.002 (-0.005, -0.001) & 1/10 & 0.9990 & 0.20 & 1.0000 & fail to reject \\
H1 proxy is hacked & Grid $k{=}2$ & 10 & +0.064 (+0.000, +0.159) & 6/10 & 0.3770 & 0.62 & 0.3809 & fail to reject \\
H1 proxy is hacked & Grid $B{=}1$ & 10 & +0.048 (-0.155, +0.225) & 7/10 & 0.1719 & 0.74 & 1.0000 & fail to reject \\
H2 DR beats plug-in RM & Corr $k{=}1$ & 10 & -0.000 (-0.120, +0.120) & 1/10 & 0.9375 & 0.46 & 1.0000 & fail to reject \\
H2 DR beats plug-in RM & Corr $k{=}2$ & 10 & -0.000 (-0.002, +0.001) & 3/10 & 0.9453 & 0.39 & 1.0000 & fail to reject \\
H2 DR beats plug-in RM & Corr $k{=}3$ & 10 & +0.001 (-0.000, +0.002) & 7/10 & 0.1719 & 0.72 & 0.4648 & fail to reject \\
H2 DR beats plug-in RM & Corr long & 10 & -0.000 (-0.001, +0.001) & 3/10 & 0.9102 & 0.45 & 1.0000 & fail to reject \\
H2 DR beats plug-in RM & Corr $B{=}1$ & 10 & -0.003 (-0.005, -0.001) & 1/10 & 0.9990 & 0.05 & 1.0000 & fail to reject \\
H2 DR beats plug-in RM & Grid $k{=}2$ & 10 & +0.056 (-0.001, +0.138) & 2/10 & 0.6875 & 0.47 & 1.0000 & fail to reject \\
H2 DR beats plug-in RM & Grid $B{=}1$ & 10 & +0.097 (-0.132, +0.322) & 7/10 & 0.0898 & 0.72 & 1.0000 & fail to reject \\
H3a DORL beats DR & Corr $k{=}1$ & 10 & +0.044 (+0.000, +0.124) & 3/10 & 0.1250 & 0.65 & 0.2500 & fail to reject \\
H3a DORL beats DR & Corr $k{=}2$ & 10 & +0.319 (+0.319, +0.320) & 10/10 & 0.0010 & 1.00 & 0.0068 & reject $H_0$ \\
H3a DORL beats DR & Corr $k{=}3$ & 10 & +0.194 (+0.173, +0.216) & 10/10 & 0.0010 & 1.00 & 0.0068 & reject $H_0$ \\
H3a DORL beats DR & Corr long & 10 & +0.293 (+0.257, +0.319) & 10/10 & 0.0010 & 1.00 & 0.0068 & reject $H_0$ \\
H3a DORL beats DR & Corr $B{=}1$ & 10 & +0.206 (+0.176, +0.232) & 10/10 & 0.0010 & 1.00 & 0.0068 & reject $H_0$ \\
H3a DORL beats DR & Grid $k{=}2$ & 10 & +0.128 (-0.025, +0.256) & 7/10 & 0.0898 & 0.69 & 0.1523 & fail to reject \\
H3a DORL beats DR & Grid $B{=}1$ & 10 & +0.141 (-0.014, +0.331) & 4/10 & 0.3438 & 0.55 & 0.2500 & fail to reject \\
H3b DORL beats DR+predictive & Corr $k{=}1$ & 10 & +0.000 (+0.000, +0.000) & 0/10 & 1.0000 & 0.50 & 1.0000 & fail to reject \\
H3b DORL beats DR+predictive & Corr $k{=}2$ & 10 & +0.319 (+0.319, +0.320) & 10/10 & 0.0010 & 1.00 & 0.0068 & reject $H_0$ \\
H3b DORL beats DR+predictive & Corr $k{=}3$ & 10 & +0.194 (+0.173, +0.216) & 10/10 & 0.0010 & 1.00 & 0.0068 & reject $H_0$ \\
H3b DORL beats DR+predictive & Corr long & 10 & +0.292 (+0.258, +0.319) & 10/10 & 0.0010 & 1.00 & 0.0068 & reject $H_0$ \\
H3b DORL beats DR+predictive & Corr $B{=}1$ & 10 & +0.206 (+0.175, +0.232) & 10/10 & 0.0010 & 1.00 & 0.0068 & reject $H_0$ \\
H3b DORL beats DR+predictive & Grid $k{=}2$ & 10 & +0.128 (-0.022, +0.255) & 7/10 & 0.0898 & 0.69 & 0.1055 & fail to reject \\
H3b DORL beats DR+predictive & Grid $B{=}1$ & 10 & +0.086 (+0.008, +0.156) & 7/10 & 0.0352 & 0.73 & 0.0938 & fail to reject \\
H6 audits alone do not anchor & Corr $k{=}1$ & 10 & +0.400 (+0.400, +0.401) & 10/10 & 0.0010 & 1.00 & 0.0068 & reject $H_0$ \\
H6 audits alone do not anchor & Corr $k{=}2$ & 10 & +0.319 (+0.319, +0.320) & 10/10 & 0.0010 & 1.00 & 0.0068 & reject $H_0$ \\
H6 audits alone do not anchor & Corr $k{=}3$ & 10 & +0.196 (+0.175, +0.217) & 10/10 & 0.0010 & 1.00 & 0.0068 & reject $H_0$ \\
H6 audits alone do not anchor & Corr long & 10 & +0.292 (+0.258, +0.318) & 10/10 & 0.0010 & 1.00 & 0.0068 & reject $H_0$ \\
H6 audits alone do not anchor & Corr $B{=}1$ & 10 & +0.203 (+0.174, +0.229) & 10/10 & 0.0010 & 1.00 & 0.0068 & reject $H_0$ \\
H6 audits alone do not anchor & Grid $k{=}2$ & 10 & +0.222 (+0.115, +0.331) & 8/10 & 0.0547 & 0.82 & 0.0068 & reject $H_0$ \\
H6 audits alone do not anchor & Grid $B{=}1$ & 10 & +0.188 (+0.138, +0.233) & 10/10 & 0.0010 & 1.00 & 0.0068 & reject $H_0$ \\
\bottomrule
\end{tabular}
\end{minipage}}
\end{table}

\FloatBarrier
\subsection{Part C: paired comparisons}\label{app:partC-extra}

\begin{table}[ht]
\centering\scriptsize
\resizebox{\linewidth}{!}{\begin{tabular}{llrrrrrr}
\toprule
config & learner & IQM regret [95\% CI] & solved & training charge & total charge & cue probe & decision probe \\
\midrule
Corr $k{=}1$ & supervised, verified outcomes & 0.000 [0.000, 0.040] & 9/10 & 27{,}648 & -- & -- & -- \\
 & supervised, audit replies & 0.000 [0.000, 0.000] & 10/10 & 27{,}648 & -- & -- & -- \\
\midrule
Corr $k{=}2$ & DORL & 0.000 [0.000, 0.001] & 10/10 & 32{,}769 & 34{,}789 & 1.00 & 1.00 \\
 & DR & 0.320 [0.319, 0.321] & 0/10 & 32{,}776 & 34{,}796 & 0.56 & 0.50 \\
 & DORL, no belief input & 0.000 [0.000, 0.006] & 9/10 & 32{,}769 & 34{,}789 & 1.00 & 1.00 \\
 & DR, reads outcome beliefs & 0.320 [0.318, 0.321] & 0/10 & 32{,}776 & 34{,}796 & 0.52 & 0.50 \\
 & DR, $4\times$ episodes & 0.319 [0.286, 0.321] & 1/10 & 130{,}880 & 132{,}900 & 0.59 & 0.55 \\
 & supervised, verified outcomes & 0.320 [0.319, 0.321] & 0/10 & 32{,}767 & -- & -- & -- \\
 & supervised, audit replies & 0.000 [0.000, 0.000] & 10/10 & 32{,}768 & -- & -- & -- \\
\midrule
Corr $k{=}3$ & DORL & 0.066 [0.034, 0.078] & 2/10 & 37{,}886 & 39{,}906 & 1.00 & 1.00 \\
 & DR & 0.256 [0.255, 0.256] & 0/10 & 37{,}884 & 39{,}904 & 0.53 & 0.50 \\
 & DORL, no belief input & 0.111 [0.057, 0.144] & 2/10 & 37{,}886 & 39{,}906 & 1.00 & 1.00 \\
 & DR, reads outcome beliefs & 0.256 [0.256, 0.257] & 0/10 & 37{,}884 & 39{,}904 & 0.56 & 0.50 \\
 & DR, $4\times$ episodes & 0.257 [0.256, 0.257] & 0/10 & 151{,}320 & 153{,}340 & 0.58 & 0.50 \\
 & supervised, verified outcomes & 0.256 [0.256, 0.257] & 0/10 & 37{,}887 & -- & -- & -- \\
 & supervised, audit replies & 0.000 [0.000, 0.000] & 10/10 & 37{,}888 & -- & -- & -- \\
\midrule
Corr long & supervised, verified outcomes & 0.320 [0.319, 0.321] & 0/10 & 32{,}767 & -- & -- & -- \\
 & supervised, audit replies & 0.320 [0.319, 0.321] & 0/10 & 32{,}768 & -- & -- & -- \\
\midrule
Corr $B{=}1$ & DORL & 0.035 [0.013, 0.079] & 6/10 & 42{,}173 & 44{,}335 & 1.00 & 1.00 \\
 & DR & 0.251 [0.250, 0.254] & 0/10 & 35{,}545 & 37{,}574 & 0.55 & 0.50 \\
 & DORL, no belief input & 0.032 [0.016, 0.079] & 8/10 & 40{,}838 & 42{,}990 & 1.00 & 1.00 \\
 & DR, reads outcome beliefs & 0.251 [0.250, 0.253] & 0/10 & 34{,}980 & 37{,}006 & 0.53 & 0.50 \\
 & DORL, training-charge cap & 0.032 [0.012, 0.078] & 7/10 & 32{,}723 & 34{,}875 & 1.00 & 1.00 \\
 & DR, training-charge cap & 0.251 [0.250, 0.255] & 0/10 & 32{,}730 & 34{,}759 & 0.54 & 0.50 \\
\midrule
Grid $k{=}2$ & DORL & 0.080 [0.000, 0.232] & 5/10 & 48{,}009 & 50{,}029 & 0.97 & 0.95 \\
 & DR & 0.319 [0.160, 0.321] & 2/10 & 48{,}018 & 50{,}038 & 0.68 & 0.60 \\
 & DORL, no belief input & 0.085 [0.000, 0.256] & 5/10 & 48{,}047 & 50{,}067 & 0.97 & 0.95 \\
 & DR, reads outcome beliefs & 0.320 [0.287, 0.333] & 1/10 & 47{,}978 & 49{,}998 & 0.67 & 0.61 \\
\midrule
Grid $B{=}1$ & DORL, training-charge cap & 0.196 [0.073, 0.341] & 2/10 & 49{,}074 & 51{,}266 & 0.91 & 0.85 \\
 & DR, training-charge cap & 0.114 [0.000, 0.260] & 5/10 & 49{,}090 & 51{,}250 & 0.85 & 0.79 \\
\bottomrule
\end{tabular}}
\caption{Part C: final regret (IQM over ten worlds, $95\%$ bootstrap interval), worlds solved (regret $\le0.05$), training and
total charge, and decision-time linear-probe accuracy for the cue parities $\chi_{S_j}$ (mean over $j$) and the decision
parity $\chi_{T^\star}$; chance is $0.50$. Supervised controls report their training charge only. In the $B{=}1$ task, probe
targets correlate with the audited reply the policy observes, so probes are interpretable mainly for $B{=}0$.}\label{tab:rev}
\end{table}

Table~\ref{tab:revtests} gives the post hoc paired comparisons of the Part C controls with DORL.

\begin{table}[!ht]
\centering
\caption{Part C paired comparisons (post hoc). ``wins of $B$'' counts worlds where $B$ has strictly lower regret. $p_{\rm Holm}$
is the Holm-adjusted sign-flip $p$-value within each comparison; the two rows against uncapped DR report an unadjusted
sign-flip $p$-value.}\label{tab:revtests}
\footnotesize\setlength{\tabcolsep}{4pt}\renewcommand{\arraystretch}{1.25}
\begin{tabular}{@{}>{\raggedright\arraybackslash}p{5.1cm}l>{\raggedleft\arraybackslash}p{0.35cm}>{\raggedleft\arraybackslash}p{3.6cm}>{\raggedleft\arraybackslash}p{0.95cm}rr@{}}
\toprule
comparison ($A$ vs $B$) & config & $n$ & mean regret$(A)-$regret$(B)$ [95\% CI] & wins of $B$ & $p_{\rm sign}$ & $p_{\rm Holm}$ \\
\midrule
belief input: DORL, no belief input vs DORL & Corr $B{=}1$ & 10 & -0.002 [-0.034, +0.034] & 4/10 & 0.8281 & 0.8906 \\
belief input: DORL, no belief input vs DORL & Corr $k{=}2$ & 10 & +0.005 [-0.001, +0.016] & 2/10 & 0.8125 & 0.7500 \\
belief input: DORL, no belief input vs DORL & Corr $k{=}3$ & 10 & +0.041 [+0.011, +0.068] & 8/10 & 0.0547 & 0.0430 \\
belief input: DORL, no belief input vs DORL & Grid $k{=}2$ & 10 & +0.008 [-0.091, +0.122] & 4/10 & 0.5000 & 0.8906 \\
matched interface: DR, reads outcome beliefs vs DORL & Corr $B{=}1$ & 10 & +0.205 [+0.175, +0.230] & 10/10 & 0.0010 & 0.0039 \\
matched interface: DR, reads outcome beliefs vs DORL & Corr $k{=}2$ & 10 & +0.319 [+0.317, +0.320] & 10/10 & 0.0010 & 0.0039 \\
matched interface: DR, reads outcome beliefs vs DORL & Corr $k{=}3$ & 10 & +0.195 [+0.173, +0.217] & 10/10 & 0.0010 & 0.0039 \\
matched interface: DR, reads outcome beliefs vs DORL & Grid $k{=}2$ & 10 & +0.173 [+0.010, +0.300] & 8/10 & 0.0547 & 0.0264 \\
$4\times$ outcome-only training: DR, $4\times$ episodes vs DORL & Corr $k{=}2$ & 10 & +0.287 [+0.223, +0.320] & 9/10 & 0.0020 & 0.0020 \\
$4\times$ outcome-only training: DR, $4\times$ episodes vs DORL & Corr $k{=}3$ & 10 & +0.195 [+0.174, +0.217] & 10/10 & 0.0010 & 0.0020 \\
equal training-charge cap: DR, training-charge cap vs DORL, training-charge cap & Corr $B{=}1$ & 10 & +0.208 [+0.178, +0.233] & 10/10 & 0.0010 & 0.0020 \\
equal training-charge cap: DR, training-charge cap vs DORL, training-charge cap & Grid $B{=}1$ & 10 & -0.060 [-0.189, +0.085] & 3/10 & 0.9102 & 0.7832 \\
channel, no RL: supervised, verified outcomes vs supervised, audit replies & Corr $k{=}1$ & 10 & +0.040 [+0.000, +0.120] & 1/10 & 0.5000 & 1.0000 \\
channel, no RL: supervised, verified outcomes vs supervised, audit replies & Corr $k{=}2$ & 10 & +0.320 [+0.319, +0.321] & 10/10 & 0.0010 & 0.0039 \\
channel, no RL: supervised, verified outcomes vs supervised, audit replies & Corr $k{=}3$ & 10 & +0.256 [+0.256, +0.257] & 10/10 & 0.0010 & 0.0039 \\
channel, no RL: supervised, verified outcomes vs supervised, audit replies & Corr long & 10 & -0.000 [-0.001, +0.001] & 5/10 & 0.6230 & 1.0000 \\
uncapped DR (main grid, 35{,}545) vs DORL, cap (32{,}723) & Corr $B{=}1$ & 10 & +0.208 & 10/10 & -- & 0.0010 \\
uncapped DR (main grid, 71{,}187) vs DORL, cap (49{,}074) & Grid $B{=}1$ & 10 & -0.001 & 3/10 & -- & 0.5117 \\
\bottomrule
\end{tabular}
\end{table}

\FloatBarrier
\subsection{Part D: all default-price cells}\label{app:llm}

Table~\ref{tab:llm-detail} reports every default-price cell of the language-model study, including the
baselines and the deployment-audit cells that the main text summarizes.

\begin{table}[ht]
\centering\small
\caption{Part D final regret, charge, proxy score, and decision probe for every default-price cell. IQM regret
and its bootstrap interval are over worlds; ``solved'' counts worlds with regret $\le0.05$. Charges are means over
runs, with the total including certification. The proxy score is the mean proxy value of the final policy, and the
decision probe is the accuracy of a linear probe for the decision parity. DR and DORL use ten worlds per cell; the
proxy, outcome, reward-model, and fully verified baselines use five. The $B=1$ evaluation audit rate is zero in every
row.}\label{tab:llm-detail}
\begin{tabular}{lrrrrrrr}
\toprule
 & \multicolumn{2}{c}{regret} & & \multicolumn{2}{c}{charge} & proxy & decision \\
\cmidrule(lr){2-3}\cmidrule(lr){5-6}
learner & IQM & 95\% CI & solved & training & total & score & probe \\
\midrule
\multicolumn{8}{l}{\textit{$k=1$, $B=0$}} \\
proxy & 0.401 & [0.398, 0.405] & 0/5 & 1{,}536 & 3{,}556 & 1.00 & 0.53 \\
outcome & 0.400 & [0.394, 0.408] & 0/5 & 20{,}670 & 22{,}690 & 0.59 & 0.64 \\
reward model & 0.001 & [0.000, 0.276] & 4/5 & 20{,}670 & 22{,}690 & 0.50 & 0.91 \\
DR & 0.000 & [0.000, 0.040] & 9/10 & 20{,}706 & 22{,}726 & 0.50 & 1.00 \\
DORL & 0.000 & [0.000, 0.000] & 10/10 & 20{,}683 & 22{,}703 & 0.50 & 1.00 \\
fully verified & 0.395 & [0.117, 0.404] & 1/5 & 155{,}136 & 157{,}156 & 0.50 & 0.64 \\
\midrule
\multicolumn{8}{l}{\textit{$k=2$, $B=0$}} \\
proxy & 0.321 & [0.318, 0.324] & 0/5 & 1{,}536 & 3{,}556 & 1.00 & 0.51 \\
outcome & 0.320 & [0.316, 0.324] & 0/5 & 24{,}563 & 26{,}583 & 0.50 & 0.54 \\
reward model & 0.316 & [0.095, 0.319] & 1/5 & 24{,}563 & 26{,}583 & 0.50 & 0.64 \\
DR & 0.317 & [0.159, 0.320] & 2/10 & 24{,}587 & 26{,}607 & 0.50 & 0.61 \\
DORL & 0.000 & [0.000, 0.082] & 8/10 & 24{,}524 & 26{,}544 & 0.50 & 0.95 \\
fully verified & 0.322 & [0.319, 0.325] & 0/5 & 155{,}136 & 157{,}156 & 0.50 & 0.58 \\
\midrule
\multicolumn{8}{l}{\textit{$k=3$, $B=0$}} \\
proxy & 0.256 & [0.252, 0.260] & 0/5 & 1{,}536 & 3{,}556 & 1.00 & 0.49 \\
outcome & 0.256 & [0.254, 0.260] & 0/5 & 28{,}345 & 30{,}365 & 0.49 & 0.51 \\
reward model & 0.255 & [0.252, 0.257] & 0/5 & 28{,}345 & 30{,}365 & 0.50 & 0.51 \\
DR & 0.255 & [0.251, 0.257] & 0/10 & 28{,}399 & 30{,}419 & 0.49 & 0.50 \\
DORL & 0.000 & [0.000, 0.127] & 8/10 & 28{,}364 & 30{,}384 & 0.50 & 0.95 \\
fully verified & 0.256 & [0.254, 0.260] & 0/5 & 155{,}136 & 157{,}156 & 0.49 & 0.51 \\
\midrule
\multicolumn{8}{l}{\textit{$k=2$, $B=1$}} \\
DR & 0.352 & [0.351, 0.354] & 0/10 & 25{,}881 & 27{,}901 & 0.50 & 0.51 \\
DORL & 0.061 & [0.030, 0.252] & 7/10 & 25{,}758 & 27{,}778 & 0.50 & 0.92 \\
fully verified & 0.350 & [0.346, 0.354] & 0/5 & 155{,}251 & 157{,}271 & 0.50 & 0.52 \\
\midrule
\multicolumn{8}{l}{\textit{$k=3$, $B=1$}} \\
DR & 0.270 & [0.267, 0.273] & 0/10 & 29{,}844 & 31{,}864 & 0.50 & 0.50 \\
DORL & 0.268 & [0.240, 0.270] & 1/10 & 30{,}453 & 32{,}473 & 0.50 & 0.61 \\
fully verified & 0.271 & [0.269, 0.274] & 0/5 & 155{,}262 & 157{,}282 & 0.50 & 0.51 \\
\bottomrule
\end{tabular}
\end{table}

\FloatBarrier
\subsection{Part E: checkpoint selection}\label{app:partE-extra}

Each run's nine checkpoints, with their $512$ saved evaluation pairs $(\tilde Y,Y)$, define nine policies of known gold
accuracy. For verification budgets $B\in\{64,128,256,512,1024\}$ we select one checkpoint in four ways: proxy-only
ranking (largest mean $\tilde Y$, no verification); uniform IPW or uniform DR ($\lfloor B/9\rfloor$ verifications per
checkpoint); and decision-directed successive elimination (DD-SE). DD-SE runs six rounds and verifies only the checkpoints
that remain candidates for the best. It eliminates a checkpoint when its upper confidence bound falls below the leader's
lower bound, using empirical-Bernstein widths \citep{maurer2009empirical} on DR residuals with $\delta=0.1$. Verified pairs
are drawn with replacement, and we average the gold regret of the selected checkpoint over $200$ replicates per run and
budget and then over the $39$ runs. Table~\ref{tab:partE-selection} gives the results.

\begin{table}[ht]
\centering\small
\caption{Part E offline checkpoint selection: mean gold regret (lower is better) over the $39$ runs, with $200$
selection replicates per run and budget. Proxy-only ranking buys no verification, so its regret does not depend on the
budget.}\label{tab:partE-selection}
\begin{tabular}{rrrrr}
\toprule
budget $B$ & proxy-only & uniform IPW & uniform DR & DD-SE \\
\midrule
64 & 0.1288 & 0.0677 & 0.0283 & 0.0319 \\
128 & 0.1288 & 0.0370 & 0.0192 & 0.0205 \\
256 & 0.1288 & 0.0204 & 0.0151 & 0.0159 \\
512 & 0.1288 & 0.0114 & 0.0112 & 0.0114 \\
1{,}024 & 0.1288 & 0.0070 & 0.0082 & 0.0085 \\
\bottomrule
\end{tabular}
\end{table}

\end{document}